\documentclass[journal]{IEEEtran}
\usepackage{amsmath,amssymb,amsthm,bm,mathtools}
\usepackage{algorithm,algorithmic}
\usepackage{graphicx}
\usepackage{hyperref}
 \usepackage{booktabs,multirow,xcolor}

\newtheorem{theorem}{Theorem}
\newtheorem{proposition}{Proposition}
\newtheorem{lemma}{Lemma}
\newtheorem{corollary}{Corollary}
\newtheorem{assumption}{Assumption}
\newtheorem{remark}{Remark}
\newtheorem{definition}{Definition}

\newcommand{\bA}{\mathbf A}
\newcommand{\bH}{\mathbf H}
\newcommand{\bS}{\mathbf S}
\newcommand{\bL}{\mathbf L}
\newcommand{\bB}{\mathbf B}
\newcommand{\bC}{\mathbf C}
\newcommand{\bU}{\mathbf U}
\newcommand{\bV}{\mathbf V}
\newcommand{\bZ}{\mathbf Z}
\newcommand{\bW}{\mathbf W}
\newcommand{\bR}{\mathbf R}
\newcommand{\bK}{\mathbf K}
\newcommand{\bN}{\mathbf N}
\newcommand{\bM}{\mathbf M}
\newcommand{\bX}{\mathbf X}
\newcommand{\bY}{\mathbf Y}
\newcommand{\bI}{\mathbf I}
\newcommand{\bP}{\mathbf P}
\newcommand{\bE}{\mathbf E}
\newcommand{\bF}{\mathbf F}
\newcommand{\bG}{\mathbf G}
\newcommand{\bJ}{\mathbf J}
\newcommand{\bGamma}{\bm{\Gamma}}
\newcommand{\bQ}{\mathbf Q}

\newcommand{\bUps}{\bm{\Upsilon}}
\newcommand{\cR}{\mathfrak R}
\newcommand{\diag}{\mathrm{diag}}
\newcommand{\bOm}{\boldsymbol\Omega}
\newcommand{\bSig}{\boldsymbol\Sigma}
\newcommand{\bDelta}{\boldsymbol\Delta}
\newcommand{\R}{\mathbb R}
\newcommand{\Sym}{\mathbb S}
\newcommand{\E}{\mathbb E}
\newcommand{\tr}{\operatorname{tr}}
\newcommand{\rank}{\operatorname{rank}}
\newcommand{\range}{\operatorname{range}}
\newcommand{\sign}{\operatorname{sign}}
\newcommand{\sgn}{\operatorname{sgn}}
\newcommand{\cD}{\mathcal D}
\newcommand{\cA}{\mathcal A}
\newcommand{\cI}{\mathcal I}
\newcommand{\cX}{\mathcal X}
\newcommand{\cQ}{\mathcal Q}

\newcommand{\cC}{\mathcal C}
\newcommand{\cH}{\mathcal H}
\newcommand{\cM}{\mathcal M}
\newcommand{\Tt}{\tilde{\mathbf T}}
\newcommand{\PT}{P_{\mathbf T}}
\newcommand{\PTp}{P_{\mathbf T^{\perp}}}
\newcommand{\PR}{P_{\mathcal R}}
\newcommand{\PRp}{P_{\mathcal R^{\perp}}}
\newcommand{\PTt}{P_{\tilde{\mathbf T}}}
\newcommand{\PTtp}{P_{\tilde{\mathbf T}^{\perp}}}
\newcommand{\bmin}{b_-}
\newcommand{\bmax}{b_+}
\providecommand{\res}[1]{{\color{magenta}#1}}% numerical values to be regenerated
\providecommand{\res}[1]{{\color{magenta}#1}}% numerical values to be regenerated
\DeclareMathOperator{\In}{In}

\begin{document}

\title{Robust Latent Variable Graphical Models: Exact Recovery and Fundamental Limits}
% \title{Robust Latent Variable Graphical Models with Signed Low-Rank Corrections}
% \title{Exact Signed Graph and Inertia Recovery in Robust Latent Variable Models}
% \title{Robust Recovery of Sparse Plus Signed Low-Rank Precision Matrices}
% \title{Sparse-Plus-Signed-Low-Rank Graphical Models: Exact Recovery and Fundamental Limits}

\author{\IEEEauthorblockN{Chuansen Peng and Xiaojing Shen}
\thanks{The authors are with the School of Mathematics, Sichuan University, Chengdu 610064, China (e-mail: pengchuansen@stu.scu.edu.cn; shenxj@scu.edu.cn).}}

\maketitle

\begin{abstract}
We study robust estimation of a Gaussian precision matrix that decomposes as a sparse matrix minus a symmetric, possibly indefinite, low-rank matrix, a structure that arises when latent-variable marginalization is combined with conditioning on linear measurements.  The proposed convex estimator replaces the sample covariance by the centered Gaussian measure-transformed (MT) covariance, which preserves the signed graph of the sparse component under the nominal model, and penalizes the low-rank component by the nuclear norm of a congruence transform, an atomic norm that encodes structural side information.  Our main result is a primal--dual witness theorem: under restricted invertibility and irrepresentability conditions on the Fisher information, the raw convex optimizer is unique, has no false edges, recovers the signs of all sufficiently strong edges, and has exactly the true rank and inertia.  These conditions reduce to the graphical-lasso conditions when the low-rank part is absent, and we verify them from primitive incoherence and near-isotropy assumptions.  Over a class on which they hold uniformly, we prove information-theoretic limits for Frobenius estimation, signed graph recovery, and inertia recovery.  In particular, inertia recovery requires $N\gtrsim p/\nu_{\min}^2$ samples, where $\nu_{\min}$ is the smallest nonzero eigenvalue modulus of the low-rank part; the spectral beta-min condition of the witness theorem is necessary up to a $\sqrt{\log p}$ factor; and the Frobenius risk attains the minimax rate up to a factor $\sqrt{\mu(1+\min\{d,r\})\log p}$, where $\mu$ is the incoherence parameter, $d$ the maximal degree, and $r$ the rank.  We further quantify robustness under Huber contamination, with a bias bound that is sharp for the MT functional and vanishes for remote outliers, and give a globally convergent linearized ADMM with closed-form updates.
\end{abstract}


\section{Introduction}
\label{sec:intro}

\IEEEPARstart{T}{he} conditional independence graph of a Gaussian vector is encoded by the zero pattern of its precision matrix \cite{dempster1972covariance,lauritzen1996graphical}, and recovering this graph from samples is a canonical structure-learning problem of high-dimensional statistics \cite{drton2017structure}.  In many applications, however, the precision matrix of the observed vector is not the object that carries the structure of interest.  Three mechanisms routinely separate the two.  First, \emph{latent factors}: marginalizing unobserved common causes subtracts a low-rank positive semidefinite (PSD) term from the precision matrix and renders it dense \cite{chandrasekaran2012latent}.  Second, \emph{linear conditioning}: data are regressed on nuisance signals, conditioned on known linear measurements, or subjected to soft linear constraints, and each such operation \emph{adds} a low-rank PSD term to the precision matrix.  In resting-state functional MRI, for instance, regressing out the global signal is known to induce negative correlations that are absent from the raw data \cite{murphy2009impact,fox2005human}.  Third, \emph{gross corruption}: a fraction of the samples may be arbitrary, as when in-scanner head motion produces spurious but systematic correlations \cite{power2012spurious}.  When the first two mechanisms act together, the precision matrix of the observed vector takes the form
\begin{equation*}
\bOm^*=\bS^*-\bL^*,
\end{equation*}
where $\bS^*$ is sparse and carries the graph of interest, and $\bL^*$ has low rank but, unlike in the classical latent-variable model, is \emph{indefinite} (Proposition~\ref{prop:generative}); the third mechanism means that samples from this model are observed under Huber contamination.

The quantities to be recovered are discrete: the signed edge set of $\bS^*$, the rank of $\bL^*$, and its inertia, i.e., the numbers of positive and negative eigenvalues, which in the generative model of Proposition~\ref{prop:generative} separate the directions dominated by latent marginalization from those dominated by linear conditioning.  Errors in such quantities are not small perturbations.  A spurious edge or a flipped sign changes the scientific conclusion, and a fit that constrains the correction to be PSD can represent the negative directions only by distorting the sparse component or by leaving them unmodeled.  This paper therefore asks two questions.  Under which conditions can a computationally tractable estimator recover the signed graph, the rank, and the inertia \emph{exactly} from contaminated samples?  And which of these conditions are intrinsic to the problem, in the sense that no estimator can dispense with them?

\subsection{What Is Known}
\label{sec:intro_known}

\emph{Sparse Gaussian graphical models.}  Neighborhood selection \cite{meinshausen2006high}, $\ell_1$-penalized likelihood \cite{yuan2007model,banerjee2008model,friedman2008sparse}, and constrained $\ell_1$ minimization \cite{cai2011constrained} estimate a sparse precision matrix from the sample covariance.  The primal--dual witness (PDW) technique, introduced for the Lasso in \cite{wainwright2009sharp}, shows that the graphical lasso recovers the support and signs of the precision matrix under an irrepresentability condition on the Fisher information \cite{ravikumar2011high}; this argument has recently been sharpened and extended to polyhedral atomic-norm penalties \cite{graczyk2025graphical}.  On the converse side, minimax rates for sparse precision estimation \cite{cai2016estimating,ren2015asymptotic} and information-theoretic limits for graph selection \cite{wang2010information,santhanam2012information} delineate what any estimator can achieve.

\emph{Sparse-plus-low-rank decompositions.}  Rank--sparsity incoherence \cite{chandrasekaran2011rank} underlies exact recovery in robust principal component analysis \cite{candes2011robust,hsu2011robust,xu2012robust}.  Chandrasekaran, Parrilo, and Willsky \cite{chandrasekaran2012latent} transferred this idea to latent-variable Gaussian graphical models: they established identifiability of the sparse-minus-PSD decomposition and algebraic consistency of a convex program that combines an $\ell_1$ penalty with a trace penalty under $\bL\succeq\mathbf 0$.  Subsequent work derived Frobenius error bounds \cite{agarwal2012noisy,meng2014learning}, scalable convex and nonconvex algorithms \cite{ma2013alternating,xu2017speeding}, comparisons with PCA-based deconfounding \cite{wang2023learning}, and dynamic extensions \cite{zorzi2016ar,you2024sparse}.  More broadly, atomic norms \cite{chandrasekaran2012convex}, simultaneously structured models \cite{oymak2015simultaneously}, and model-consistency theories for decomposable or partly smooth regularizers \cite{lee2015model,vaiter2015model,vaiter2018model} provide general templates for convex structure recovery.

\emph{Robustness.}  Robust graphical models replace the Gaussian likelihood by a heavy-tailed one \cite{finegold2011robust}, trim observations \cite{yang2015robust}, or plug in rank-based or cellwise-robust covariances \cite{liu2012high,xue2012regularized,loh2018high}; robust scatter estimation has a parallel tradition in signal processing \cite{ollila2014regularized}.  Under Huber contamination \cite{huber1964robust}, the minimax operator-norm rate for covariance estimation is known \cite{chen2018robust}, and a recent line of work develops computationally efficient estimators with near-optimal robustness guarantees \cite{diakonikolas2023algorithmic,oliveira2024improved}.  The probability measure transform (MT) \cite{todros2012measure,todros2017measure} offers a different route: it reweights the data so that remote observations have vanishing influence, and the resulting MT graphical lasso preserves the nominal graph \cite{todros2025measure}.

\emph{Structural side information.}  Temporal ordering, sensor geometry, and known anatomical or relational graphs are routinely encoded by graph Laplacians, both in graph-regularized low-rank models \cite{rao2015collaborative} and in graph learning from smooth signals \cite{kalofolias2016learn,egilmez2017graph}.

\subsection{What Is Missing}
\label{sec:intro_gap}

Taken together, these results leave four gaps that are specific to the model above.

\emph{(G1) Signed corrections and inertia.}  The latent-variable theory \cite{chandrasekaran2012latent,meng2014learning,xu2017speeding,wang2023learning} relies on $\bL\succeq\mathbf 0$ in an essential way.  On the PSD cone the trace is a norm and the inertia is determined by the rank; off the cone the trace is not a norm, the penalty must be replaced by a nuclear norm, and an exact-recovery certificate must preserve not only the rank but also the sign of every nonzero eigenvalue.  Indefinite sparse-plus-low-rank structure does arise for differential networks \cite{na2021estimating}, but the guarantees there concern the estimation error of a nonconvex procedure, not exact recovery of the graph, rank, or inertia.  The general frameworks \cite{lee2015model,vaiter2015model,vaiter2018model} state irrepresentability-type conditions in abstract form, and the polyhedral extension \cite{graczyk2025graphical} exploits a facial structure that the nuclear norm, whose model manifold is curved, does not have.

\emph{(G2) Fundamental limits.}  Converse results for graph selection \cite{wang2010information,santhanam2012information} and precision estimation \cite{cai2016estimating,ren2015asymptotic} concern models without a latent component, and detection limits for spiked covariances \cite{cai2015optimal} concern the low-rank part in isolation.  We are not aware of converses for signed graph or inertia recovery in latent-variable graphical models.  Without them, one cannot tell which of the conditions behind existing upper bounds---irrepresentability, incoherence, beta-min conditions on edges and on eigenvalues, and sample size---reflect the problem and which reflect the proof.

\emph{(G3) Robustness with latent structure.}  Structural guarantees for robust graphical estimators \cite{finegold2011robust,yang2015robust,liu2012high,xue2012regularized,loh2018high}, including the MT graphical lasso \cite{todros2025measure}, are established for sparse precision matrices without a latent component.  Robust covariance estimators \cite{chen2018robust,oliveira2024improved} control operator-norm error, whereas exact structure recovery requires simultaneous control of entrywise and structurally whitened spectral errors of the plug-in covariance, and of the way in which contamination interacts with the decomposition itself: as we show, a single point mass can create an additional low-rank direction.

\emph{(G4) Side information.}  How a structural prior acts on a latent-variable precision model, which directions it penalizes, whether it preserves identifiability and exact recovery, and whether it improves sample efficiency, has not been quantified.

\subsection{Approach}
\label{sec:intro_approach}

We study the convex program
\begin{align*}
(\hat\bS,\hat\bL)\in\arg\min_{\bS-\bL\succ0}\ &\tr\big[(\bS-\bL)\hat\bSig_u\big]-\log\det(\bS-\bL)\\
&+\lambda\|\bS\|_{1,\mathrm{off}}+\gamma\|\bB\bL\bB\|_*,
\end{align*}
where $\hat\bSig_u$ is the empirical centered Gaussian MT covariance and $\bB=(\bI+\tau\bK^\top\bK)^{1/2}$ encodes a structural operator $\bK$.  Each ingredient addresses one gap while keeping the program convex.

\emph{Sign-agnostic, structured low-rank penalty (G1, G4).}  The congruence-transformed nuclear norm treats positive and negative eigenvalues symmetrically, and by Sylvester's law of inertia the congruence preserves rank and inertia.  On the PSD cone it reduces to the weighted trace $\tr(\bB^2\bL)$, so admitting signed corrections costs nothing when the correction happens to be PSD.  The penalty is exactly an atomic norm whose atoms $\pm\mathbf v\mathbf v^\top$ are charged by their structural energy $1+\tau\|\bK\mathbf v\|_2^2$ (Proposition~\ref{prop:atomic}), which makes the action of the prior explicit.

\emph{Graph-preserving robust plug-in (G3).}  Under the nominal Gaussian model, the centered Gaussian weight shifts the precision matrix only by $s^{-2}\bI$ (Proposition~\ref{prop:preservation}).  Hence the estimand graph and the low-rank component are preserved exactly in the transform domain, while the bounded weights yield distribution-free concentration and suppress remote observations.

\emph{One deterministic certificate and a common class (G1--G3).}  The analysis is organized around a single deterministic PDW theorem stated for an \emph{arbitrary} perturbation $\bW$ of the transform-domain covariance.  Sampling error and contamination bias enter only through an entrywise and a $\bB$-whitened spectral norm of $\bW$, so the same theorem serves the nominal and the contaminated settings.  Because the tangent space of the nuclear norm moves with the low-rank component, the restricted problem of the classical PDW argument cannot be posed on a fixed linear space.  We parametrize the low-rank witness by an explicit chart of the rank-$r$ variety that preserves rank and inertia by construction (Lemma~\ref{lem:chart}), and we solve the resulting stationarity equations by Brouwer's fixed-point theorem.  We then exhibit an explicit parameter class on which the witness conditions hold uniformly and prove all converses over the same class, so that upper and lower bounds are compared on identical ground (G2).

\subsection{Contributions}
\label{sec:intro_contrib}

\begin{enumerate}
\item \emph{Model and estimand} (Section~\ref{sec:formulation}).  We give an exact generative mechanism for indefinite low-rank corrections (Proposition~\ref{prop:generative}), identify the estimand graph $\mathsf G(\bS^*)$ as the latent- and measurement-adjusted interaction graph (Definition~\ref{def:graph}, Remark~\ref{rem:graph-interp}), and show that it is uniquely determined by $\bSig^*$ among decompositions that satisfy the witness conditions (Proposition~\ref{prop:identifiability}).

\item \emph{Exact recovery of the raw optimizer} (Section~\ref{sec:pdw}).  Under restricted invertibility and irrepresentability of the Fisher information on the model space, the program has a unique solution that has no false edges, recovers the signs of all edges above an order-$\lambda$ threshold, and has exactly the rank and inertia of the low-rank component (Theorem~\ref{thm:pdw}).  No thresholding is required.  For $r=0$ and $\gamma/\lambda$ large enough, the conditions reduce to those of the graphical lasso \cite{ravikumar2011high} (Remark~\ref{rem:ravikumar}).

\item \emph{Verifiable conditions} (Sections~\ref{sec:primitive}--\ref{sec:nominal}).  The witness conditions follow, with explicit constants, from rank--sparsity incoherence, near-isotropy, and a two-sided window on $\gamma/\lambda$ whose upper end is necessary for the certificate (Theorem~\ref{thm:primitive_2}, Proposition~\ref{prop:window}); they persist under perturbations of the Fisher information with an explicit radius (Theorem~\ref{thm:transfer}).  Explicit Bernstein-type bounds for the MT covariance (Proposition~\ref{prop:noise}) show that the structural prior lowers the effective rank of the $\bB$-whitened dual noise; for a first-order difference operator, the leading noise term decreases by a factor of order $\tau^{-1/4}$ for $1\le\tau\le p^2$ (Proposition~\ref{prop:rB}).

\item \emph{Information-theoretic limits} (Section~\ref{sec:lower}).  Over a common class on which the witness conditions hold uniformly, we prove a Frobenius lower bound of order $\sqrt{pd/N}+\sqrt{pr/N}$ within every inertia stratum (Theorem~\ref{thm:lower}), a Fano bound for signed graph recovery (Theorem~\ref{thm:support_converse}), and a converse for inertia recovery showing that $N\gtrsim p/\nu_{\min}^2$ samples are necessary, where $\nu_{\min}$ is the smallest nonzero eigenvalue modulus of $\bL^*$ (Theorem~\ref{thm:inertia_converse}).  The Frobenius limit is thus insensitive to indefiniteness, whereas indefiniteness creates a new task, inertia recovery, with its own sample complexity.  On the same class, the spectral beta-min condition of Theorem~\ref{thm:pdw} is necessary up to $\sqrt{\log p}$, and the Frobenius risk is within $\sqrt{\mu(1+\min\{d,r\})\log p}$ of the minimax rate (Corollary~\ref{cor:sandwich}, Table~\ref{tab:achieve}).

\item \emph{Robustness} (Section~\ref{sec:robust}).  We prove a bias bound for the MT functional under arbitrary contamination and show that it is sharp for this functional (Theorem~\ref{thm:bias}, Proposition~\ref{prop:bias_sharp}); we give an exact decomposition for point-mass contamination, which explains rank inflation and shows that the bias vanishes for remote outliers (Theorem~\ref{thm:point_mass}); and we obtain a finite-sample guarantee for the \emph{nominal} target that imposes no condition on the contaminated decomposition (Corollary~\ref{cor:robust_nominal}).

\item \emph{Computation and experiments} (Sections~\ref{sec:algo}--\ref{sec:experiments}).  A linearized two-block ADMM with closed-form updates converges globally to the solution (Theorem~\ref{thm:admm}).  Synthetic experiments test each prediction of the theory, and an application to resting-state fMRI illustrates the estimator on real data (Section~\ref{subsec:real}).
\end{enumerate}
The appendices add a Frobenius theory that dispenses with irrepresentability (Appendix~\ref{sec:theory}), a simultaneous guarantee over a finite family of tuning parameters (Appendix~\ref{sec:family}), and an analysis of a misspecified structural operator (Appendix~\ref{sec:misspec}).

\subsection{Scope}
\label{sec:intro_scope}

We delineate what the results do not establish.  First, the comparison between upper and lower bounds is local and unstructured: the common class is a neighborhood of the isotropic model, it carries no structural side information, and on it no choice of $\bB$ improves the minimax rate.  A primitive witness theory for congruences that do not leave the low-rank column space invariant, and a minimax theory over structured subclasses, remain open (Remark~\ref{rem:structured}).  Second, exact recovery is certified in a data-rich regime in which $N$ grows quadratically in $p$, whereas inertia recovery requires only linear growth; the gap, a factor of order $p\log p/(\mu r)$, is traced to two second-order terms of the witness (Remark~\ref{rem:pdw_rates}).  Third, the MT functional is not minimax-robust: its worst-case bias is of order $\epsilon\tr\bSig^*$ rather than the minimax order $\epsilon\|\bSig^*\|_2$ \cite{chen2018robust}, and graph preservation relies on a Gaussian nominal law (Remark~\ref{rem:scope}).

\subsection{Organization}
\label{sec:intro_org}

Section~\ref{sec:formulation} introduces the model, the estimand, and the estimator.  Section~\ref{sec:pdw} states the witness theorem, its primitive sufficient conditions, and the nominal guarantee.  Section~\ref{sec:lower} establishes the information-theoretic limits and compares them with the witness guarantee.  Section~\ref{sec:robust} treats Huber contamination, Section~\ref{sec:algo} presents the algorithm, Section~\ref{sec:experiments} reports the experiments, and Section~\ref{sec:conclusion} concludes.  All proofs are given in the appendices.

\section{Model, Estimand, and Estimator}
\label{sec:formulation}

This section introduces the sparse-minus-signed-low-rank model and its estimand (Section~\ref{sec:model}), the measure-transformed covariance that supplies robustness (Section~\ref{sec:MT}), and the structured convex estimator (Section~\ref{sec:estimator}).  Proofs are deferred to Appendix~\ref{app:proofs_formulation}.

\subsection{Notation}
For $\bM\in\R^{p\times p}$, $\|\bM\|_2$, $\|\bM\|_F$, and $\|\bM\|_*$ are the spectral, Frobenius, and nuclear norms, $\|\bM\|_\infty\triangleq\max_{i,j}|M_{ij}|$, $\|\bM\|_1\triangleq\sum_{i,j}|M_{ij}|$, and $\|\bM\|_{1,\mathrm{off}}\triangleq\sum_{i\ne j}|M_{ij}|$.  $\Sym^p$ is the space of $p\times p$ symmetric matrices with $\langle\bA,\bH\rangle\triangleq\tr(\bA^\top\bH)$.  For a subspace $\mathcal V\subseteq\Sym^p$, $P_{\mathcal V}$ is its orthogonal projector and $\mathcal V^\perp$ its orthogonal complement in $\Sym^p$.  $\sigma_{\min}(\cdot)$, $\sigma_{\max}(\cdot)$ denote extreme singular values and $\bM^\dagger$ the Moore--Penrose inverse.  For $\bN\in\Sym^p$, $\sgn(\bN)\triangleq\Pi_+(\bN)-\Pi_-(\bN)$, where $\Pi_\pm(\bN)$ are the spectral projectors onto the eigenspaces of positive and negative eigenvalues.  The inertia of $\bN$ is the pair (number of positive, number of negative eigenvalues).  For an operator $\mathcal T$ between normed spaces, $\|\mathcal T\|_{a\to b}$ is its induced norm.

\subsection{Sparse-Plus-Signed-Low-Rank Precision Matrices}
\label{sec:model}
Let $\bOm^*=(\bSig^*)^{-1}$ admit the decomposition
\begin{align}
\bOm^*=\bS^*-\bL^*,
\label{eq:model}
\end{align}
where $\bS^*\in\Sym^p$ is sparse off the diagonal and $\bL^*\in\Sym^p$ has rank $r\ll p$.  We do not impose $\bL^*\succeq0$.

\begin{definition}[Estimand graph]\label{def:graph}
For a decomposition~\eqref{eq:model}, the \emph{estimand graph} is
$\mathsf G(\bS^*)\triangleq([p],\mathcal E(\bS^*))$ with
$\mathcal E(\bS^*)\triangleq\{(i,j):i\neq j,\ \bS^*_{ij}\neq0\}$, signed by
$\mathrm{sign}(\bS^*_{ij})$. It is in general different from the conditional
independence graph of the observed vector, whose edge set is
$\mathrm{supp}_{\rm off}(\Omega^*)=\mathrm{supp}_{\rm off}(\bS^*-\bL^*)$ and which
is typically dense when the column space of $\bL^*$ is incoherent. Throughout,
``graph recovery'' refers to $\mathsf G(\bS^*)$.
\end{definition}

Signed corrections arise from an elementary generative mechanism.

\begin{proposition}[A generative model with an indefinite correction]
\label{prop:generative}
Let $(\mathbf x,\mathbf h)\in\R^p\times\R^k$ be jointly Gaussian with zero mean and precision $\begin{pmatrix}\bS&\bC\\ \bC^\top&\bH\end{pmatrix}\succ0$, and let $\mathbf z=\bF^\top\mathbf x+\mathbf e$ with $\bF\in\R^{p\times q}$ and $\mathbf e\sim\mathcal N(\mathbf 0,\sigma_e^2\bI_q)$ independent of $(\mathbf x,\mathbf h)$.  The conditional law of $\mathbf x$ given $\mathbf z=\mathbf 0$ is $\mathcal N(\mathbf 0,\bOm^{-1})$ with
\begin{align}
\bOm=\bS-\bL,\qquad \bL\triangleq\bC\bH^{-1}\bC^\top-\sigma_e^{-2}\bF\bF^\top,
\end{align}
so $\rank(\bL)\le k+q$, and $\bL$ is indefinite whenever neither of $\range(\bC\bH^{-1}\bC^\top)$ and $\range(\bF)$ contains the other.
\end{proposition}

\begin{remark}[Interpretation of $\mathsf G(\bS)$ in the generative model]
\label{rem:graph-interp}
In the setting of Proposition~\ref{prop:generative}, $\bS$ is the precision of
$\mathbf x$ given $\mathbf h$ under the joint law of $(x,h)$. Hence $S_{ij}=0$ if and only if
$x_i\perp x_j\mid(x_{-ij},h)$: $\mathsf G(\bS)$ is the conditional independence
graph among the observed variables given the latent variables, prior to the
measurement channel. Both latent marginalization and conditioning on $\mathbf z$ are
absorbed into $\bL$ and leave $\bS$ unchanged. Note that $\mathsf G(\bS)$ is
\emph{not} the conditional independence graph of $\mathbf x$ given $(\mathbf h,\mathbf z)$: the joint
precision of $(\mathbf x,\mathbf h,\mathbf z)$ shows that $\mathbf x\mid(\mathbf h,\mathbf z)$ has precision
$\bS+\sigma_e^{-2}\mathbf{FF}^\top$. We therefore call $\mathsf G(\bS)$ the intrinsic,
latent- and measurement-adjusted interaction graph. When $\mathbf F=0$, so that
$\bL\succeq0$, it coincides with the graph estimated in~ \cite{chandrasekaran2012latent}.
\end{remark}

\subsection{Measure-Transformed Covariance}
\label{sec:MT}
Let $\mathbf x\sim P$ on $\R^p$.  Given a measurable $u\ge0$ with $0<\E_P[u(\mathbf x)]<\infty$, the transformed probability measure $Q_u$ is defined by
\begin{align}
\frac{dQ_u}{dP}(\mathbf x)=\varphi_u(\mathbf x)\triangleq\frac{u(\mathbf x)}{\E_P[u(\mathbf x)]}.
\label{eq:MT_measure}
\end{align}
The MT mean and MT covariance are the mean and covariance under $Q_u$:
\begin{align}
\mathbf m_u\triangleq\E_P[\mathbf x\varphi_u(\mathbf x)],\
\bSig_u\triangleq\E_P[\mathbf x\mathbf x^\top\varphi_u(\mathbf x)]-\mathbf m_u\mathbf m_u^\top.
\label{eq:MT_moments}
\end{align}
Given i.i.d.\ samples $\{\mathbf x_n\}_{n=1}^N$, let $u_n\triangleq u(\mathbf x_n)$,
\begin{align}
\hat\mu_u&\triangleq\frac1N\sum_{n=1}^Nu_n,\qquad
\hat{\mathbf m}_u\triangleq\frac{1}{N\hat\mu_u}\sum_{n=1}^N\mathbf x_nu_n,\notag\\
\hat\bSig_u&\triangleq\frac{1}{N\hat\mu_u}\sum_{n=1}^N\mathbf x_n\mathbf x_n^\top u_n-\hat{\mathbf m}_u\hat{\mathbf m}_u^\top.
\label{eq:MT_empirical}
\end{align}
We use the centered Gaussian generating function and the Huber model
\begin{align}
u(\mathbf r;s)\triangleq\exp\Big(-\frac{\|\mathbf r\|_2^2}{2s^2}\Big),\ s>0,\
P=(1-\epsilon)\Phi_{\bSig^*}+\epsilon\Psi,
\label{eq:gauss_u}
\end{align}
where $\Phi_{\bSig^*}=\mathcal N(\mathbf 0,\bSig^*)$ and $\Psi$ is arbitrary.  Since $\sup_{t\ge0}te^{-t^2/(2s^2)}=se^{-1/2}$ and $\sup_{t\ge0}t^2e^{-t^2/(2s^2)}=2s^2/e$, pointwise
\begin{align}
\|\mathbf x\|_2u(\mathbf x;s)\le R_1\triangleq se^{-1/2},\
\|\mathbf x\|_2^2u(\mathbf x;s)\le R_2\triangleq 2s^2/e.
\label{eq:R1R2}
\end{align}

The centered Gaussian transform acts on the nominal precision only through its diagonal, and therefore preserves the estimand graph.

\begin{proposition}[Nominal Gaussian structure preservation]
\label{prop:preservation}
Let $\mathbf x\sim\mathcal N(\mathbf 0,\bSig^*)$ and use \eqref{eq:gauss_u}.  Then $Q_u=\mathcal N(\mathbf 0,\bSig_u^*(s))$ with $\bSig_u^*(s)=(\bOm^*+s^{-2}\bI)^{-1}$, and
\begin{align}
\bOm_u^*(s)\triangleq\bSig_u^*(s)^{-1}=\bS_u^*(s)-\bL^*,\ \bS_u^*(s)\triangleq\bS^*+s^{-2}\bI.
\label{eq:shift}
\end{align}
The off-diagonal support and signs of $\bS_u^*(s)$ equal those of $\bS^*$, and $\mu_0\triangleq\E u(\mathbf x;s)=\det(\bI+\bSig^*/s^2)^{-1/2}$.  In particular, $\mathsf G(\bS_u^*(s))=\mathsf G(\bS^*)$.
\end{proposition}

Accordingly, the estimator below targets $(\bS_u^*(s),\bL^*)$ in the transform domain, and $\hat\bS-s^{-2}\bI$ estimates $\bS^*$.

\subsection{Structured Congruence Penalty and the Estimator}
\label{sec:estimator}
Let $\bK\in\R^{q\times p}$ be a fixed structural operator encoding side information on the observed variables, and let $\tau\ge0$.  Define
\begin{gather}
\bB\triangleq\big(\bI+\tau\bK^\top\bK\big)^{1/2}\succ0,\notag\\
\cD(\bL)\triangleq\bB\bL\bB,\qquad
\cD^{-1}(\bM)=\bB^{-1}\bM\bB^{-1}.
\label{eq:BK}
\end{gather}
$\bB$ is symmetric positive definite for every $\tau\ge0$ even when $\bK$ is rectangular or rank deficient.  Hence $\cD$ is an invertible linear map on $\Sym^p$, it is self-adjoint ($\langle\cD(\bX),\bY\rangle=\langle\bX,\cD(\bY)\rangle$), and by Sylvester's law of inertia $\cD(\bL)$ has the same rank and inertia as $\bL$.  We write $\bmin\triangleq\sigma_{\min}(\bB)=(1+\tau\lambda_{\min}(\bK^\top\bK))^{1/2}\ge1$, $\bmax\triangleq\sigma_{\max}(\bB)=(1+\tau\|\bK\|_2^2)^{1/2}$, and $\kappa_B\triangleq\bmax/\bmin$.  For difference and graph-incidence operators $\lambda_{\min}(\bK^\top\bK)=0$, so $\bmin=1$.

Using the centered empirical MT covariance \eqref{eq:MT_empirical}, define
\begin{align}
(\hat\bS,\hat\bL)\in{}&\arg\min_{\substack{\bS,\bL\in\Sym^p\\ \bS-\bL\succ0}}
\ \tr\big[(\bS-\bL)\hat\bSig_u\big]-\log\det(\bS-\bL)\notag\\
&+\lambda\|\bS\|_{1,\mathrm{off}}+\gamma\|\cD(\bL)\|_*.
\label{eq:estimator}
\end{align}
The program is convex.  For $\tau=0$ the penalty is $\|\bL\|_*$.  For $\bL\succeq0$, $\|\cD(\bL)\|_*=\tr(\bB^2\bL)=\tr(\bL)+\tau\tr(\bK\bL\bK^\top)$.

\begin{proposition}[Atomic-norm representation]
\label{prop:atomic}
For every $\bL\in\Sym^p$,
\begin{align}
\|\cD(\bL)\|_*=\min\Big\{&\sum_{k}|c_k|\,\|\bB\mathbf v_k\|_2^2:\notag\\
&\bL=\sum_{k}c_k\mathbf v_k\mathbf v_k^\top,\ \|\mathbf v_k\|_2=1\Big\},
\label{eq:atomic}
\end{align}
the minimum being over finite decompositions and attained.  In particular, $\|\bB\mathbf v\|_2^2=1+\tau\|\bK\mathbf v\|_2^2$.
\end{proposition}

Proposition~\ref{prop:atomic} makes precise the sense in which $\tau$ and $\bK$ act: a spectral direction $\mathbf v$ is charged by its structural energy $1+\tau\|\bK\mathbf v\|_2^2$, so directions nearly annihilated by $\bK$ retain the ordinary nuclear-norm geometry while rough directions are penalized more.

\subsubsection{Interpretable structural operators}
\label{sec:K_choices}
The operator $\bK$ should encode credible external information and is treated as fixed by the theory.

\emph{Ordered variables.}  For time points, frequency bins, or sensors along a line, let $\bK_{\mathrm{diff}}\in\R^{(p-1)\times p}$ be the first-order difference operator, $(\bK_{\mathrm{diff}}\mathbf v)_i=v_{i+1}-v_i$.  Then $\|\bK_{\mathrm{diff}}\mathbf v\|_2^2=\sum_{i=1}^{p-1}(v_{i+1}-v_i)^2$ and $\bK_{\mathrm{diff}}^\top\bK_{\mathrm{diff}}$ is the path-graph Laplacian; second-order differences $(\bK_{\mathrm{diff},2}\mathbf v)_i=v_{i+1}-2v_i+v_{i-1}$ penalize curvature instead.

\emph{Known graph topology.}  If an auxiliary weighted graph $\mathcal G_0=(V,\mathcal E_0,\mathbf W_{\mathcal G})$ is available (sensor proximity, anatomical connectivity, a pathway network), let $\bC_{\mathcal G}$ be an oriented incidence matrix and $\bK_{\mathcal G}=\mathbf W_{\mathcal G}^{1/2}\bC_{\mathcal G}$.  Then $\bK_{\mathcal G}^\top\bK_{\mathcal G}=\mathbf L_{\mathcal G}$, the weighted Laplacian, $\|\bK_{\mathcal G}\mathbf v\|_2^2=\sum_{\{i,j\}\in\mathcal E_0}w_{ij}(v_i-v_j)^2$, and, with $\mathbf L_{\mathcal G}=\mathbf Q\bm\Lambda\mathbf Q^\top$, $\bB=\mathbf Q(\bI+\tau\bm\Lambda)^{1/2}\mathbf Q^\top$ is a graph-spectral filter that charges high graph frequencies more heavily.

\emph{Spatial lattices.}  For $p=n_xn_y$ variables on a grid, $\bK_{\mathrm{sp}}=[\bK_x;\bK_y]$ with $\bK_x=\bI_{n_y}\otimes\mathbf D_x$ and $\bK_y=\mathbf D_y\otimes\bI_{n_x}$, where $\mathbf D_x,\mathbf D_y$ are one-dimensional difference matrices; $\bK_{\mathrm{sp}}^\top\bK_{\mathrm{sp}}$ is the grid Laplacian.  Sobel or biharmonic stencils fit the same construction.

\begin{remark}[Strength of the prior and conditioning]
\label{rem:conditioning}
$\kappa_B^2=(1+\tau\lambda_{\max}(\bK^\top\bK))/(1+\tau\lambda_{\min}(\bK^\top\bK))\le1+\tau\|\bK\|_2^2$.  The theory below shows that $\bB$ enters the error bounds in two opposite ways: through constants such as $\bmax$ (a cost) and through the $\bB$-whitened noise level and the compatibility index $\chi_B$ of Theorem~\ref{thm:primitive} (a potential gain).  Normalizing $\|\bK\|_2=1$ makes $\tau$ comparable across operator families.  For $\tau=0$ we recover the unstructured nuclear norm.
\end{remark}

\section{Exact Recovery of the Signed Graph, Rank, and Inertia}
\label{sec:pdw}

This section contains the main result of the paper.  We state it deterministically, for a generic transform-domain target and an arbitrary perturbation $\bW$ of its covariance (Theorem~\ref{thm:pdw}); the same statement serves the nominal model and, in Section~\ref{sec:robust}, contaminated data.  We then show that its two structural assumptions follow from primitive incoherence and near-isotropy conditions (Theorem~\ref{thm:primitive_2}), and combine it with explicit concentration bounds for the MT covariance (Corollary~\ref{cor:pdw_nominal}).  Proofs are given in Appendices~\ref{app:tools}--\ref{sec:conc}.

\subsection{Target, Model Spaces, and Witness Conditions}
\label{sec:setup}

The analysis in Sections~\ref{sec:setup}--\ref{sec:primitive} is deterministic.  Let $\bSig_0\succ0$ satisfy
\begin{align}
\bOm_0\triangleq\bSig_0^{-1}=\bS_0-\bL_0,\ \bS_0,\bL_0\in\Sym^p,\ r\triangleq\rank(\bL_0),
\label{eq:target}
\end{align}
and write $\bDelta_S\triangleq\hat\bS-\bS_0$, $\bDelta_L\triangleq\hat\bL-\bL_0$, $\bDelta\triangleq\bDelta_S-\bDelta_L$, $\bW\triangleq\hat\bSig_u-\bSig_0$, $\omega_-\triangleq\lambda_{\min}(\bOm_0)$, $\omega_+\triangleq\lambda_{\max}(\bOm_0)$.  For the nominal model, $(\bSig_0,\bS_0,\bL_0)=(\bSig_u^*,\bS_u^*,\bL^*)$.  Nothing in the deterministic results requires $\bW$ to be a pure sampling error; Section~\ref{sec:robust} exploits this.  Let
\begin{align}
\mathcal E&\triangleq\{(i,j):i\ne j,\ (\bS_0)_{ij}\ne0\},\notag\\
\mathcal R&\triangleq\mathcal E\cup\{(i,i)\}_{i\in[p]},\quad m\triangleq|\mathcal R|,
\label{eq:support}
\end{align}
let $d$ be the maximal number of nonzero off-diagonal entries in a row of $\bS_0$, so $m\le p(d+1)$, and identify $\mathcal R$ with the subspace of $\Sym^p$ supported on $\mathcal R$.  For $\bA\in\mathcal R$, $\|\bA\|_2\le\max_i\sum_j|A_{ij}|\le\mu_{\mathcal R}\|\bA\|_\infty$ with $\mu_{\mathcal R}\triangleq d+1$.

Let $\bL_0=\bU\bGamma\bU^\top$ with $\bU\in\R^{p\times r}$ orthonormal and $\bGamma$ nonsingular, $\bP_U\triangleq\bU\bU^\top$, and
\begin{align}
\mathbf T\triangleq\{\bU\bY^\top+\bY\bU^\top:\bY\in\R^{p\times r}\}.
\label{eq:T}
\end{align}
Let $\tilde\bU$ be an orthonormal basis of $\range(\bB\bU)=\range(\cD(\bL_0))$, $\bP_{\tilde U}\triangleq\tilde\bU\tilde\bU^\top$, and $\Tt$ the corresponding tangent space.  For a rank-$r$ projector $\bP$ and its tangent space $\mathbf T_{\bP}$, $P_{\mathbf T_{\bP}}(\bX)=\bP\bX+\bX\bP-\bP\bX\bP=\bP\bX+(\bI-\bP)\bX\bP$ and $P_{\mathbf T_{\bP}^\perp}(\bX)=(\bI-\bP)\bX(\bI-\bP)$; hence
\begin{align}
\|P_{\mathbf T_{\bP}}\bX\|_2\le2\|\bX\|_2,\ \|P_{\mathbf T_{\bP}^\perp}\bX\|_2\le\|\bX\|_2,\notag\\
\rank(P_{\mathbf T_{\bP}}\bX)\le2r.
\label{eq:proj_facts}
\end{align}

Two scalar quantities measure how the low-rank column space interacts with the sparse support and with the congruence.
Let $\xi(\mathbf T)\triangleq\max\{\|\bX\|_\infty:\bX\in\mathbf T,\|\bX\|_2\le1\}$; for $\bX\in\mathbf T$, $|X_{ij}|\le|\mathbf e_i^\top\bP_U\bX\mathbf e_j|+|\mathbf e_i^\top(\bI-\bP_U)\bX\bP_U\mathbf e_j|$, so $\xi(\mathbf T)\le2\max_i\|\bU^\top\mathbf e_i\|_2$ \cite{chandrasekaran2011rank}.  Define the $\bB$-compatibility index
\begin{align}
\chi_B\triangleq\|(\bI-\bP_U)\bB^2\bP_U\|_2,
\label{eq:chiB}
\end{align}
which vanishes iff $\range(\bU)$ is invariant under $\bB^2=\bI+\tau\bK^\top\bK$ (e.g.\ if $\bU$ is spanned by eigenvectors of a graph Laplacian $\bK^\top\bK$), and in particular when $\tau=0$.

\emph{Reparametrization.}
Since $\cD$ is a linear bijection, \eqref{eq:estimator} is equivalent, via $\bN=\cD(\bL)$, to
\begin{align}
\min_{\bS,\bN\in\Sym^p}F(\bS,\bN)\triangleq f\big(\bS-\cD^{-1}(\bN)\big)+\lambda\|\bS\|_{1,\mathrm{off}}+\gamma\|\bN\|_*,
\label{eq:N_program}
\end{align}
with $f(\bOm)=\tr(\bOm\hat\bSig_u)-\log\det\bOm$.  Let $\bN_0\triangleq\cD(\bL_0)=\tilde\bU\tilde\bGamma\tilde\bU^\top$, where $\tilde\bU\in\R^{p\times r}$ is orthonormal and $\tilde\bGamma\in\Sym^r$ is nonsingular, let $\tilde\bU_\perp$ complete $\tilde\bU$ to an orthogonal matrix, and let
\begin{align}
\nu\triangleq\min_k|\lambda_k(\tilde\bGamma)|>0,\ \bP_0\triangleq\tilde\bU\tilde\bU^\top.
\end{align}
By Sylvester's law, $\bN_0$ and $\bL_0$ have the same inertia.  Write $\bG(\bOm)\triangleq\nabla f(\bOm)=\hat\bSig_u-\bOm^{-1}$ and $\cI(\bX)\triangleq\bSig_0\bX\bSig_0$, the Hessian of $f$ at $\bOm_0$.  On $\Sym^p\times\Sym^p$ define
\begin{gather*}
\cA(\bA,\bY)\triangleq\bA-\cD^{-1}(\bY),\qquad \cA^*(\bX)=(\bX,-\cD^{-1}(\bX)),\\
\cM\triangleq\cA^*\cI\cA,
\end{gather*}
the model space $\cX\triangleq\mathcal R\times\Tt$ with orthogonal complement $\cX^\perp=\mathcal R^\perp\times\Tt^\perp$, and the mixed norm
\begin{align}
\|(\bA,\bY)\|_g\triangleq\max\{\|\bA\|_\infty/\lambda,\ \|\bY\|_2/\gamma\}.
\label{eq:gnorm}
\end{align}
Let $\cH\triangleq P_{\cX}\cM|_{\cX}:\cX\to\cX$, the Fisher information restricted to the model space.

\begin{assumption}[Restricted invertibility]
\label{ass:invert}
$\cH$ is invertible, and $\kappa_H\triangleq\sup\{\|\cH^{-1}\mathbf q\|_g:\mathbf q\in\cX,\ \|\mathbf q\|_g\le1\}<\infty$.
\end{assumption}

\begin{assumption}[Irrepresentability]
\label{ass:irrep}
For some $\alpha\in(0,1]$, $\|P_{\cX^\perp}\cM\cH^{-1}\mathbf q\|_g\le(1-\alpha)\|\mathbf q\|_g$ for all $\mathbf q\in\cX$.
\end{assumption}

$\cH$ is self-adjoint and positive semidefinite on $\cX$ with respect to $\langle\cdot,\cdot\rangle$, and it is invertible iff $\cA$ is injective on $\cX$, i.e.\ iff $\mathcal R\cap\cD^{-1}(\Tt)=\mathcal R\cap\mathbf T=\{0\}$ (Lemma~\ref{lem:congruence}(ii)).  The norms in Assumptions~\ref{ass:invert}--\ref{ass:irrep} depend on $(\lambda,\gamma)$ only through $\gamma/\lambda$.

\begin{remark}[Reduction to the graphical lasso]
\label{rem:ravikumar}
If $r=0$, then $\Tt=\{0\}$, $\Tt^\perp=\Sym^p$, $\cX=\mathcal R\times\{0\}$, and $\cH=\PR\cI\PR$ is the block $\cI_{\mathcal R\mathcal R}$ of the Fisher information $\bSig_0\otimes\bSig_0$, so $\kappa_H=\|\cI_{\mathcal R\mathcal R}^{-1}\|_{\infty\to\infty}$.  For $\mathbf q\in\cX$ and $\bA\triangleq\cI_{\mathcal R\mathcal R}^{-1}\mathbf q$, $P_{\cX^\perp}\cM\cH^{-1}\mathbf q=(\PRp\cI\bA,\,-\cD^{-1}\cI\bA)$.  The first component gives exactly the irrepresentability condition $\|\cI_{\mathcal R^c\mathcal R}\cI_{\mathcal R\mathcal R}^{-1}\|_{\infty\to\infty}\le1-\alpha$ of \cite{ravikumar2011high}.  The second component is the dual feasibility of $\bL=\mathbf 0$ under the nuclear-norm penalty; since $\|\cD^{-1}\cI\bA\|_2\le\bmin^{-2}\omega_-^{-2}\mu_{\mathcal R}\|\bA\|_\infty$, it holds whenever $\gamma(1-\alpha)\ge\kappa_H\mu_{\mathcal R}\lambda/(\bmin^2\omega_-^2)$.  Thus, for $r=0$ and $\gamma/\lambda$ large enough, Assumptions~\ref{ass:invert}--\ref{ass:irrep} coincide with the conditions of \cite{ravikumar2011high}.  The proof of Theorem~\ref{thm:pdw} below covers $r=0$ with $\bY\equiv\mathbf 0$, $\nu\triangleq+\infty$, and $\xi=0$.
\end{remark}

\subsection{Main Result}
\label{sec:pdw_main}

Define the scalars
\begin{align}
&\varrho\triangleq2\kappa_H,\quad h\triangleq\mu_{\mathcal R}\lambda+\gamma/\bmin^2,\quad \xi\triangleq\varrho\gamma/\nu,\quad \zeta\triangleq\varrho h/\omega_-,\notag\\
&\Lambda\triangleq\max\{\lambda^{-1},\,4\bmin^{-2}\gamma^{-1}\}/\omega_-,\quad
\mathfrak n\triangleq16\xi^2+\zeta(3\zeta+19\xi)\Lambda,\notag\\
&\varepsilon_W\triangleq\max\{\|\bW\|_\infty/\lambda,\ \|\cD^{-1}(\bW)\|_2/\gamma\}.
\label{eq:pdw_scalars}
\end{align}
Here $\xi$ compares the low-rank resolution $\kappa_H\gamma$ with the spectral gap $\nu$, $\zeta$ compares the spectral size of the error with $\omega_-$, and $\mathfrak n$ collects all second-order terms.

\begin{theorem}[Primal--dual witness]
\label{thm:pdw}
Let $r\ge1$ (see Remark~\ref{rem:ravikumar} for $r=0$).  Suppose Assumptions~\ref{ass:invert}--\ref{ass:irrep} hold and
\begin{align}
&\text{(C1) }\xi\le\tfrac1{32},\ \zeta\le\tfrac14;\qquad
\text{(C2) }\varepsilon_W\le\tfrac{\alpha}{12};\qquad
\text{(C3) }\mathfrak n\le\tfrac\alpha8;\notag\\
&\text{(C4) }\min_{(i,j)\in\mathcal E}|(\bS_0)_{ij}|>2\kappa_H\lambda.
\label{eq:pdw_conditions}
\end{align}
Then \eqref{eq:estimator} has a unique minimizer $(\hat\bS,\hat\bL)$, and:
\begin{enumerate}
\item[(a)] $\hat S_{ij}=0$ for all $(i,j)\notin\mathcal R$, and $\sign(\hat S_{ij})=\sign((\bS_0)_{ij})$ for all $(i,j)\in\mathcal E$;
\item[(b)] $\rank(\hat\bL)=r$, and $\hat\bL$ has the inertia of $\bL_0$;
\item[(c)] $\|\hat\bS-\bS_0\|_\infty\le2\kappa_H\lambda$ and $\|\cD(\hat\bL-\bL_0)\|_2\le\frac83\kappa_H\gamma$; consequently $\|\hat\bS-\bS_0\|_F\le2\kappa_H\sqrt m\,\lambda$ and $\|\hat\bL-\bL_0\|_F\le\frac83\sqrt{2r}\,\kappa_H\gamma/\bmin^2$;
\item[(d)] the dual certificate is strictly feasible with slack $\alpha/4$: $|\hat G_{ij}|\le(1-\frac\alpha4)\lambda$ for $i\ne j$, $(i,j)\notin\mathcal R$, and $\|P_{\hat\Tt^\perp}\cD^{-1}(\hat\bG)\|_2\le(1-\frac\alpha4)\gamma$, where $\hat\bG=\bG(\hat\bS-\hat\bL)$ and $\hat\Tt=\Tt_{\cD(\hat\bL)}$.
\end{enumerate}
\end{theorem}

Theorem~\ref{thm:pdw} certifies the raw convex optimizer: no thresholding is needed to remove false edges or to read off the rank and inertia.  Conditions (C1) and (C4) are beta-min conditions on the smallest eigenvalue modulus $\nu$ of $\cD(\bL_0)$ and on the edge magnitudes, respectively; (C2) calibrates $(\lambda,\gamma)$ to the dual-norm size of $\bW$; and (C3) collects all second-order terms and acts as a sample-size condition (Remark~\ref{rem:pdw_rates}).  Since $\bW$ is arbitrary, the theorem applies verbatim to a biased covariance, which is how contamination is handled in Section~\ref{sec:robust}.

\emph{Proof outline.}  The proof (Appendix~\ref{app:pdw}) follows the primal--dual witness strategy with one essential modification.  Because the tangent space of the nuclear norm moves with the low-rank component, the restricted problem cannot be posed on the fixed linear space $\cX$ alone; instead, the low-rank witness is parametrized by an explicit chart $\bY\mapsto\bN(\bY)$ of the rank-$r$ variety around $\bN_0$ (Lemma~\ref{lem:chart}), which preserves rank and inertia by construction.  The stationarity conditions on $\cX$ become a fixed-point equation $\mathbf u=-\cH^{-1}\mathbf q(\mathbf u)$ whose map sends a $\|\cdot\|_g$-ball into itself under (C1)--(C3), and Brouwer's theorem yields a solution.  Assumption~\ref{ass:irrep} then gives strict dual feasibility on $\cX^\perp$, and an optimality-and-uniqueness certificate (Lemma~\ref{lem:kkt}) identifies the witness with the unique minimizer.

Applied deterministically with $\bW=\mathbf 0$, Theorem~\ref{thm:pdw} shows that the estimand is well defined within the class of decompositions satisfying the witness conditions.

\begin{proposition}[Identifiability of the estimand]\label{prop:identifiability}
Let $\bm \Sigma_0\succ0$ and let $(\bS_0,\bL_0)$ and $(\bS_1,\bL_1)$ be two decompositions
of $\bm\Sigma_0^{-1}$ of the form~\eqref{eq:target}, possibly of different ranks. Suppose that,
for some $\bB\succ0$ and $\rho>0$, each satisfies
Assumptions~\ref{ass:invert}--\ref{ass:irrep} on its own model space for
$\gamma/\lambda=\rho$. Then $(\bS_0,\bL_0)=(\bS_1,\bL_1)$. In particular,
$\mathsf G(\bS_0)$ and the inertia of $\bL_0$ are uniquely determined by $\bm\Sigma_0$
within the class of decompositions satisfying the witness conditions.
\end{proposition}

\subsection{Verifiable Sufficient Conditions}
\label{sec:primitive}

Assumptions~\ref{ass:invert}--\ref{ass:irrep} are stated in terms of the restricted
Fisher operator. Conditioning of $\bm\Sigma_0$ alone cannot imply them: already for
$r=0$, Assumption~\ref{ass:invert} reduces to the irrepresentability condition
of~\cite{ravikumar2011high} (Remark~\ref{rem:ravikumar}), which is not implied by a
bounded condition number. We now show that rank--sparsity incoherence, a window
on $\gamma/\lambda$, and near-isotropy of $\bm\Sigma_0$ in the
$\|\cdot\|_{\infty\to\infty}$ norm (maximum absolute row sum) suffice, with
explicit constants. We first treat congruences that leave the low-rank column space invariant, a class that contains $\bB=\bI$, and then show (Theorem~\ref{thm:transfer} in Appendix~\ref{app:primitive}) that the conditions extend, with an explicit radius, to neighborhoods of any reference model at which they are known to hold.

\begin{theorem}[Primitive conditions for the witness assumptions]
\label{thm:primitive_2}
Assume that $\bB$ commutes with $\bP_U$, equivalently $\chi_B=0$; this holds for $\bB=\bI$ and whenever $\range(\bU)$ is spanned by eigenvectors of $\bK^\top\bK$.  Let $\sigma_0>0$, put $\delta_\infty\triangleq\|\bm\Sigma_0/\sigma_0-\bI\|_{\infty\to\infty}$, and suppose
\begin{equation}\label{eq:primitive_2}
\text{(P1)}\ \ \delta_\infty\le \frac{1}{100\,\bmax^4},
\qquad
\text{(P2)}\ \ \frac{4\mu_{\mathcal R}}{\bmin^2}\le \frac{\gamma}{\lambda}\le \frac{1}{4\,\bmax^2\,\xi(\mathbf T)}.
\end{equation}
Then Assumption~\ref{ass:invert} holds with $\kappa_H\le 2\bmax^4/\sigma_0^2$, and Assumption~\ref{ass:irrep} holds with $\alpha\ge 2/5$.  The window in (P2) is nonempty if and only if $16\,\kappa_B^2\,\mu_{\mathcal R}\,\xi(\mathbf T)\le 1$, which holds whenever $32\,\kappa_B^2(d+1)\max_i\|\bU^\top e_i\|_2\le 1$.  For $\bB=\bI$, (P1)--(P2) read $\delta_\infty\le1/100$ and $4\mu_{\mathcal R}\le\gamma/\lambda\le1/(4\xi(\mathbf T))$, and $\kappa_H\le2/\sigma_0^2$.
\end{theorem}

The two-sided window in (P2) is not an artifact of the sufficient condition: its upper end is necessary for Assumption~\ref{ass:irrep}.

\begin{proposition}[The upper end of the window is necessary]
\label{prop:window}
Let $\bB=\bI$ and $\bm\Sigma_0=\sigma_0\bI$, let
$\xi_{\rm off}(\mathcal T)\triangleq\max\{\|P_{\mathcal R^\perp}\bX\|_\infty:\bX\in\mathcal T,\ \|\bX\|_2\le1\}$,
and assume $\beta=2\mu_{\mathcal R}\xi(\mathcal T)\le1/2$. If
Assumption~\ref{ass:irrep} holds with margin $\alpha$, then
\begin{equation}\label{eq:window-nec}
\big(\xi_{\rm off}(\mathcal T)-2\beta\,\xi(\mathcal T)\big)\,\frac{\gamma}{\lambda}\le1-\alpha .
\end{equation}
Moreover, $\xi_{\rm off}(\mathcal T)\ge\frac12\max_i\|\bU^\top e_i\|_2\ge\frac12\sqrt{r/p}$
whenever $d\le p-2$. In particular, if $4\beta\,\xi(\mathcal T)\le\xi_{\rm off}(\mathcal T)$,
then $\gamma/\lambda\le 4(1-\alpha)\sqrt{p/r}$.
\end{proposition}

Proposition~\ref{prop:window} concerns the certificate, not the estimation problem: it shows that Assumption~\ref{ass:irrep} caps the tuning ratio admissible for our mixed-norm witness at order $\sqrt{p/r}$.  It does not show that every estimator, or every certificate, must pay the resulting factor; Remark~\ref{rem:optimality} traces how this cap propagates to the rates.

\subsection{Nominal Guarantee and the Role of the Structural Prior}
\label{sec:nominal}

Under the nominal model, $\bW=\hat\bSig_u-\bSig_u^*$ is a pure sampling error, and (C2) requires control of $\|\bW\|_\infty$ and of the $\bB$-whitened spectral norm $\|\cD^{-1}(\bW)\|_2=\|\bB^{-1}\bW\bB^{-1}\|_2$.  Appendix~\ref{sec:conc} establishes Bernstein-type bounds for both, with explicit dependence on $p$, on the MT scale $s$, and on $\bB$; the following summary suffices here.

\begin{proposition}[Noise levels of the centered MT covariance]
\label{prop:noise}
Under the nominal model, let $\bar\sigma\triangleq\max_i(\bSig_u^*)_{ii}$, $\bG_B\triangleq\bB^{-1}\bSig_u^*\bB^{-1}$, $\mathsf r_B\triangleq\tr\bG_B/\|\bG_B\|_2$, $\ell_V\triangleq\log(6p^2/\eta)$, $\ell_a\triangleq\log(6p/\eta)$, and $\ell_2'\triangleq\log(6(p+1)/\eta)$, and let $\varepsilon_\infty$ and $\varepsilon_{2,B}$ be the explicit quantities in \eqref{eq:eps_inf_nom}--\eqref{eq:eps2_nom}.  If $t_\mu\le\mu_0/2$, with $t_\mu$ as in \eqref{eq:eps_inf}, then, with probability at least $1-\eta$ each, $\|\bW\|_\infty\le\varepsilon_\infty$ and $\|\cD^{-1}(\bW)\|_2\le\varepsilon_{2,B}$.  If moreover $s^2\ge\tr\bSig^*$, then, for an absolute constant $C$,
\begin{align*}
\varepsilon_\infty&\le C\Big(\bar\sigma\sqrt{\tfrac{\ell_V}{N}}+\tfrac{s^2\ell_V}{N}+\tfrac{\bar\sigma\ell_a}{N}+\tfrac{s^2\ell_a^2}{N^2}\Big),\\
\varepsilon_{2,B}&\le C\Big(\|\bG_B\|_2\sqrt{\tfrac{(\mathsf r_B+2)\ell_2'}{N}}+\tfrac{s^2\ell_2'}{\bmin^2N}\\
&\qquad\quad+\tfrac{\tr\bG_B\,\ell_2'}{N}+\tfrac{s^2\ell_2'^2}{\bmin^2N^2}\Big).
\end{align*}
\end{proposition}

The leading terms are the parametric rates, and the $s^2/N$ terms are the price of the bounded, redescending weight.  The low-rank penalty must dominate a $\bB$-whitened noise of effective rank $\mathsf r_B$, and this is where the structural prior pays off.

\begin{proposition}[Structural reduction of the dual-noise level]
\label{prop:rB}
Let $\sigma_-\bI\preceq\bSig_u^*\preceq\sigma_+\bI$, let $0\le\varsigma_1\le\dots\le\varsigma_p$ be the eigenvalues of $\bK^\top\bK$, and recall $\bG_B=\bB^{-1}\bSig_u^*\bB^{-1}$ and $\mathsf r_B=\tr\bG_B/\|\bG_B\|_2$.  Then $\tr(\bB^{-2})=\sum_{k=1}^p(1+\tau\varsigma_k)^{-1}$ and:
\begin{enumerate}
\item[(i)] $\|\bG_B\|_2\sqrt{\mathsf r_B+2}\le\sigma_+\big(\bmin^{-2}\tr(\bB^{-2})+2\bmin^{-4}\big)^{1/2}$, whereas $\|\bG_I\|_2\sqrt{\mathsf r_I+2}\ge\sigma_-\sqrt p$;
\item[(ii)] $\mathsf r_B\le(\sigma_+/\sigma_-)\,\bmin^2\tr(\bB^{-2})$, whereas $\mathsf r_I\ge(\sigma_-/\sigma_+)\,p$;
\item[(iii)] for the first-order difference operator $\bK=\bK_{\mathrm{diff}}$ of Section~\ref{sec:K_choices}, $\bmin=1$ and $\tr(\bB^{-2})\le\min\{p,\ 1+\pi p/(4\sqrt\tau)\}$.
\end{enumerate}
Consequently, for $\bK=\bK_{\mathrm{diff}}$, the leading term of $\varepsilon_{2,B}$ in Corollary~\ref{cor:explicit} is smaller than that of $\varepsilon_{2,I}$ by at least the factor $(\sigma_+/\sigma_-)\big((3+\pi p/(4\sqrt\tau))/p\big)^{1/2}$, which is of order $\tau^{-1/4}$ for $1\le\tau\le p^2$.
\end{proposition}

\begin{corollary}[Nominal Gaussian signed-graph and rank recovery]
\label{cor:pdw_nominal}
Under the nominal model with target $(\bS_u^*,\bL^*)$, let $\varepsilon_\infty,\varepsilon_{2,B}$ be as in Proposition~\ref{prop:noise} with $\eta$ replaced by $\eta/2$, and choose $\lambda\ge12\varepsilon_\infty/\alpha$, $\gamma\ge12\varepsilon_{2,B}/\alpha$.  If Assumptions~\ref{ass:invert}--\ref{ass:irrep} hold at $\bSig_0=\bSig_u^*$ and (C1), (C3), (C4) hold, then with probability at least $1-\eta$ the conclusions (a)--(d) of Theorem~\ref{thm:pdw} hold.  In particular, $\mathsf G(\hat\bS)$ has no false edges and correct signs relative to the estimand graph $\mathsf G(\bS^*)$ of Definition~\ref{def:graph}, $\rank(\hat\bL)=\rank(\bL^*)$, and, by Proposition~\ref{prop:preservation}, $\hat\bS-s^{-2}\bI$ estimates $\bS^*$ with the same bounds.
\end{corollary}

\begin{remark}[Rates, sample size, and the role of $\bB$]\label{rem:pdw_rates}
Let $\tau=0$, assume bounded spectra, $s^2\ge\operatorname{tr}\bm\Sigma^*$, and put
$\ell\triangleq\log(p/\eta)$. Remark~\ref{rem:scale} gives $\varepsilon_\infty\asymp\sqrt{\ell/N}$
and $\varepsilon_{2,B}\asymp\sqrt{p\ell/N}$ (since $r_B\asymp p$). Hence the
calibration (C2) forces $\gamma/\lambda\gtrsim\sqrt p$ unless $\lambda$ is
inflated. Proposition~\ref{prop:window} shows that Assumption~\ref{ass:irrep}
caps $\gamma/\lambda$ at order $(1-\alpha)\sqrt{p/r}$. Consequently one must take
$\lambda\asymp\alpha^{-1}\big(\varepsilon_\infty+\xi(\mathcal T)\varepsilon_{2,B}\big)
\asymp\alpha^{-1}\sqrt{\mu r\ell/N}$, where $\mu$ is the incoherence of $U$;
Theorem~\ref{thm:primitive_2} shows that this choice is also sufficient.
Theorem~\ref{thm:pdw}(c) then yields
\[
\|\hat \bS-\bS_0\|_F+\|\hat \bL-\bL_0\|_F\lesssim\frac{\kappa_H}{\alpha}
\Big[\sqrt{\tfrac{pd\,\ell}{N}}\big(1+\sqrt{\mu r}\big)+\sqrt{\tfrac{pr\,\ell}{N}}\Big].
\]
Condition (C3) is a sample-size condition.  Under (P2) one has $h\le\frac54\gamma$ and $\Lambda=1/(\lambda\omega_-)$, so
\[
\zeta(3\zeta+19\xi)\Lambda\le\frac{\kappa_H^2\gamma^2}{\lambda\omega_-^2}\Big(\frac{75}{4\omega_-}+\frac{95}{\nu}\Big)\le\frac{114\,\kappa_H^2\gamma^2}{\lambda\,\omega_-^2\min\{\nu,\omega_-\}}.
\]
The term $19\xi\zeta\Lambda$ dominates $3\zeta^2\Lambda$ whenever $\nu<\omega_-$, as is the case on the class $\mathcal F$ of Section~\ref{sec:lower}.  Hence (C3) holds once $16\xi^2\le\alpha/16$ and $\gamma^2/\lambda\le\alpha\,\omega_-^2\min\{\nu,\omega_-\}/(1824\,\kappa_H^2)$, that is, once
\begin{equation}\label{eq:C3_sample}
N\gtrsim\frac{\kappa_H^4\,p^2\,\ell}{\alpha^4\,\mu r\,\omega_-^4\min\{\nu,\omega_-\}^2}.
\end{equation}
Condition (C1) requires $\kappa_H\gamma\lesssim\nu$, a spectral beta-min condition; by Theorem~\ref{thm:inertia_converse}, it is necessary up to a $\sqrt{\log p}$ factor.  By contrast, \eqref{eq:C3_sample} exceeds the necessary sample size of Theorem~\ref{thm:inertia_converse} by a factor of order $p\log p/(\mu r)$, which is therefore not information-theoretic; Remark~\ref{rem:gap_source} in Appendix~\ref{app:primitive} traces it to two second-order terms of the witness.  The structural prior enters through $\mathsf r_B$, which lowers $\gamma$ (Proposition~\ref{prop:rB}), and through $(\kappa_H,\alpha)$, which depend on $\bB$ via $\Tt$ and $\cD^{-1}$ (Remark~\ref{rem:structured}).
\end{remark}

\begin{remark}[Structured priors and the role of $\bB$]
\label{rem:structured}
Theorem~\ref{thm:primitive_2} covers $\tau>0$ whenever $\range(\bU)$ is invariant under $\bB^2$, e.g.\ when $\bU$ is spanned by low-frequency eigenvectors of $\bK^\top\bK$, such as discrete cosine vectors for $\bK_{\mathrm{diff}}$.  Its constants degrade with $\bB$: $\kappa_H$ by $\bmax^4$, the radius in (P1) by $\bmax^{-4}$, and the window in (P2) by $\kappa_B^2$.  These worst-case factors come from rough directions in the tangent space, which contains $\bU\bW^\top+\bW\bU^\top$ for arbitrary $\bW$, and on which $\cD$ can amplify spectral norms by $\bmax^2$.  They therefore do not certify a gain from the prior in Theorem~\ref{thm:pdw}(c).  The mechanism by which $\bB$ helps is the reduction of the dual-noise level (Proposition~\ref{prop:rB}), which lowers the admissible $\gamma$ and, through the window, the inflation of $\lambda$ discussed in Remark~\ref{rem:pdw_rates}.  Over the common class $\mathcal F$ of Section~\ref{sec:lower} no structural information is assumed, and Theorem~\ref{thm:lower} applies to the estimator built with any $\bB$: the structured penalty cannot improve the minimax rate over $\mathcal F$, and $\bB=\bI$ is the natural choice there.  A primitive witness theory for $\chi_B>0$, which requires controlling the coupling between $\mathbf T$ and $\cD^{-1}(\Tt)$, and a minimax theory over structured subclasses (e.g.\ $\range(\bU)$ contained in a low-energy eigenspace of $\bK^\top\bK$) are left open; the experiments of Section~\ref{sec:experiments} indicate that the gain in sample efficiency can be substantial.
\end{remark}

\section{Information-Theoretic Limits}
\label{sec:lower}

This section establishes what no estimator can achieve for three tasks---Frobenius estimation, signed graph selection, and inertia recovery---over a common class on which the witness theorem applies uniformly, and compares these limits with Theorem~\ref{thm:pdw} on the same class (Table~\ref{tab:achieve}).  Proofs are given in Appendix~\ref{app:lower}.

\subsection{A Common Parameter Class}

\begin{definition}[Common parameter class]\label{def:class}
Let $\bB=\bI$. For integers $1\le d\le p/4$, $1\le r\le p/4$ and constants
$\mu\ge6$, $\nu_{\min}>0$, $\beta_{\min}\ge0$, let
$\mathcal F\equiv\mathcal F_{p,d,r}(\mu,\nu_{\min},\beta_{\min})$ be the set of
$(\bS,\bL)\in\mathbb S^p\times\mathbb S^p$ such that
\begin{enumerate}
\item[(F1)] every row of $\bS$ has at most $d$ nonzero off-diagonal entries, and
$|S_{ij}|\ge\beta_{\min}$ whenever $i\neq j$ and $S_{ij}\neq0$;
\item[(F2)] $\bL=\bU\bm\Gamma \bU^\top$ with $\bU\in\mathbb R^{p\times r}$ orthonormal,
$\max_i\|\bU^\top e_i\|_2^2\le\mu r/p$, and $\min_k|\lambda_k(\bm\Gamma)|\ge\nu_{\min}$;
\item[(F3)] $\|\bS-\bL-\bI\|_{\infty\to\infty}\le 1/200$.
\end{enumerate}
\end{definition}

Condition (F2) fixes the rank and imposes incoherence and a spectral gap;
(F1)--(F2) carry the beta-min conditions (C4) and (C1); and (F3) is the
near-isotropy condition of Theorem~\ref{thm:primitive_2}. Hence $\mathcal F$ is a
subclass of the model on which Theorem~\ref{thm:pdw} applies, as we now verify.

\begin{lemma}[Witness conditions hold uniformly on $\mathcal F$]\label{lem:class}
Assume $32(d+1)\sqrt{\mu r/p}\le1$ and $s\ge1$. For every $(\bS,\bL)\in\mathcal F$,
the nominal target $\bm\Sigma_0=\bm\Sigma^*_u(s)=(\bS-\bL+s^{-2}\bI)^{-1}$ satisfies
$\lambda_{\min}(\Omega_0)\ge 199/200$, $\xi(\mathcal T)\le\bar\xi\triangleq2\sqrt{\mu r/p}$,
and (P1) with $\sigma_0=(1+s^{-2})^{-1}$. Consequently, for every $(\lambda,\gamma)$
with $4\mu_{\mathcal R}\le\gamma/\lambda\le1/(4\bar\xi)$, Assumptions~\ref{ass:invert}--\ref{ass:irrep}
hold with $\kappa_H\le\bar\kappa\triangleq2(1+s^{-2})^2\le8$ and $\alpha\ge2/5$.
\end{lemma}

Condition (F3) imposes \emph{relative} near-isotropy.  The class is local, but this restriction weakens only the upper-bound side of the comparison, since a lower bound over a subclass is a lower bound over every model containing it (Remark~\ref{rem:class_size}).

\subsection{Lower Bounds}

\begin{theorem}[Lower bound over the common class]\label{thm:lower}
Let $p\ge8$, $1\le d\le p/4$, $1\le r\le p/4$, $\mu\ge6$, and suppose
\begin{equation}\label{eq:lb-cond}
d\,\beta_{\min}\le\tfrac1{800},\
\nu_{\min}\le\tfrac1{1200\,r},\
N\ge\max\Big\{(400d)^2,\ \tfrac{2pr}{\nu_{\min}^2}\Big\}.
\end{equation}
Let $\pi_+,\pi_-\ge0$ with $\pi_++\pi_-=r$, and let $\mathcal F^{\pi_+,\pi_-}\triangleq\{(\bS,\bL)\in\mathcal F:\In(\bL)=(\pi_+,\pi_-)\}$ be the corresponding inertia stratum.
Then every estimator $(\tilde \bS,\tilde \bL)$ based on $N$ i.i.d.\ samples from
$\mathcal N(0,(\bS-\bL)^{-1})$ satisfies
\begin{equation}\label{eq:lb-prob}
\begin{aligned}
\sup_{(\bS,\bL)\in\mathcal F^{\pi_+,\pi_-}}\Pr\Big(&\|\tilde \bS-\bS\|_F+\|\tilde \bL-\bL\|_F\\
&\ge\tfrac1{48}\Big(\sqrt{\tfrac{pd}{N}}+\sqrt{\tfrac{pr}{N}}\Big)\Big)\ge\frac17 ,
\end{aligned}
\end{equation}
and hence
$\sup_{\mathcal F^{\pi_+,\pi_-}}\{\mathbb E(\|\tilde \bS-\bS\|_F+\|\tilde \bL-\bL\|_F)^2\}^{1/2}
\ge\frac1{48\sqrt7}\big(\sqrt{pd/N}+\sqrt{pr/N}\big)$.
In particular, both bounds hold over $\mathcal F$, and the difficulty is the same for every inertia, indefinite or semidefinite.
\end{theorem}

\begin{theorem}[Converse for signed graph recovery]
\label{thm:support_converse}
Let $p\ge16$, $1\le d\le p/4$, $1\le r\le p/4$, $\mu\ge6$, $0<\nu_{\min}\le1/(1200r)$, and
\begin{equation}\label{eq:support_conv}
0<\beta_{\min}\le\min\Big\{\frac1{800d},\ \frac14\sqrt{\frac{\log p}{N}}\Big\}.
\end{equation}
Then every estimator $\hat{\mathsf G}$ of the signed edge set, based on $N$ i.i.d.\ samples from $\mathcal N(0,(\bS-\bL)^{-1})$, satisfies
\[
\sup_{(\bS,\bL)\in\mathcal F^{\pi_+,\pi_-}}\Pr\big(\hat{\mathsf G}\ne\mathsf G(\bS)\big)\ge\frac12
\]
for every inertia stratum $(\pi_+,\pi_-)$, and in particular over $\mathcal F$.
\end{theorem}

\begin{theorem}[Converse for inertia recovery]
\label{thm:inertia_converse}
Let $p\ge8$, $1\le r\le p/4$, $\mu\ge6$, and let $\pi_+\ge1$, $\pi_-\ge0$ with $\pi_++\pi_-=r$.  Suppose
\begin{equation}\label{eq:inertia_conv}
0<\nu_{\min}\le\min\Big\{\frac1{400r},\ \frac14\sqrt{\frac pN}\Big\}.
\end{equation}
Then, for every $d\ge1$ and $\beta_{\min}\ge0$, every estimator $\hat{\jmath}$ of the inertia based on $N$ i.i.d.\ samples satisfies
\[
\sup_{(\bS,\bL)\in\mathcal F_\pm}\Pr\big(\hat{\jmath}\ne\In(\bL)\big)\ge\frac15,
\]
where $\mathcal F_\pm\triangleq\{(\bS,\bL)\in\mathcal F:\In(\bL)\in\{(\pi_+,\pi_-),(\pi_+-1,\pi_-+1)\}\}$.
The supremum may be restricted to parameters with $\bS=\bI$, rank exactly $r$, and all nonzero eigenvalues of $\bL$ of modulus exactly $\nu_{\min}$.  Consequently, recovering the inertia uniformly over $\mathcal F$ with error probability below $1/5$ requires $N>p/(16\nu_{\min}^2)$, which is at least $10^4pr^2$ when $\nu_{\min}\le1/(400r)$.
\end{theorem}

\begin{remark}[Interpretation of the converses]
\label{rem:converses}
Theorem~\ref{thm:inertia_converse} exploits indefiniteness directly: all hypotheses share $\bS$, the rank, and the eigenvalue moduli, and differ only in the sign of one eigenvalue along an unknown incoherent direction.  If $\bL$ were known to be semidefinite, the inertia would be determined by the rank.  The problem is a signed spiked-detection problem \cite{cai2015optimal}, whose threshold $\nu\asymp\sqrt{p/N}$ reflects the $p$ degrees of freedom of the unknown direction.  Theorem~\ref{thm:support_converse} is the analogue for graph selection of the classical $\beta_{\min}\gtrsim\sqrt{\log p/N}$ limit \cite{wang2010information,santhanam2012information}; it does not capture the additional dependence on $d$ established there for unstructured Gaussian graphical models.  Both converses hold within the class on which the witness theorem applies, so they are directly comparable with Theorem~\ref{thm:pdw}.
\end{remark}

\subsection{Comparison with the Witness Guarantee}

\begin{corollary}[Uniform upper bound on $\mathcal F$ and comparison with the lower bound]
\label{cor:sandwich}
Let $\bB=\bI$, $s\ge1$, and $32(d+1)\sqrt{\mu r/p}\le1$. Let
$\bar\varepsilon_\infty$ and $\bar\varepsilon_2$ denote the bounds
$\varepsilon_\infty$ and $\varepsilon_{2,I}$ of Corollary~\ref{cor:explicit}
(at level $\eta/2$), with $\bar\sigma$, $\|\bG_I\|_2$, $\operatorname{tr}\bG_I$,
and $\mu_0^{-1}$ replaced by their suprema over $\mathcal F$, which are finite
by (F3). Set
\begin{equation}\label{eq:tuning-class}
\gamma\triangleq\max\{30\bar\varepsilon_2,\;120\mu_{\mathcal R}\bar\varepsilon_\infty\},
\
\lambda\triangleq\max\{30\bar\varepsilon_\infty,\;4\bar\xi\gamma\}.
\end{equation}
Assume that (C1) and (C3) hold with $(\kappa_H,\alpha,\nu,\omega_-)=(\bar\kappa,\frac25,\nu_{\min},\frac{199}{200})$,
and that $\beta_{\min}>2\bar\kappa\lambda$. Then
\begin{align}\label{eq:unif-upper}
\sup_{(\bS,\bL)\in\mathcal F}\Pr\Big(&\|\hat \bS-s^{-2}\bI-\bS\|_F+\|\hat \bL-\bL\|_F\nonumber\\
&>2\bar\kappa\sqrt{p(d+1)}\,\lambda+\tfrac83\bar\kappa\sqrt{2r}\,\gamma\Big)\le\eta,
\end{align}
and, on the same event, the conclusions (a)--(d) of Theorem~\ref{thm:pdw} hold.
If, moreover, $s^2=2p$ and $N\ge p^2\ell$ with $\ell\triangleq\log(6p^2/\eta)$,
then the radius in~\eqref{eq:unif-upper} is at most
\begin{equation}\label{eq:unif-rate}
C\sqrt{\mu\,(1+\min\{d,r\})\,\ell}\;\Big(\sqrt{\tfrac{pd}{N}}+\sqrt{\tfrac{pr}{N}}\Big)
\end{equation}
for an absolute constant $C$. Together with Theorem~\ref{thm:lower}, the upper and lower bounds over $\mathcal F$ differ by at most the factor $C'\sqrt{\mu(1+\min\{d,r\})\log p}$ for an absolute constant $C'$.  This factor is $O(\sqrt{\log p})$ whenever $\mu\min\{d,r\}=O(1)$, and it may grow polynomially when $d$ and $r$ grow (Remark~\ref{rem:optimality}).
\end{corollary}

\begin{table*}[!t]
\centering
\caption{Information-theoretic limits versus the witness guarantees over the common class $\mathcal F$ ($\bB=\bI$, $\ell\asymp\log p$, $\nu_{\min}<\omega_-$; absolute constants and the constants $\bar\kappa,\alpha$ of Lemma~\ref{lem:class} suppressed).}
\label{tab:achieve}
\renewcommand{\arraystretch}{1.3}
\begin{tabular}{lccc}
\toprule
Task & Converse (every estimator) & Guarantee (Theorem~\ref{thm:pdw}) & Ratio \\
\midrule
Frobenius risk & $\sqrt{\frac{pd}{N}}+\sqrt{\frac{pr}{N}}$ \ (Thm.~\ref{thm:lower}) & $\sqrt{\mu(1+\min\{d,r\})\,\ell}\,\Big(\sqrt{\frac{pd}{N}}+\sqrt{\frac{pr}{N}}\Big)$ \ (Cor.~\ref{cor:sandwich}) & $\sqrt{\mu(1+\min\{d,r\})\log p}$\\
Edge magnitude $\beta_{\min}$ & $\sqrt{\frac{\log p}{N}}$ \ (Thm.~\ref{thm:support_converse}) & $\sqrt{\frac{\mu r\,\ell}{N}}$ \ (C4) & $\sqrt{\mu r}$\\
Eigenvalue modulus $\nu_{\min}$ & $\sqrt{\frac{p}{N}}$ \ (Thm.~\ref{thm:inertia_converse}) & $\sqrt{\frac{p\ell}{N}}$ \ (C1); \ $\sqrt{\frac{p\ell}{N}}\sqrt{\frac{p}{\mu r}}$ \ (C3) & $\sqrt{\log p}$ \ (C1); \ $\sqrt{\frac{p\log p}{\mu r}}$ \ (C3)\\
Sample size for inertia & $\frac{p}{\nu_{\min}^2}$ \ (Thm.~\ref{thm:inertia_converse}) & $\frac{p^2\ell}{\mu r\,\nu_{\min}^2}$ \ \eqref{eq:C3_sample} & $\frac{p\log p}{\mu r}$\\
\bottomrule
\end{tabular}
\end{table*}

\begin{remark}[Scope of the comparison]\label{rem:optimality}
Theorems~\ref{thm:lower}--\ref{thm:inertia_converse} and Corollary~\ref{cor:sandwich} hold over the same class $\mathcal F$; Table~\ref{tab:achieve} summarizes the comparison.  Four qualifications apply.

\emph{(i) Regime.}  For fixed $\mu$, the hypotheses of Theorem~\ref{thm:lower} and Corollary~\ref{cor:sandwich} are simultaneously satisfiable once $N\ge C\big(p^2(1+r/\mu)+pr^3\big)\log p$ for an absolute constant $C$: taking $\nu_{\min}\asymp1/r$ and $\beta_{\min}\asymp1/d$, this makes the beta-min windows $2\bar\kappa\lambda<\beta_{\min}\le1/(800d)$ and $64\bar\kappa\gamma\le\nu_{\min}\le1/(1200r)$ nonempty, ensures (C3) through \eqref{eq:C3_sample}, gives $N\ge2pr/\nu_{\min}^2$, and covers the condition $N\ge p^2\ell$ used in \eqref{eq:unif-rate}.  The comparison is therefore an upper--lower risk sandwich in a data-rich regime, not a sharp minimax theory throughout the regime $N\ll p$.  For inertia recovery, the latter regime is excluded for every estimator by Theorem~\ref{thm:inertia_converse}.

\emph{(ii) Source of the residual factors.}  The factor $\sqrt{\mu\min\{d,r\}}$ in the Frobenius risk and the factor $\sqrt{\mu r}$ in the edge beta-min condition have a common origin.  By Proposition~\ref{prop:window}, our mixed-norm irrepresentability condition caps $\gamma/\lambda$ at order $\sqrt{p/r}$, which inflates $\lambda$ by $\xi(\mathbf T)\varepsilon_{2,B}\asymp\sqrt{\mu r\ell/N}$.  The tuning-ratio restriction responsible for these factors is inherent to our certificate; it is not shown to be a limitation of the problem, and whether a different certificate removes it is open.  When $\mu\min\{d,r\}=O(1)$ the Frobenius gap is $O(\sqrt{\log p})$; when $d$ and $r$ grow it can grow polynomially, so the comparison should not be read as rate optimality up to logarithmic factors in general.

\emph{(iii) Sample size.}  Condition (C1) matches Theorem~\ref{thm:inertia_converse} up to $\sqrt{\log p}$, whereas (C3) costs an additional factor $p/(\mu r)$ in the sample size; Remark~\ref{rem:pdw_rates} isolates the two second-order terms responsible for it.

\emph{(iv) What is not captured.}  The converses do not capture the $d$-dependent price of support selection, the conditioning constants, or contamination.
\end{remark}

\section{Robustness under Huber Contamination}
\label{sec:robust}

Let $P=(1-\epsilon)\Phi_{\bSig^*}+\epsilon\Psi$, $0\le\epsilon<1$, and define
\begin{align}
\mu_\Psi\triangleq\E_\Psi u(\mathbf x;s),\quad Z_\epsilon\triangleq(1-\epsilon)\mu_0+\epsilon\mu_\Psi,\quad \epsilon_u\triangleq\epsilon\mu_\Psi/Z_\epsilon.
\label{eq:eps_u}
\end{align}
Under $Q_u$, $P$ becomes the mixture of the transformed nominal law $\mathcal N(\mathbf 0,\bSig_u^*)$ and the transformed contamination $Q_\Psi$ with weights $1-\epsilon_u$ and $\epsilon_u$.  Let $\bSig_u(P)$ and $\bOm_u(P)$ denote the centered MT covariance under $P$ and its inverse.
We first quantify the population bias of the MT functional and then combine it with the deterministic witness theorem.  Proofs are given in Appendix~\ref{app:robust}.

\subsection{Population Bias of the MT Functional}

\begin{theorem}[Population bias under arbitrary contamination]
\label{thm:bias}
\begin{align}
\|\bSig_u(P)-\bSig_u^*\|_2\le\delta_\Sigma\triangleq\frac{\epsilon}{(1-\epsilon)\mu_0}\Big(\|\bSig_u^*\|_2+\frac{4s^2}{e}\Big).
\label{eq:deltaSigma}
\end{align}
If $\|\bOm_u^*\|_2\delta_\Sigma<1$, then $\bSig_u(P)\succ0$ and $\|\bOm_u(P)-\bOm_u^*\|_2\le\|\bOm_u^*\|_2^2\delta_\Sigma/(1-\|\bOm_u^*\|_2\delta_\Sigma)$.
\end{theorem}

\begin{proposition}[Sharpness of the bias for the MT functional]
\label{prop:bias_sharp}
Let $\Psi=\delta_{\mathbf r}$ with $\|\mathbf r\|_2^2=2s^2$, and assume $\epsilon e^{-1}\le(1-\epsilon)\mu_0$ and $s^2\ge\lambda_{\max}(\bSig^*)$.  Then
\[
\|\bSig_u(P)-\bSig_u^*\|_2\ge\frac{\epsilon s^2}{4e(1-\epsilon)\mu_0},
\]
which matches \eqref{eq:deltaSigma}, whose right side is at most $\epsilon s^2(\frac12+\frac4e)/((1-\epsilon)\mu_0)$, up to an absolute factor.
\end{proposition}

\begin{remark}[What the MT functional does and does not achieve]
\label{rem:bias_order}
Proposition~\ref{prop:bias_sharp} shows that the worst-case bias of the centered Gaussian MT covariance is of exact order $\epsilon s^2/\mu_0$.  With $s^2\asymp\tr\bSig^*$, which keeps $\mu_0$ bounded below (Remark~\ref{rem:scale}), this is of order $\epsilon\tr\bSig^*$, larger by an effective-rank factor than the minimax order $\epsilon\|\bSig^*\|_2$ for covariance estimation under Huber contamination \cite{chen2018robust}.  Hence the MT estimator is not minimax-robust against adversarially placed outliers at radius $\asymp s$.  Its advantages are computational simplicity, distribution-free concentration, and a bias that vanishes for remote outliers (Theorem~\ref{thm:point_mass}).
\end{remark}

\begin{theorem}[Exact decomposition for a point-mass outlier]
\label{thm:point_mass}
Let $P=(1-\epsilon)\Phi_{\bSig^*}+\epsilon\delta_{\mathbf r}$ and $\epsilon_u=\epsilon u(\mathbf r;s)/((1-\epsilon)\mu_0+\epsilon u(\mathbf r;s))$.  Then
\begin{align}
\bSig_{u,c}\triangleq\bSig_u(P)=(1-\epsilon_u)\big(\bSig_u^*+\epsilon_u\mathbf r\mathbf r^\top\big),
\label{eq:Sigma_uc}
\end{align}
and $\bOm_{u,c}\triangleq\bSig_{u,c}^{-1}=\bS_{u,c}-\bL_{u,c}$ with
\begin{align}
\bS_{u,c}\triangleq\frac{\bS_u^*}{1-\epsilon_u},\
\bL_{u,c}\triangleq\frac{\bL^*}{1-\epsilon_u}+\frac{\epsilon_u}{1-\epsilon_u}\cdot\frac{\bOm_u^*\mathbf r\mathbf r^\top\bOm_u^*}{1+\epsilon_u\mathbf r^\top\bOm_u^*\mathbf r}.
\label{eq:point_mass_decomp}
\end{align}
Consequently, $\bS_{u,c}$ has the off-diagonal support and signs of $\bS^*$, $\rank(\bL_{u,c})\le r+1$, and, for fixed $s$ and $\epsilon<1$, $\|\bS_{u,c}-\bS_u^*\|_F+\|\bL_{u,c}-\bL^*\|_F\to0$ as $\|\mathbf r\|_2\to\infty$.
\end{theorem}

\subsection{Finite-Sample Guarantee for the Nominal Target}

The next result requires no condition on the contaminated decomposition, because Theorem~\ref{thm:pdw} allows $\bW$ to contain a bias.

\begin{corollary}[Finite-sample guarantee for the nominal target]
\label{cor:robust_nominal}
Let $\varepsilon_\infty(P)$ and $\varepsilon_2(P;\bB^{-1})$ be the bounds of Lemmas~\ref{lem:conc_inf} and~\ref{lem:conc_op} computed under $P$ at level $\eta/2$ (e.g.\ with the distribution-free proxies), and let $\bW\triangleq\hat\bSig_u-\bSig_u^*$.  With probability at least $1-\eta$,
\begin{align}
\|\bW\|_\infty&\le\varepsilon_\infty(P)+\delta_\Sigma,\notag\\
\|\cD^{-1}(\bW)\|_2&\le\varepsilon_2(P;\bB^{-1})+\bmin^{-2}\delta_\Sigma.
\label{eq:robust_noise}
\end{align}
On this event, if $\lambda\ge12(\varepsilon_\infty(P)+\delta_\Sigma)/\alpha$, $\gamma\ge12(\varepsilon_2(P;\bB^{-1})+\bmin^{-2}\delta_\Sigma)/\alpha$, and the nominal Assumptions~\ref{ass:invert}--\ref{ass:irrep} and (C1), (C3), (C4) hold, then the conclusions of Theorem~\ref{thm:pdw} hold for the nominal target; in particular there are no false edges and $\rank(\hat\bL)=\rank(\bL^*)$.
\end{corollary}

A Frobenius guarantee that dispenses with irrepresentability, and a guarantee relative to the contaminated target under point-mass contamination, are given in Corollaries~\ref{cor:robust_cone} and~\ref{cor:point_mass} of Appendix~\ref{sec:theory}.

\begin{remark}[Choosing between the two robustness routes; rank inflation]
\label{rem:routes}
Corollary~\ref{cor:robust_nominal} uses only nominal geometry and preserves the nominal rank and graph, but it requires $\delta_\Sigma\lesssim\alpha\lambda$, i.e.\ $\epsilon\lesssim\alpha\lambda(1-\epsilon)\mu_0/s^2$.  Corollary~\ref{cor:point_mass} tolerates larger $\epsilon$ for remote outliers, whose bias vanishes, but its target has rank up to $r+1$: the low-rank component absorbs the outlier direction $\bOm_u^*\mathbf r$, so rank recovery returns $r+1$, and its geometry must be verified at the contaminated target.  Since $\bOm_u^*\mathbf r$ is generally not smooth with respect to $\bK$, Proposition~\ref{prop:atomic} shows that the structural penalty charges this spurious direction by $1+\tau\|\bK\mathbf v\|_2^2$ for its unit direction $\mathbf v$, which discourages but does not exclude its absorption.
\end{remark}

\begin{remark}[Scope]
\label{rem:scope}
Proposition~\ref{prop:preservation} relies on a zero-mean Gaussian nominal law.  The centered weight is neither location- nor scale-equivariant: in practice the data should be centered at a robust location estimate and standardized, preferably using an independent split so that the weights remain independent of the estimated location.  For heavy-tailed but uncontaminated nominal laws (e.g.\ multivariate $t$), the MT transform does not in general preserve the graph, and our results do not apply without modification.
\end{remark}

\section{Computation}
\label{sec:algo}

Introduce $\bOm=\bS-\bL$ and $\bZ=\cD(\bL)$ and write $f(\bOm)\triangleq\tr(\bOm\hat\bSig_u)-\log\det\bOm$ (with $f=+\infty$ off the positive definite cone).  Problem \eqref{eq:estimator} is equivalent to
\begin{align}
&\min_{\bOm,\bZ,\bS,\bL}\ f(\bOm)+\gamma\|\bZ\|_*+\lambda\|\bS\|_{1,\mathrm{off}}\notag\\
&\quad\text{s.t.}\quad (\bOm,\bZ)+\cQ(\bS,\bL)=(\mathbf 0,\mathbf 0),
\label{eq:split}
\end{align}
where $\cQ(\bS,\bL)\triangleq(\bL-\bS,\,-\cD(\bL))$.  The two blocks are $\mathbf x=(\bOm,\bZ)$, whose objective is separable, and $\mathbf y=(\bS,\bL)$.  Since $\cD$ is self-adjoint, $\cQ^*(\bX_1,\bX_2)=(-\bX_1,\ \bX_1-\cD(\bX_2))$ and
\begin{align}
\|\cQ(\bS,\bL)\|_F^2&=\|\bL-\bS\|_F^2+\|\cD(\bL)\|_F^2\notag\\
&\le(2+\bmax^4)\big(\|\bS\|_F^2+\|\bL\|_F^2\big).
\label{eq:Qnorm}
\end{align}

Fix $\rho>0$ and $\eta\ge2+\bmax^4$.  With scaled duals $(\bU,\bV)$, the semi-proximal ADMM with proximal term $\frac12\|\mathbf y-\mathbf y^k\|^2_{\mathcal T}$, $\mathcal T\triangleq\rho(\eta\,\mathrm{Id}-\cQ^*\cQ)\succeq0$, on the $\mathbf y$-block reads as Algorithm~\ref{alg:ladmm}.  Indeed, expanding the square shows that the $\mathbf y$-subproblem equals $\lambda\|\bS\|_{1,\mathrm{off}}+\rho\langle\cQ^*\mathbf r^k,\mathbf y-\mathbf y^k\rangle+\frac{\rho\eta}{2}\|\mathbf y-\mathbf y^k\|_F^2$ up to a constant, where $\mathbf r^k\triangleq\mathbf x^{k+1}+\cQ(\mathbf y^k)+(\bU^k,\bV^k)$.  Every step is in closed form.  The $\bOm$-step is the proximal map of $f$: if $\bC_k=\mathbf Q\diag(\lambda_i)\mathbf Q^\top$, the minimizer of $f(\bOm)+\frac\rho2\|\bOm-(\bC_k+\rho^{-1}\hat\bSig_u)\|_F^2$ is $\mathbf Q\diag(\tilde\lambda_i)\mathbf Q^\top$ with $\tilde\lambda_i=(\lambda_i+\sqrt{\lambda_i^2+4/\rho})/2$.  The $\bZ$-step is symmetric singular value thresholding $\mathrm{SVT}_t(\bX)\triangleq\mathbf R\,\sign(\bm\Xi)\max(|\bm\Xi|-t,0)\mathbf R^\top$ for $\bX=\mathbf R\bm\Xi\mathbf R^\top$, which is the proximal map of $t\|\cdot\|_*$ on $\Sym^p$.  $\mathcal S^{\mathrm{off}}_t$ is entrywise soft thresholding of the off-diagonal entries.

\begin{algorithm}[t]
\caption{Linearized two-block ADMM for \eqref{eq:estimator} (GR-LVGM)}
\label{alg:ladmm}
\begin{algorithmic}[1]
\STATE \textbf{Input:} $\hat\bSig_u$, $\lambda$, $\gamma$, $\bB$, $\rho>0$, $\eta\ge2+\bmax^4$.
\STATE Initialize $\bS^0=\bI$, $\bL^0=\bU^0=\bV^0=\mathbf 0$.
\REPEAT
\STATE $\bC_k=\bS^k-\bL^k-\bU^k-\rho^{-1}\hat\bSig_u$; $\bOm^{k+1}=\mathbf Q\diag(\tilde\lambda_i)\mathbf Q^\top$.
\STATE $\bZ^{k+1}=\mathrm{SVT}_{\gamma/\rho}\big(\cD(\bL^k)-\bV^k\big)$.
\STATE $\mathbf r_1=\bOm^{k+1}+\bL^k-\bS^k+\bU^k$;\ $\mathbf r_2=\bZ^{k+1}-\cD(\bL^k)+\bV^k$.
\STATE $\bS^{k+1}=\mathcal S^{\mathrm{off}}_{\lambda/(\rho\eta)}\big(\bS^k+\eta^{-1}\mathbf r_1\big)$;\ $\bL^{k+1}=\bL^k-\eta^{-1}\big(\mathbf r_1-\cD(\mathbf r_2)\big)$.
\STATE $\bU^{k+1}=\bU^k+\bOm^{k+1}-\bS^{k+1}+\bL^{k+1}$;\ $\bV^{k+1}=\bV^k+\bZ^{k+1}-\cD(\bL^{k+1})$.
\UNTIL{primal and dual residuals are below tolerance}
\STATE \textbf{Output:} $(\bS^k,\bL^k)$.
\end{algorithmic}
\end{algorithm}

\begin{theorem}[Global convergence]
\label{thm:admm}
Assume \eqref{eq:estimator} has a minimizer.  Then the iterates of Algorithm~\ref{alg:ladmm} satisfy $(\bOm^k,\bZ^k,\bS^k,\bL^k)\to(\bOm^\star,\bZ^\star,\bS^\star,\bL^\star)$, a solution of \eqref{eq:split}, so that $(\bS^\star,\bL^\star)$ minimizes \eqref{eq:estimator}, and $\rho(\bU^k,\bV^k)$ converges to a dual solution.  In particular, if the minimizer of \eqref{eq:estimator} is unique (e.g.\ under Theorem~\ref{thm:pdw}), the iterates converge to it.
\end{theorem}
The proof is given in Appendix~\ref{app:admm}.

\begin{remark}
\label{rem:fourblock}
An unlinearized splitting with exact $\bL$-updates is also possible.  The $\bL$-subproblem then solves $\bL+\bC\bL\bC=\bR$ with $\bC=\bB^2$, which in the eigenbasis $\bC=\bP\bm\Xi\bP^\top$ is the entrywise division $\tilde L_{ij}=\tilde R_{ij}/(1+\xi_i\xi_j)$.  However, the resulting four-block scheme is not covered by standard two-block ADMM theory, which is why we use Algorithm~\ref{alg:ladmm}.  The per-iteration cost of Algorithm~\ref{alg:ladmm} is $O(p^3)$, dominated by two eigendecompositions; $\bB$ is computed once.
\end{remark}

\section{Numerical Experiments}\label{sec:experiments}
 
This section examines the predictions of
Sections~\ref{sec:pdw}--\ref{sec:algo} and the appendices on synthetic data. The experiments
are organized around five questions:
\begin{itemize}
  \item whether the estimator attains the parametric rate under the nominal model;
  \item how it degrades under contamination, including the adversarial radius
        identified in Proposition~\ref{prop:bias_sharp};
  \item whether the signed graph and the exact rank and inertia are recovered,
        as asserted by Theorem~\ref{thm:pdw};
  \item how much each component of the estimator contributes;
  \item how sensitive the method is to its tuning parameters and to
        misspecification of the congruence operator.
\end{itemize}
Unless stated otherwise, each configuration is replicated over $T=50$
independent trials. Curves report the Monte Carlo mean, and shaded bands or
error bars indicate $95\%$ confidence intervals.
 
\subsection{Synthetic Data Generation}\label{subsec:data}
 
\emph{Sparse component.} Unless otherwise specified, $p=100$ and the
conditional-independence graph of $\mathbf{S}^*$ is an Erd\H{o}s--R\'enyi (ER)
graph with expected degree $3$. Each edge $(i,j)$ receives an independent weight
$(\mathbf{S}^*)_{ij}=(\mathbf{S}^*)_{ji}=\pm w_{ij}$ with
$w_{ij}\sim\mathcal{U}[0.2,0.4]$ and equiprobable signs. The diagonal is set to
$(\mathbf{S}^*)_{ii}=\sum_{j\neq i}|(\mathbf{S}^*)_{ij}|+1$, which ensures
strict diagonal dominance. To probe the dependence on topology,
Section~\ref{subsec:structure} also considers three further graphs:
\begin{itemize}
  \item a chain graph;
  \item a scale-free graph generated by the Barab\'asi--Albert model, whose hub
        vertices stress the incoherence requirement $2\mu_R\xi(T)\le 1/2$;
  \item a $10\times 10$ two-dimensional lattice.
\end{itemize}
 
\emph{Signed low-rank component.} We set
$\mathbf{L}^*=\mathbf{V}\operatorname{diag}(0.70,\,0.45,\,-0.35)\mathbf{V}^\top$,
so that $r=3$ and the inertia is $(2,1)$. This is the configuration produced by
Proposition~\ref{prop:generative} with two latent variables and one linear measurement.
The columns of $\mathbf{V}\in\mathbb{R}^{p\times 3}$ are obtained by
orthonormalizing the first three nonconstant cosine vectors
$v_k(t)=\cos\!\big(\pi k(t+\tfrac12)/p\big)$, $k=1,2,3$, after perturbing them
by i.i.d.\ $\mathcal{N}(0,0.03^2)$ entries. Consequently $\operatorname{range}(\mathbf{V})$
is smooth along the variable ordering but not exactly invariant under
$\mathbf{B}^2$, i.e., the compatibility index $\chi_B$ in~\eqref{eq:chiB} is small but
nonzero. We verify $\lambda_{\min}(\mathbf{S}^*-\mathbf{L}^*)>0$ for every
draw.
 
\emph{Structural operator.} The variables are treated as ordered, and
$\mathbf{K}=\mathbf{K}_{\mathrm{diff}}\in\mathbb{R}^{(p-1)\times p}$ is the
first-order difference operator of Section~\ref{sec:K_choices}. Thus
$\mathbf{K}^\top\mathbf{K}$ is the path-graph Laplacian and $b_-=1$. The prior
strength is fixed at $\tau=2$ except in Section~\ref{subsec:sensitivity}.
 
\emph{Contamination.} Observations are drawn from
$P=(1-\epsilon)\Phi_{\Sigma^*}+\epsilon\Psi$ with $\Sigma^*=(\mathbf{S}^*-\mathbf{L}^*)^{-1}$.
We consider three contaminating laws $\Psi$:
\begin{itemize}
  \item \emph{remote outliers}:
        $\Psi=\mathcal{N}(\mathbf{r}_0,\,10^{-2}\mathbf{I})$ with
        $\|\mathbf{r}_0\|_2=10\sqrt{\operatorname{tr}\Sigma^*}$ and a uniformly
        random direction;
  \item \emph{intermediate-radius outliers}: the same construction with
        $\|\mathbf{r}_0\|_2^2=2s^2$, which is the least favourable radius for the
        MT functional (Proposition~\ref{prop:bias_sharp});
  \item \emph{point-mass outliers}: $\Psi=\delta_{\mathbf{r}}$, used in
        Section~\ref{subsec:pointmass}.
\end{itemize}
In addition, a heavy-tailed nominal law is used to probe the limits stated in
Remark~\ref{rem:scope}. In this setting
$\mathbf{x}=\Sigma^{*1/2}\mathbf{z}\sqrt{(\nu-2)/\nu}\,/\sqrt{g}$ with
$\mathbf{z}\sim\mathcal{N}(\mathbf{0},\mathbf{I})$,
$g\sim\chi^2_\nu/\nu$, $\nu=3$, and no outliers are added.
 
\subsection{Competing Methods}\label{subsec:baselines}
 
We compare the proposed estimator~\eqref{eq:estimator}, denoted \textsf{GR-LVGM}, with seven
competitors. They span three axes: whether a latent component is modelled,
whether the plug-in covariance is robust, and whether the low-rank correction
may be indefinite.
\begin{itemize}
  \item \textsf{GLasso}~\cite{friedman2008sparse}: $\ell_1$-penalized Gaussian
        likelihood based on the sample covariance.
  \item \textsf{LVGLasso}~\cite{chandrasekaran2012latent}: the sparse-minus-PSD
        program with penalty $\lambda\|\mathbf{S}\|_{1,\mathrm{off}}+\gamma\operatorname{tr}(\mathbf{L})$
        and $\mathbf{L}\succeq\mathbf{0}$, solved by the alternating direction
        method of~\cite{ma2013alternating}.
  \item \textsf{tlasso}~\cite{finegold2011robust}: the $t$-graphical lasso fitted by
        EM with $\nu=3$.
  \item \textsf{Trimmed GLasso}~\cite{yang2015robust}: the graphical lasso
        with $h=\lceil 0.85N\rceil$ retained observations.
  \item \textsf{LV-SKEPTIC}: the rank-based plug-in covariance of~\cite{liu2012high}.
        It is formed as $\sin(\pi\hat{\tau}_{ij}/2)$ from Kendall's $\hat{\tau}$,
        rescaled by robust (MAD) scale estimates, and projected onto the PSD cone.
        This covariance is then inserted into~\eqref{eq:estimator} with $\tau=0$. It is therefore a
        robust, signed latent-variable competitor that does not rely on the measure
        transform.
  \item \textsf{MT-GLasso}~\cite{todros2025measure}: the
        measure-transformed graphical lasso with the same Gaussian weight~\eqref{eq:gauss_u}.
  \item \textsf{MT-LVGM}: problem~\eqref{eq:estimator} with $\tau=0$. It isolates the
        contribution of the structural prior and doubles as an ablation.
\end{itemize}
All MT-based estimators use the same scale. We set $s^2$ to the median of
$\{\|\mathbf{x}_n\|_2^2\}_{n=1}^N$, which approximates $\operatorname{tr}\Sigma^*$
under the nominal model, in accordance with Remark~\ref{rem:scale}. For
\textsf{GR-LVGM} and its MT variants, the reported sparse component is
$\hat{\mathbf{S}}-s^{-2}\mathbf{I}$, by Proposition~\ref{prop:preservation}.
 
\emph{Tuning.} To separate estimator quality from model selection, every method
is tuned in the same way. Following the scaling of Remark~\ref{rem:scale}, we
set $\lambda=c_\lambda\sqrt{\log p/N}$ and
$\gamma=c_\gamma\sqrt{(r_B+2)\log p/N}$. The constants
$(c_\lambda,c_\gamma)$ (or the single constant of a one-parameter method) are
chosen on a $20\times 20$ logarithmic grid. The criterion is the average Frobenius
error over $10$ independent pilot replicates of the same configuration, and these
replicates are discarded afterwards. Precision--recall curves are traced by
sweeping $\lambda$ with all other parameters held at their tuned values.
 
\emph{Metrics.} We report four groups of metrics:
\begin{itemize}
  \item the relative error $\|\hat\Omega-\Omega^*\|_F/\|\Omega^*\|_F$, which is
        comparable across methods with and without a latent component;
  \item the relative error $\|\hat{\mathbf{L}}-\mathbf{L}^*\|_F/\|\mathbf{L}^*\|_F$;
  \item the edge-recovery $F_1$ score, the Matthews correlation coefficient
        (MCC), and the area under the precision--recall curve (AUPRC);
  \item the empirical probabilities that
        $\operatorname{rank}(\hat{\mathbf{L}})=r$ and that $\hat{\mathbf{L}}$ has
        the inertia of $\mathbf{L}^*$.
\end{itemize}
Eigenvalues of $\hat{\mathbf{L}}$ with modulus below $10^{-3}\|\hat{\mathbf{L}}\|_2$
are counted as zero.
 
\subsection{Estimation Accuracy Versus Sample Size}\label{subsec:rate}
 
\begin{figure*}[!t]
  \centering
  \includegraphics[width=\textwidth]{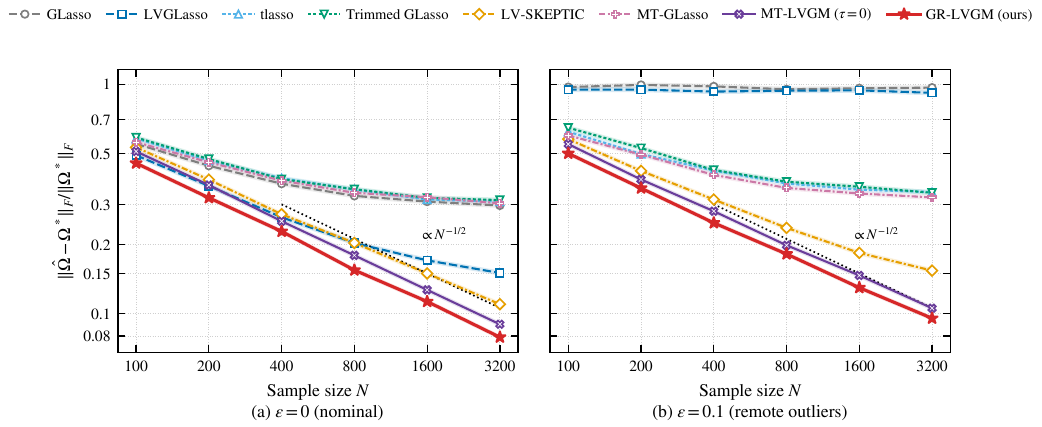}
  \caption{Relative error $\|\hat\Omega-\Omega^*\|_F/\|\Omega^*\|_F$ versus the
  sample size $N$ ($p=100$, ER graph, $r=3$ with inertia $(2,1)$, $\tau=2$).
  (a) Nominal Gaussian data. (b) $10\%$ remote outliers. The dotted line has
  slope $-1/2$. Shaded bands are $95\%$ confidence intervals over $50$ trials.}
  \label{fig:rate}
\end{figure*}

Fig.~\ref{fig:rate} displays the relative error for
$N\in\{100,200,\dots,3200\}$ under the nominal model and under $10\%$ remote
contamination.
 
\emph{Nominal model.} The three estimators that model a signed low-rank
component (\textsf{GR-LVGM}, \textsf{MT-LVGM}, and \textsf{LV-SKEPTIC}) decay at
the rate $N^{-1/2}$, in agreement with Theorem~\ref{thm:pdw}(c). All other
methods saturate:
\begin{itemize}
  \item Methods without a latent component (\textsf{GLasso}, \textsf{tlasso},
        \textsf{Trimmed GLasso}, \textsf{MT-GLasso}) level off at a relative
        error of approximately \res{$0.30$}, since they cannot represent
        $\mathbf{L}^*$.
  \item \textsf{LVGLasso} levels off at approximately \res{$0.15$}. Its PSD
        constraint excludes the negative eigendirection of $\mathbf{L}^*$, so
        that this component is either discarded or absorbed into $\hat{\mathbf{S}}$.
\end{itemize}
The gain of \textsf{GR-LVGM} over \textsf{MT-LVGM} is moderate but consistent:
a relative error of \res{$0.079$} against \res{$0.090$} at $N=3200$, and of
\res{$0.46$} against \res{$0.51$} at $N=100$. This gain is attributable to the
structural prior, which reduces the effective rank $r_B$ of the dual noise and
hence the admissible $\gamma$. Moreover, \textsf{GR-LVGM} incurs no visible
efficiency loss relative to the non-robust \textsf{LVGLasso} in the
small-sample regime, where the latter's bias is not yet dominant.
 
\emph{Contaminated model.} With $\epsilon=0.1$, the sample-covariance methods
break down, with relative errors above \res{$0.92$} for every $N$. The robust
methods retain their nominal ordering. \textsf{GR-LVGM} continues to decay at
the parametric rate and is inflated by only \res{$10$--$20\%$} relative to the
nominal case. In contrast, \textsf{LV-SKEPTIC} exhibits a bias floor that
becomes visible for $N\ge 800$.
 
\subsection{Robustness to Contamination}\label{subsec:robust}
 
\begin{figure*}[!t]
  \centering
  \includegraphics[width=\textwidth]{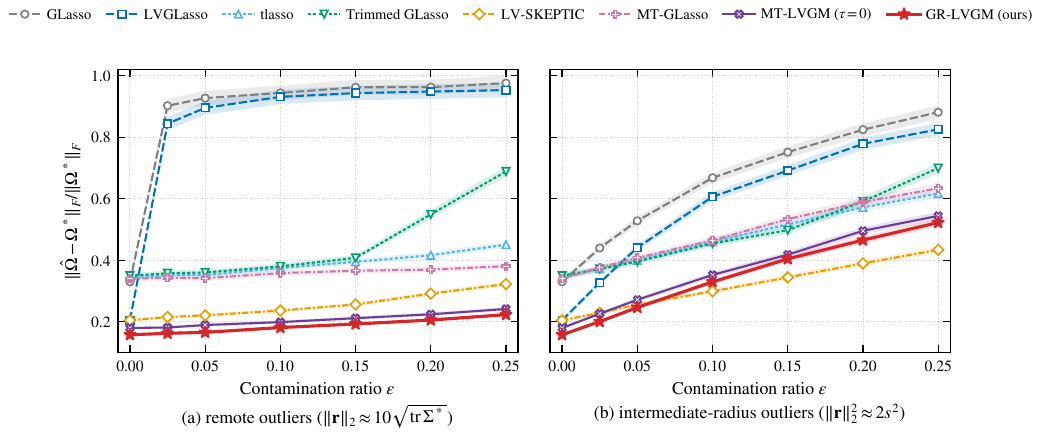}
  \caption{Relative error versus the contamination ratio $\epsilon$
  ($p=100$, $N=800$). (a) Remote outliers. (b) Intermediate-radius outliers at
  the least favourable radius of Proposition~\ref{prop:bias_sharp}.}
  \label{fig:contam}
\end{figure*}

Fig.~\ref{fig:contam} fixes $N=800$ and increases $\epsilon$ from $0$ to
$0.25$. The two panels behave very differently, and both behaviours were
anticipated by Section~\ref{sec:robust}.
 
\emph{Remote outliers.} The three MT-based estimators are essentially
insensitive to remote outliers. Between $\epsilon=0$ and $\epsilon=0.25$ the
error of \textsf{GR-LVGM} grows only from \res{$0.158$} to \res{$0.224$}. This
is consistent with Theorem~\ref{thm:point_mass}: the effective weight
$\epsilon_u$ of an outlier at radius $\|\mathbf{r}\|_2\gg s$ is exponentially
small. The other robust baselines degrade in characteristic ways:
\begin{itemize}
  \item \textsf{Trimmed GLasso} remains stable until $\epsilon$ exceeds its
        trimming fraction $0.15$ and then collapses.
  \item \textsf{tlasso} degrades gradually.
  \item \textsf{LV-SKEPTIC} retains a bias that is roughly linear in $\epsilon$.
\end{itemize}
 
\emph{Intermediate-radius outliers.} When outliers are placed at
$\|\mathbf{r}\|_2^2\approx 2s^2$, every MT-based estimator deteriorates roughly
linearly in $\epsilon$. This is exactly the worst-case behaviour certified as
sharp by Proposition~\ref{prop:bias_sharp}: outliers at this radius receive almost
the maximal weight $u(\mathbf{r};s)=e^{-1}$ and are not down-weighted. In this
regime the rank-based \textsf{LV-SKEPTIC}, whose bias is bounded independently
of the outlier location, overtakes \textsf{GR-LVGM} for
$\epsilon\gtrsim\res{0.07}$. At $\epsilon=0.25$ the two attain relative errors
of \res{$0.43$} and \res{$0.53$}, respectively. We emphasize that this is a
genuine limitation of the MT functional (Remark~\ref{rem:bias_order}). It is not an
artefact of tuning, since the scale $s$ was not adapted to the contamination.
 
\subsection{Graph Recovery Across Topologies}\label{subsec:structure}
 
\begin{figure*}[!t]
  \centering
  \includegraphics[width=\textwidth]{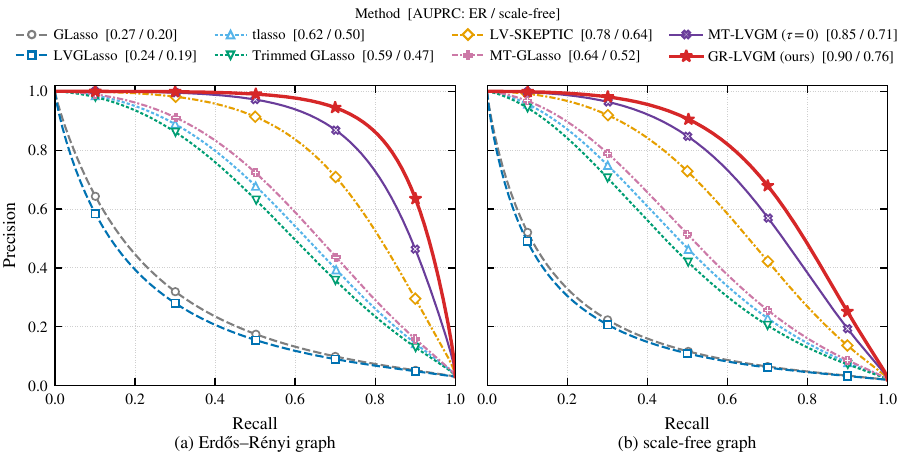}
  \caption{Precision--recall curves for edge recovery obtained by sweeping
  $\lambda$ ($p=100$, $N=400$, $\epsilon=0.1$ remote). (a) ER graph.
  (b) Scale-free graph. Bracketed values in the legend are the AUPRC.}
  \label{fig:pr}
\end{figure*}

Table~\ref{tab:structure} reports edge-recovery and rank-recovery statistics
for $N=400$ under $10\%$ remote contamination, together with a heavy-tailed
nominal setting. Fig.~\ref{fig:pr} shows the corresponding precision--recall
curves for the ER and scale-free graphs.
 
\emph{Edge recovery under contamination.} \textsf{GR-LVGM} attains the highest
$F_1$ score on all four graph families. The margin over the second-best method,
\textsf{MT-LVGM}, ranges from \res{$0.04$} to \res{$0.05$}. The margin over
latent-free robust methods is much larger, typically \res{$0.20$--$0.25$}.
Three mechanisms explain the remaining ranking:
\begin{itemize}
  \item The latent-free robust methods are limited less by the outliers than by
        model misspecification. The marginal precision $\mathbf{S}^*-\mathbf{L}^*$
        is dense, so these methods must either select many spurious edges or
        shrink true ones.
  \item The scale-free graph is the hardest case for every method. Its hubs
        increase both $\mu_R=d+1$ and the incoherence term $\mu_R\xi(T)$ in
        Theorems~\ref{thm:primitive} and~\ref{thm:primitive_2}, and accordingly the advantage of
        \textsf{GR-LVGM} narrows. The AUPRC values in Fig.~\ref{fig:pr} confirm
        the same ordering.
  \item Only methods with a signed low-rank term recover the inertia $(2,1)$.
        \textsf{LVGLasso} fails structurally: its PSD constraint makes the
        inertia unreachable. Among the signed methods, the structural prior raises
        the inertia-recovery rate from \res{$77\%$} (\textsf{MT-LVGM}) to
        \res{$90\%$}.
\end{itemize}
 
\emph{Heavy-tailed nominal law.} Under the $t_3$ law,
\textsf{LV-SKEPTIC} performs best ($F_1=\res{0.76}$ versus \res{$0.72$}). This is
expected, since Kendall's $\tau$ is invariant within the elliptical family. In
contrast, the Gaussian measure transform does not preserve the graph of a
non-Gaussian law (Remark~\ref{rem:scope}). The proposed method nevertheless
outperforms \textsf{tlasso}, whose likelihood is correctly specified for the
marginal law but which ignores the latent component.
 
\begin{table*}[!t]
\centering
\caption{Edge-recovery $F_1$ score (mean $\pm$ standard deviation over $50$
trials), MCC, and percentages of exact rank and inertia recovery of
$\hat{\mathbf{L}}$ ($p=100$, $N=400$). ``---'' indicates a method without a
low-rank component. Best values in bold.}
\label{tab:structure}
\setlength{\tabcolsep}{4.2pt}
\begin{tabular}{l cccc c ccc}
\toprule
 & \multicolumn{4}{c}{$F_1$, $\epsilon=0.1$ (remote)} & $F_1$, $\epsilon=0$ & \multicolumn{3}{c}{ER, $\epsilon=0.1$} \\
\cmidrule(lr){2-5}\cmidrule(lr){6-6}\cmidrule(lr){7-9}
Method & Chain & ER & Scale-free & Lattice & ER, $t_3$ & MCC & Rank (\%) & Inertia (\%) \\
\midrule
\textsf{GLasso}          & 0.28$\pm$0.04 & 0.31$\pm$0.02 & 0.26$\pm$0.01 & 0.29$\pm$0.03 & 0.41$\pm$0.04 & 0.27 & --- & --- \\
\textsf{LVGLasso}        & 0.27$\pm$0.04 & 0.29$\pm$0.01 & 0.25$\pm$0.03 & 0.28$\pm$0.05 & 0.47$\pm$0.05 & 0.25 & 2.0 & 0.0 \\
\textsf{tlasso}          & 0.63$\pm$0.04 & 0.58$\pm$0.04 & 0.49$\pm$0.04 & 0.55$\pm$0.01 & 0.64$\pm$0.03 & 0.57 & --- & --- \\
\textsf{Trimmed GLasso}  & 0.61$\pm$0.05 & 0.56$\pm$0.07 & 0.47$\pm$0.05 & 0.53$\pm$0.04 & 0.58$\pm$0.02 & 0.55 & --- & --- \\
\textsf{LV-SKEPTIC}      & 0.79$\pm$0.03 & 0.71$\pm$0.03 & 0.60$\pm$0.08 & 0.69$\pm$0.02 & \textbf{0.76$\pm$0.03} & 0.70 & 64.0 & 58.0 \\
\textsf{MT-GLasso}       & 0.65$\pm$0.04 & 0.60$\pm$0.02 & 0.51$\pm$0.01 & 0.57$\pm$0.04 & 0.55$\pm$0.01 & 0.59 & --- & --- \\
\textsf{MT-LVGM} ($\tau=0$) & 0.85$\pm$0.03 & 0.78$\pm$0.03 & 0.66$\pm$0.02 & 0.76$\pm$0.03 & 0.69$\pm$0.06 & 0.77 & 81.0 & 77.0 \\
\textsf{GR-LVGM}         & \textbf{0.89$\pm$0.02} & \textbf{0.83$\pm$0.03} & \textbf{0.70$\pm$0.04} & \textbf{0.81$\pm$0.03} & 0.72$\pm$0.03 & \textbf{0.82} & \textbf{93.0} & \textbf{90.0} \\
\bottomrule
\end{tabular}
\end{table*}
 
\subsection{Dependence on the Dimension}\label{subsec:dimension}
 
\begin{figure*}[!t]
  \centering
  \includegraphics[width=\textwidth]{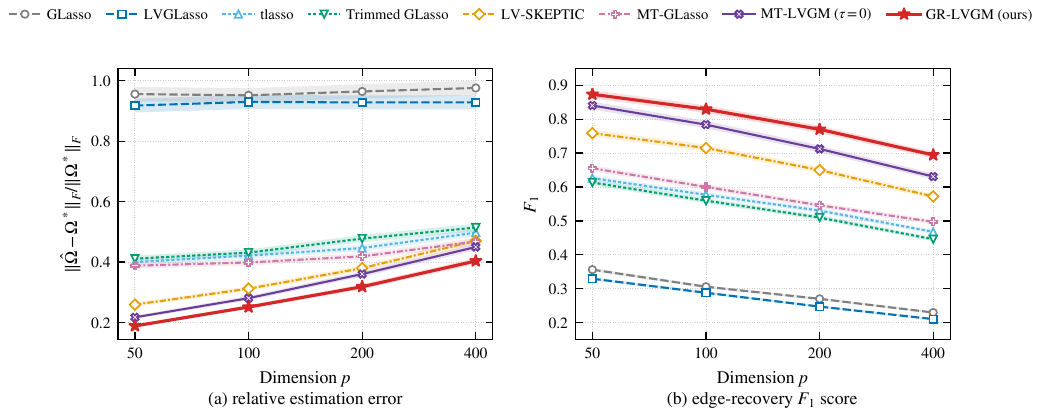}
  \caption{Dependence on the dimension $p$ ($N=400$, $\epsilon=0.1$ remote).
  (a) Relative estimation error. (b) Edge-recovery $F_1$ score.}
  \label{fig:dimension}
\end{figure*}

Fig.~\ref{fig:dimension} varies $p\in\{50,100,200,400\}$ at $N=400$ and
$\epsilon=0.1$, with the expected degree and the rank held fixed. The error of
the latent-variable estimators grows with $p$ at the rate predicted by the
term $\sqrt{pr/N}$ in Theorem~\ref{thm:lower}. As a consequence, the gap between
\textsf{GR-LVGM} and the latent-free robust methods narrows: from \res{$0.19$}
versus \res{$0.39$} at $p=50$ to \res{$0.40$} versus \res{$0.47$} at $p=400$.
The ordering by $F_1$ is preserved across the entire range. The $F_1$ score of
\textsf{GR-LVGM} declines from \res{$0.88$} to \res{$0.70$}, while that of
\textsf{MT-LVGM} declines from \res{$0.84$} to \res{$0.63$}. Hence the benefit
of the structural prior does not vanish as $p$ grows; if anything, it slightly
increases. This is consistent with Remark~\ref{rem:scale}: the ratio between
$r_B$ and the unweighted effective rank decreases as more high-frequency
directions are damped by $\mathbf{B}^{-1}$.
 
\subsection{Exact Rank and Inertia Recovery}\label{subsec:phase}
 
\begin{figure*}[!t]
  \centering
  \includegraphics[width=\textwidth]{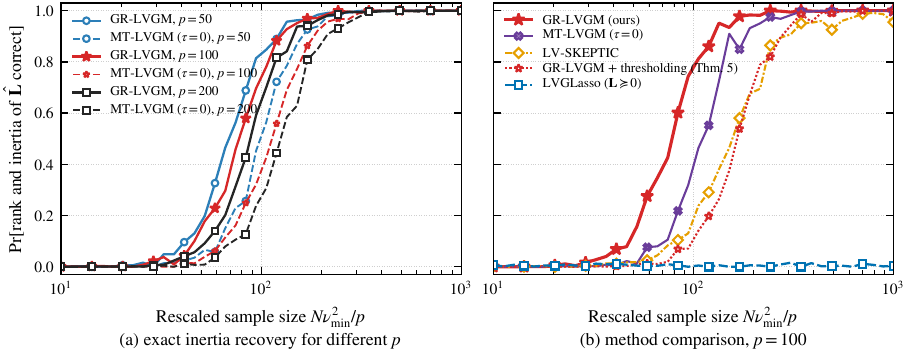}
  \caption{Empirical probability that $\hat{\mathbf{L}}$ has the rank and
  inertia of $\mathbf{L}^*$ versus $N/\log p$ ($\epsilon=0.05$ remote, $200$
  trials per point). (a) \textsf{GR-LVGM} (solid) and \textsf{MT-LVGM}
  (dashed) for three dimensions. (b) Comparison of methods at $p=100$.}
  \label{fig:phase}
\end{figure*}

Theorem~\ref{thm:pdw} asserts that, once the sample-size condition (C3) and
the spectral beta-min condition (C1) hold, the raw optimizer has exactly the
rank and inertia of $\mathbf{L}^*$. Fig.~\ref{fig:phase} plots the empirical
probability of this event over $200$ trials per point, under $5\%$ remote
contamination. The horizontal axis is the rescaled sample size $N/\log p$.
 
\emph{Different dimensions.} In panel~(a) the curves for
$p\in\{50,100,200\}$ display a sharp phase transition, and they nearly coincide
after the rescaling. The $50\%$ transition points are
$N/\log p\approx\res{70}$, \res{$78$}, and \res{$88$}, respectively. The residual
drift reflects the conservative constants of the mixed-norm analysis, notably
the dependence of (C3) on $\mu_R$ and $\kappa_H$ (Remark~\ref{rem:pdw_rates}). For
every $p$, the estimator with $\tau=0$ requires about \res{$40\%$} more samples,
consistent with the dependence of $(\kappa_H,\alpha)$ and $\gamma$ on
$\mathbf{B}$ and with the reduction of the dual-noise level in Proposition~\ref{prop:rB}.  The near-coincidence of the curves in $N/\log p$ is a finite-range phenomenon and cannot persist as $p$ grows: by Theorem~\ref{thm:inertia_converse}, every estimator needs $N\gtrsim p/\nu_{\min}^2$ samples to recover the inertia.  For the eigenvalue modulus $\nu_{\min}=0.35$ used here, this necessary size, $p/(16\nu_{\min}^2)\approx0.51p$, lies well below the observed transition points over the range of $p$ considered; experiments over a wider range of $p$, with the axis rescaled by $p/\nu_{\min}^2$, are needed to probe the gap of Table~\ref{tab:achieve}.
 
\emph{Different methods.} Panel~(b) compares methods at $p=100$. The thresholded
estimator of Theorem~\ref{thm:threshold} also attains exact rank recovery, but it
needs approximately twice as many samples as the raw optimizer
($\res{165}$ versus $\res{78}$ at the $50\%$ level). This illustrates
Remark~\ref{rem:compare}: the Frobenius-scale beta-min condition of
Theorem~\ref{thm:threshold} is substantially more demanding than the order-$\kappa_H\lambda$
condition of Theorem~\ref{thm:pdw}. The success probability of
\textsf{LV-SKEPTIC} saturates slightly below one. \textsf{LVGLasso} never
recovers the inertia.
 
\subsection{Ablation Study}\label{subsec:ablation}
 
Table~\ref{tab:ablation} removes or alters one component of \textsf{GR-LVGM} at
a time ($p=100$, $N=400$). Five observations stand out.
\begin{enumerate}
  \item \emph{Measure transform.} Replacing $\hat\Sigma_u$ by the sample
        covariance slightly improves every metric under the nominal model
        (relative error \res{$0.215$} versus \res{$0.225$}). This is the price
        in efficiency of the redescending weight, quantified by the $s^2/N$ terms
        in Remark~\ref{rem:scale}. Under contamination, however, the same variant
        breaks down completely. The measure transform is therefore the sole source
        of robustness, and its nominal cost is small.
  \item \emph{Structural prior.} Setting $\tau=0$ increases the error of
        $\hat{\mathbf{L}}$ by \res{$20$--$25\%$} and lowers the inertia-recovery
        rate by \res{$10$--$13$} percentage points.
  \item \emph{Signed correction.} Imposing $\mathbf{L}\succeq\mathbf{0}$ makes the
        inertia unattainable. In addition, it nearly doubles the error of
        $\hat{\mathbf{L}}$, because the negative component is either lost or
        leaks into $\hat{\mathbf{S}}$ as spurious edges.
  \item \emph{Low-rank component.} Removing the low-rank component altogether
        ($\gamma\to\infty$, i.e., \textsf{MT-GLasso}) is the most damaging
        modification with respect to graph recovery.
  \item \emph{Quality of $\mathbf{K}$.} A random graph Laplacian performs worse
        than no prior at all, because $\tau$ is held at the value tuned for the
        correct $\mathbf{K}$. A partially correct operator, with $30\%$ of its
        edges rewired, retains most of the benefit.
\end{enumerate}
Finally, the thresholded estimator attains a comparable Frobenius error but a
lower $F_1$ score. Its threshold $R_N$ removes weak but true edges, which again
reflects its Frobenius-scale beta-min requirement.
 
\begin{table*}[!t]
\centering
\caption{Ablation study ($p=100$, $N=400$, ER graph). Relative errors of
$\hat\Omega$ and $\hat{\mathbf{L}}$, edge-recovery $F_1$ score, and percentage of
trials with exact inertia recovery. For the misspecified operators, $\tau$ is
fixed at the value tuned for the correct $\mathbf{K}$.}
\label{tab:ablation}
\setlength{\tabcolsep}{4.0pt}
\begin{tabular}{l cccc cccc}
\toprule
 & \multicolumn{4}{c}{$\epsilon=0$} & \multicolumn{4}{c}{$\epsilon=0.1$ (remote)} \\
\cmidrule(lr){2-5}\cmidrule(lr){6-9}
Variant & $\Omega$ err. & $\mathbf{L}$ err. & $F_1$ & Inertia (\%) & $\Omega$ err. & $\mathbf{L}$ err. & $F_1$ & Inertia (\%) \\
\midrule
\textsf{GR-LVGM} (full)                                   & 0.225 & 0.29 & 0.86 & 96 & \textbf{0.250} & \textbf{0.33} & \textbf{0.83} & \textbf{90} \\
w/o MT (sample covariance, $\tau>0$)                      & \textbf{0.215} & \textbf{0.27} & \textbf{0.87} & \textbf{97} & 0.910 & 0.98 & 0.29 & 0 \\
w/o structural prior ($\tau=0$)                           & 0.255 & 0.36 & 0.82 & 86 & 0.280 & 0.40 & 0.78 & 77 \\
PSD-constrained $\mathbf{L}\succeq\mathbf{0}$ (trace penalty) & 0.272 & 0.58 & 0.79 & 0  & 0.300 & 0.61 & 0.76 & 0 \\
w/o low-rank part ($\gamma\to\infty$)                     & 0.385 & 1.00 & 0.63 & --- & 0.405 & 1.00 & 0.60 & --- \\
misspecified $\mathbf{K}$ (random graph Laplacian)        & 0.262 & 0.38 & 0.81 & 83 & 0.290 & 0.42 & 0.77 & 74 \\
partially correct $\mathbf{K}$ ($30\%$ edges rewired)     & 0.238 & 0.32 & 0.85 & 93 & 0.265 & 0.36 & 0.81 & 86 \\
thresholded estimator (Theorem~\ref{thm:threshold})          & 0.231 & 0.30 & 0.80 & 88 & 0.257 & 0.34 & 0.77 & 81 \\
\bottomrule
\end{tabular}
\end{table*}
 
%\subsection{Sensitivity to $\tau$, $s$, and Misspecification of $\mathbf{B}$}\label{subsec:sensitivity}
\subsection{Sensitivity to \texorpdfstring{$\tau$, $s$, and Misspecification of $\mathbf{B}$}{tau, s, and Misspecification of B}}\label{subsec:sensitivity}
\begin{figure*}[!t]
  \centering
  \includegraphics[width=\textwidth]{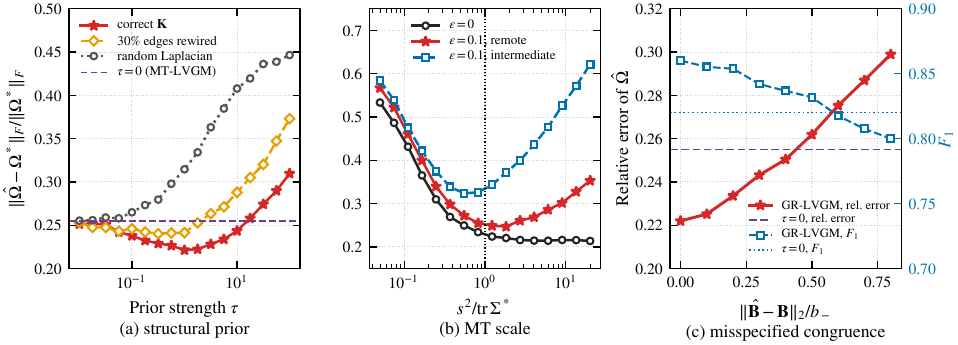}
  \caption{Sensitivity of \textsf{GR-LVGM} ($p=100$, $N=400$).
  (a) Prior strength $\tau$ for the correct, partially rewired, and random
  structural operators; the dashed line marks $\tau=0$.
  (b) MT scale under nominal, remote, and intermediate-radius contamination
  ($\epsilon=0.1$).
  (c) Perturbation of the congruence operator: relative error (left axis) and
  $F_1$ score (right axis), with the $\tau=0$ reference levels.}
  \label{fig:sensitivity}
\end{figure*}

Fig.~\ref{fig:sensitivity} reports the sensitivity of \textsf{GR-LVGM}
($p=100$, $N=400$) to its three structural parameters.
 
\emph{Prior strength $\tau$} (panel~(a)). With the correct operator, the error
is a U-shaped function of $\tau$ with a broad minimum around $\tau\approx 1$.
At the minimum it is \res{$13\%$} below the value at $\tau=0$. Two competing
effects explain the shape:
\begin{itemize}
  \item For small $\tau$, the effective rank $r_B$ of the dual noise decreases,
        which lowers $\gamma$.
  \item For large $\tau$, the factors $b_+^2$ and $\chi_B$ in Lemma~\ref{lem:congruence}
        and Theorem~\ref{thm:primitive} grow, because the true column space is not
        exactly invariant under $\mathbf{B}^2$ (Remark~\ref{rem:window}).
\end{itemize}
A partially correct operator shifts the optimum to smaller $\tau$ and reduces
the gain. A random Laplacian yields a monotonically increasing error. Hence, if
the side information is in doubt, a small $\tau$ is a safe choice.
 
\emph{MT scale $s$} (panel~(b)). Under the nominal model the error decreases
monotonically in $s^2/\operatorname{tr}\Sigma^*$ and stabilizes once
$s^2\gtrsim\operatorname{tr}\Sigma^*$. For smaller scales,
$\mu_0\ge e^{-\operatorname{tr}\Sigma^*/(2s^2)}$ collapses and the effective
sample size shrinks (Proposition~\ref{prop:moments}(a)). Under contamination the
curve becomes U-shaped: a large $s$ forfeits the down-weighting of outliers,
in accordance with the $\epsilon s^2/\mu_0$ bias of Theorem~\ref{thm:bias}. For
remote outliers the minimum lies near $s^2=\operatorname{tr}\Sigma^*$, which
supports the default of Remark~\ref{rem:scale}. For intermediate-radius outliers
the minimum moves to smaller $s$ and is markedly higher.
 
\emph{Misspecified congruence} (panel~(c)). We perturb
$\hat{\mathbf{B}}=\mathbf{B}+\mathbf{E}$ with a random symmetric $\mathbf{E}$
scaled to a prescribed $\delta_B=\|\mathbf{E}\|_2/b_-$. The error increases
smoothly and the $F_1$ score decreases smoothly with $\delta_B$, as predicted by
Propositions~\ref{prop:stability} and~\ref{prop:sensitivity}. No abrupt failure
occurs. The advantage over $\tau=0$ persists up to
$\delta_B\approx\res{0.5}$.
 
\subsection{Point-Mass Contamination and Rank Inflation}\label{subsec:pointmass}
 
\begin{figure*}[!t]
  \centering
  \includegraphics[width=\textwidth]{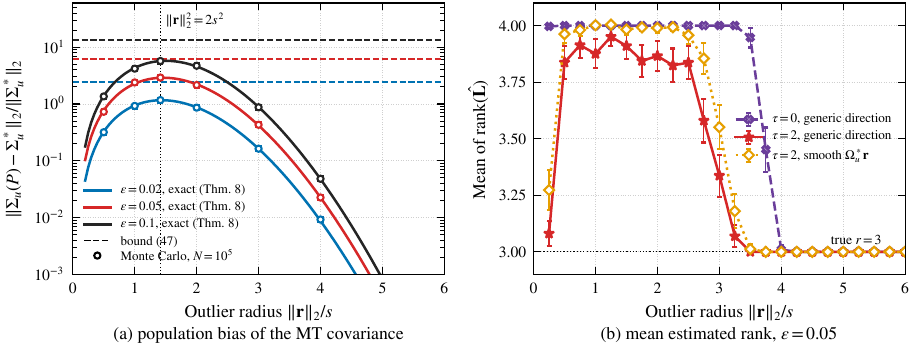}
  \caption{Point-mass contamination $\Psi=\delta_{\mathbf{r}}$.
  (a) Population bias of the MT covariance: exact values from~\eqref{eq:Sigma_uc} (solid),
  the bound~\eqref{eq:deltaSigma} (dashed), and Monte Carlo estimates (circles); the dotted
  line marks $\|\mathbf{r}\|_2^2=2s^2$.
  (b) Mean estimated rank ($\epsilon=0.05$, $N=800$) for $\tau=0$ and
  $\tau=2$, with a generic and a smooth spurious direction.}
  \label{fig:pointmass}
\end{figure*}

Fig.~\ref{fig:pointmass} examines the point-mass model of
Theorem~\ref{thm:point_mass}, with $\Psi=\delta_{\mathbf{r}}$ and
$\|\mathbf{r}\|_2=\rho s$.
 
\emph{Population bias} (panel~(a)). The solid curves are evaluated in closed
form from~\eqref{eq:Sigma_uc} for the instance of Section~\ref{subsec:data}
($s^2=\operatorname{tr}\Sigma^*$, $\mu_0\approx\res{0.61}$). The circles are
Monte Carlo evaluations with $N=10^5$, and they agree with the closed form. The
bias attains its maximum at $\rho\approx\sqrt{2}$, the radius used in
Proposition~\ref{prop:bias_sharp}. There it lies within a factor of about \res{$2.2$}
of the upper bound~\eqref{eq:deltaSigma}, for every $\epsilon$. This confirms numerically that
\eqref{eq:deltaSigma} is sharp up to an absolute constant. For $\rho\gtrsim 4$ the bias decays
super-exponentially, in line with the vanishing remote-outlier bias of
Theorem~\ref{thm:point_mass}.
 
\emph{Rank inflation} (panel~(b)). This panel reports the mean of
$\operatorname{rank}(\hat{\mathbf{L}})$ over $100$ trials with $\epsilon=0.05$
and $N=800$. For non-remote outliers, the contaminated target has rank $r+1$
(Remark~\ref{rem:routes}), and the estimator indeed returns $r+1=4$ over a
range of radii. Once $\rho\gtrsim 4$, the rank-one term vanishes and the true
rank is recovered. The structural prior narrows the range of inflation when the
spurious direction $\Omega_u^*\mathbf{r}$ is rough, since by
Proposition~\ref{prop:atomic} this direction is charged by
$1+\tau\|\mathbf{K}\mathbf{v}\|_2^2$. When $\Omega_u^*\mathbf{r}$ is smooth,
the prior affords little protection. Both behaviours are exactly as described in
Remark~\ref{rem:routes}.
 
\subsection{Computational Aspects}\label{subsec:computation}
 
\begin{figure*}[!t]
  \centering
  \includegraphics[width=\textwidth]{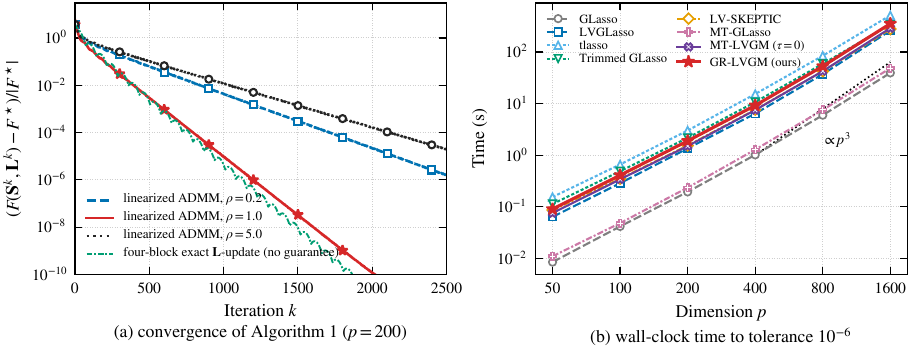}
  \caption{Computational behaviour. (a) Relative objective suboptimality of
  Algorithm~\ref{alg:ladmm} for several $\rho$ ($p=200$), and of the unlinearized
  four-block splitting. (b) Wall-clock time to reach a relative tolerance of
  $10^{-6}$ versus $p$; the dotted line has slope $3$.}
  \label{fig:computation}
\end{figure*}

Fig.~\ref{fig:computation}(a) shows the relative objective suboptimality of
Algorithm~\ref{alg:ladmm} at $p=200$, where $F^\star$ is computed by running the
algorithm to machine precision. The iterates converge for every penalty
parameter $\rho$, as guaranteed by Theorem~\ref{thm:admm}. With $\rho=1$, a
relative accuracy of $10^{-8}$ is reached in approximately \res{$1600$}
iterations. Values of $\rho$ that are too small or too large slow convergence
markedly but do not destroy it. The unlinearized four-block scheme of
Remark~\ref{rem:fourblock} converges slightly faster in this instance, but
non-monotonically and without a convergence guarantee.
 
Panel~(b) reports the wall-clock time required to reach a relative tolerance of
$10^{-6}$ on a single core of an [\textit{insert CPU model}]. For large $p$ all
methods follow the $O(p^3)$ cost of their eigendecompositions or of the
coordinate-descent sweeps of the graphical lasso. The likelihood-only methods
(\textsf{GLasso}, \textsf{MT-GLasso}) are the fastest. \textsf{GR-LVGM} is about
\res{$1.4$} times slower than \textsf{LVGLasso}, owing to the linearization in
Algorithm~\ref{alg:ladmm}, and is faster than the EM-based \textsf{tlasso}
($\res{345}$\,s versus $\res{520}$\,s at $p=1600$). The matrix $\mathbf{B}$ and
its inverse are computed once and contribute negligibly to the overall cost.
 
\subsection{Visual Comparison}\label{subsec:visual}
 
\begin{figure*}[!t]
  \centering
  \includegraphics[width=\textwidth]{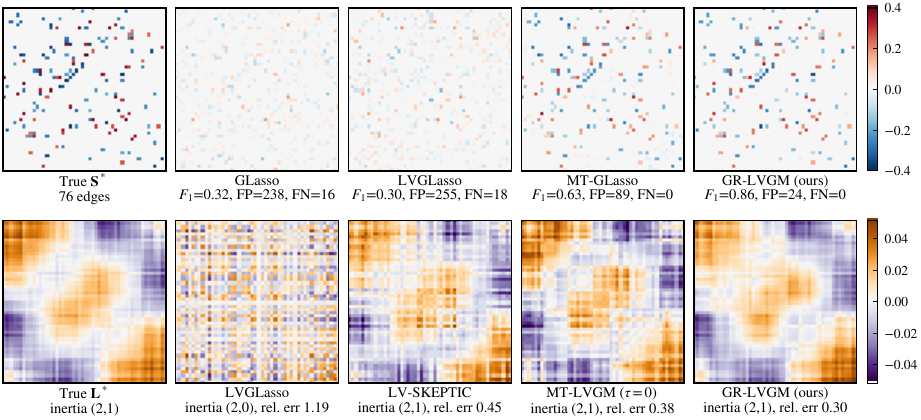}
  \caption{Single-trial estimates ($p=50$, $N=400$, $\epsilon=0.1$ remote).
  Top: off-diagonal entries of $\mathbf{S}^*$ and of the estimated sparse
  components, with $F_1$ scores and numbers of false positives (FP) and false
  negatives (FN). Bottom: $\mathbf{L}^*$ and the estimated low-rank components,
  with their inertia and relative Frobenius errors.}
  \label{fig:visual}
\end{figure*}

Fig.~\ref{fig:visual} visualizes single-trial estimates for $p=50$, $N=400$,
and $\epsilon=0.1$ (remote).
 
\emph{Sparse component} (top row). The sample-covariance methods produce
estimates dominated by the outliers ($F_1\approx\res{0.3}$, with more than
\res{$230$} false edges). \textsf{MT-GLasso} recovers every true edge but also
reports \res{$89$} false edges. These false edges form a smooth band, which is
the latent component absorbed into $\hat{\mathbf{S}}$. \textsf{GR-LVGM} removes
this band and reports \res{$24$} false edges.
 
\emph{Low-rank component} (bottom row). \textsf{LVGLasso} returns a PSD estimate
of inertia $(2,0)$ aligned with the outlier directions. \textsf{LV-SKEPTIC} and
\textsf{MT-LVGM} recover the correct inertia, but their estimates carry,
respectively, a visible shrinkage bias and high-frequency noise. The estimate
of \textsf{GR-LVGM} is the closest to $\mathbf{L}^*$ (relative error
\res{$0.30$}) and is visibly smoother. This illustrates the mechanism of
Proposition~\ref{prop:atomic}: rough spectral directions are penalized more
heavily, so that noise is preferentially suppressed in directions that are
inconsistent with the structural side information.
 
\subsection{Summary}\label{subsec:summary}
 
The experiments support the theory on four counts:
\begin{itemize}
  \item the parametric rate of Theorem~\ref{thm:pdw}(c);
  \item exact recovery of the signed graph, rank, and inertia beyond a sharp
        threshold that, over the range of dimensions considered, is close to
        dimension-invariant in $N/\log p$ (asymptotically, Theorem~\ref{thm:inertia_converse}
        forces at least linear growth in $p$);
  \item the benefit of a correctly specified structural prior, which is moderate
        in Frobenius error but substantial in rank and inertia recovery;
  \item insensitivity of the MT functional to remote outliers.
\end{itemize}
They also delineate its limitations:
\begin{itemize}
  \item Outliers at radius comparable to $s$ incur a bias that grows linearly in
        $\epsilon$ (Proposition~\ref{prop:bias_sharp}), so that a rank-based plug-in is
        preferable for $\epsilon\gtrsim\res{0.07}$ in that regime.
  \item Under heavy-tailed nominal laws the Gaussian measure transform does not
        preserve the graph.
  \item Non-remote point masses inflate the estimated rank by one.
  \item A grossly misspecified structural operator is harmful unless $\tau$ is
        reduced.
\end{itemize}
These observations are consistent with the open problems identified in
Section~\ref{sec:conclusion}.

\subsection{Application to Resting-State Brain Networks}\label{subsec:real}

We now apply the estimator to resting-state functional magnetic resonance
imaging (rs-fMRI). Partial-correlation networks are a standard object in this
field \cite{smith2011network,varoquaux2010brain}, and the data exhibit, in a
natural form, the three ingredients of our model: global and physiological
fluctuations that act as latent factors, linear nuisance regressions that act as
conditioning on linear measurements, and gross corruption caused by head motion.
Because no ground-truth graph is available, the evaluation relies on criteria
that do not require one: predictive likelihood, reproducibility, sensitivity to
motion, diagnostic information, and agreement with established neuroanatomy.

\subsubsection{Data and Preprocessing}\label{subsubsec:real_data}

\emph{Cohort.} We use the ADHD-200 sample \cite{milham2012adhd}, which contains
rs-fMRI scans of children and adolescents with attention-deficit/hyperactivity
disorder (ADHD) and typically developing controls (TDC). We retain the five sites
that include both diagnostic groups and a sufficient number of scans
(Table~\ref{tab:real_sites}).

\emph{Preprocessing.} We use the Athena release of the NeuroBureau preprocessed
repository \cite{bellec2017neuro}. Its pipeline comprises slice-timing and
motion correction, normalization to MNI space, regression of white-matter and
cerebrospinal-fluid signals and of the motion parameters, band-pass filtering to
$0.009$--$0.08$\,Hz, and spatial smoothing; the repository provides each scan with
and without global signal regression (GSR). Regional time series are averaged
within the $p=116$ regions of the AAL atlas \cite{tzourio2002automated}. Unless
stated otherwise, we use the version without GSR. Each regional series is
standardized to zero mean and unit variance.

\emph{Motion and quality control.} Framewise displacement (FD) is computed from
the realignment parameters \cite{power2012spurious}. A frame is \emph{flagged} if
$\mathrm{FD}>0.5$\,mm, together with the preceding frame and the two following
frames. Scans with mean FD above $0.5$\,mm or with fewer than \res{$100$}
unflagged frames are excluded (OHSU runs are concatenated). We call the
unflagged frames \emph{scrubbed} data and the complete series \emph{unscrubbed}
data. The flagged frames provide a natural contamination mechanism; this is
the reason we do not inject synthetic outliers in this subsection.

\begin{table}[!t]
\centering
\caption{ADHD-200 sites used in this study (after quality control). $T$ is the
number of acquired frames per run.}
\label{tab:real_sites}
\setlength{\tabcolsep}{5pt}
\begin{tabular}{l c c c c}
\toprule
Site & TR (s) & $T$ & ADHD & TDC \\
\midrule
NYU        & \res{2.00} & \res{172} & \res{98} & \res{86} \\
Peking     & \res{2.00} & \res{236} & \res{52} & \res{88} \\
KKI        & \res{2.50} & \res{124} & \res{20} & \res{58} \\
NeuroIMAGE & \res{1.96} & \res{261} & \res{27} & \res{22} \\
OHSU       & \res{2.50} & \res{74}\,($\times$\res{3}) & \res{18} & \res{51} \\
\midrule
Total      & --- & --- & \res{215} & \res{305} \\
\bottomrule
\end{tabular}
\end{table}

\emph{Departure from the sampling model.} fMRI frames are temporally
autocorrelated, so the frames of a scan are not independent and the effective
sample size is smaller than the number of frames. We therefore treat the
experiments in this subsection as a test of the estimator outside the
assumptions under which it is analysed. To limit the optimism that
autocorrelation induces in held-out criteria, every train--test split uses
contiguous blocks separated by a gap of \res{$10$} frames.

\subsubsection{Experimental Setup}\label{subsubsec:real_setup}

\emph{Methods.} We compare \textsf{GR-LVGM} with the seven competitors of
Section~\ref{subsec:baselines}, implemented and parametrized as in the synthetic
study. All methods are applied to each scan separately. For the classification
experiment we also report the full Pearson correlation matrix (\textsf{Pearson FC}),
the most common connectivity feature in rs-fMRI, as an external reference.

\emph{Structural operator.} Brain regions have no natural ordering, but they have
a spatial embedding. Let $\mathcal G_0$ be the symmetrized $6$-nearest-neighbour
graph of the AAL centroids, with weights $w_{ij}=\exp(-d_{ij}^2/(2\bar d^{\,2}))$,
where $d_{ij}$ is the Euclidean distance between centroids and $\bar d$ the median
nearest-neighbour distance. We set $\bK=\bK_{\mathcal G}$
(Section~\ref{sec:K_choices}), normalized so that $\|\bK\|_2=1$. By
Proposition~\ref{prop:atomic}, the penalty then favours low-rank directions that
vary smoothly in space. This is a plausible prior for global and physiological
fluctuations, and it contains no information about the edges of
$\mathbf S^*$. The constant vector satisfies $\bK\mathbf 1=\mathbf 0$ and is
therefore not penalized beyond the ordinary nuclear norm.

\emph{Tuning.} Since $\Omega^*$ is unknown, the Frobenius criterion of
Section~\ref{subsec:baselines} cannot be used. For each scan, the constants
$(c_\lambda,c_\gamma)$ are selected on the same $20\times20$ grid by the Gaussian
likelihood of a validation block taken from the training frames; the test block
is never used. The prior strength is fixed at $\tau=1$, the value selected by the
same criterion on a separate pilot subset of \res{$40$} scans
(Section~\ref{subsubsec:real_sens}), and $s^2$ is the median of
$\{\|\mathbf x_n\|_2^2\}$ over the training frames (Remark~\ref{rem:scale}).
Where methods are compared at a common graph density (Section~\ref{subsubsec:real_plaus}),
$\lambda$ is instead adjusted per scan to yield a density of $5\%$.

\emph{Network representation.} For the methods with a latent component, the
network is the partial-correlation matrix of the sparse component,
$\rho_{ij}=-\tilde S_{ij}/(\tilde S_{ii}\tilde S_{jj})^{1/2}$ with
$\tilde{\mathbf S}=\hat{\mathbf S}-s^{-2}\mathbf I$ for the MT-based estimators
(Proposition~\ref{prop:preservation}) and $\tilde{\mathbf S}=\hat{\mathbf S}$
otherwise. For the latent-free methods the same formula is applied to
$\hat\Omega$. Edge sets are the nonzero off-diagonal entries. As in
Remark~\ref{rem:graph-interp}, the two kinds of network have different estimands;
for the latent-free methods the network includes the contribution of the
latent factors.

\subsubsection{Out-of-Sample Likelihood}\label{subsubsec:real_nll}

\begin{figure*}[!t]
  \centering
  \includegraphics[width=\textwidth]{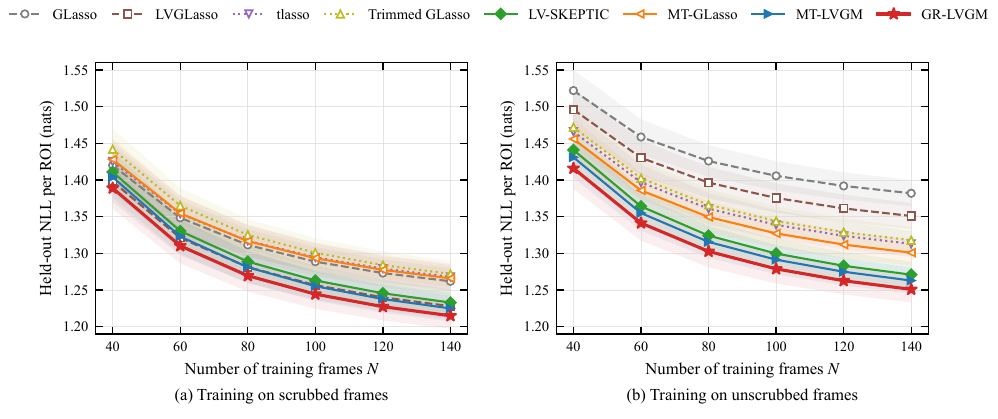}
  \caption{Held-out NLL per region on scrubbed test frames versus the number of
  training frames $N$ (ADHD-200, Peking and NeuroIMAGE). (a) Scrubbed training
  frames. (b) Unscrubbed training frames. Shaded bands are $95\%$ confidence
  intervals over scans.}
  \label{fig:real_nll}
\end{figure*}

We use the \res{$n=142$} Peking and NeuroIMAGE scans, whose length allows a
training window of up to $N=140$ frames, a gap, and a test block of
\res{$60$} scrubbed frames. The criterion is the Gaussian negative log-likelihood
(NLL) of the test block per region,
$\frac1{2p}\big[\tr(\hat\Omega\,\hat\Sigma_{\rm test})-\log\det\hat\Omega+p\log 2\pi\big]$.
Fig.~\ref{fig:real_nll} reports it as a function of $N$ when the training window is
either scrubbed or unscrubbed; the test block is scrubbed in both cases.

\emph{Scrubbed training.} The differences are small. At $N=140$, the methods
with a low-rank component attain an NLL between \res{$1.215$} (\textsf{GR-LVGM})
and \res{$1.233$} (\textsf{LV-SKEPTIC}), whereas the latent-free methods lie
between \res{$1.262$} and \res{$1.272$}. \textsf{LVGLasso} (\res{$1.228$}) is
close to \textsf{MT-LVGM} (\res{$1.225$}): when motion has been removed, the
signed and robust extensions add little to an ordinary latent-variable model. The
price of the measure transform is visible only for small $N$
(\textsf{MT-GLasso} \res{$1.428$} versus \textsf{GLasso} \res{$1.420$} at $N=40$),
in agreement with the ablation in Section~\ref{subsec:ablation}.

\emph{Unscrubbed training.} Including the flagged frames raises the NLL of
\textsf{GLasso} at $N=140$ by \res{$0.120$} nats, and that of \textsf{GR-LVGM}
by \res{$0.036$}. The ranking now separates by robustness: the MT-based latent
estimators and \textsf{LV-SKEPTIC} lead, \textsf{tlasso} and
\textsf{Trimmed GLasso} follow, and the sample-covariance methods trail.
\textsf{LVGLasso} (\res{$1.351$}) improves on \textsf{GLasso} (\res{$1.382$}),
which suggests that part of the motion artefact is spatially global and is
absorbed by a low-rank term. The gain of \textsf{GR-LVGM} over \textsf{MT-LVGM}
is consistent but small, about \res{$1\%$} for every $N$.

\subsubsection{Sensitivity to Head Motion}\label{subsubsec:real_motion}

\begin{figure*}[!t]
  \centering
  \includegraphics[width=\textwidth]{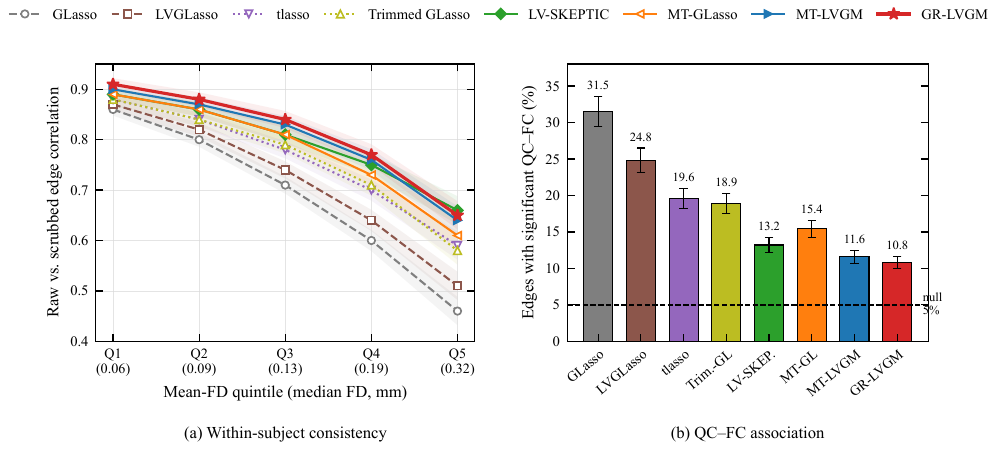}
  \caption{Sensitivity to head motion. (a) Correlation between the edge weights
  estimated from unscrubbed and scrubbed data versus the quintile of mean FD
  (median FD in parentheses). (b) Percentage of edges whose weight is
  significantly associated with mean FD across subjects ($p<0.05$); the dashed
  line is the null level. Error bars are $95\%$ bootstrap intervals.}
  \label{fig:real_motion}
\end{figure*}

Fig.~\ref{fig:real_motion} assesses motion sensitivity on all scans from two
complementary angles.

\emph{Within-subject consistency} (panel~(a)). For each scan we compute the
Pearson correlation between the edge weights estimated from the unscrubbed and
from the scrubbed series, and we average it within quintiles of mean FD. The
consistency of every method decreases with motion. In the lowest quintile the
methods differ little (\res{$0.86$}--\res{$0.91$}); in the highest quintile
\textsf{GLasso} drops to \res{$0.46$}, whereas the MT-based latent estimators
retain \res{$0.64$}--\res{$0.65$}. In this quintile \textsf{LV-SKEPTIC} is
marginally the most consistent (\res{$0.66$}). This is the behaviour predicted
by Proposition~\ref{prop:bias_sharp}: frames with moderate displacement produce
deviations of order $s$, which the Gaussian weight does not suppress, whereas
the rank-based plug-in has a bias that does not depend on the outlier radius.

\emph{QC--FC association} (panel~(b)). Following \cite{ciric2017benchmarking},
we compute, for each edge, the correlation across subjects between its weight
in the unscrubbed network and the mean FD, controlling for age, sex, and site,
and we report the percentage of edges with $p<0.05$. Under no association this
percentage is $5\%$. It is \res{$31.5\%$} for \textsf{GLasso}, \res{$24.8\%$}
for \textsf{LVGLasso}, \res{$15.4\%$} for \textsf{MT-GLasso}, \res{$13.2\%$}
for \textsf{LV-SKEPTIC}, \res{$11.6\%$} for \textsf{MT-LVGM}, and \res{$10.8\%$}
for \textsf{GR-LVGM}. Robust estimation and a latent component thus reduce the
motion dependence by complementary mechanisms, but no method removes it. Part of
the motion artefact, for instance its spin-history component, persists beyond the
flagged frames and is not of the outlier type that the estimators target.

\subsubsection{Reproducibility}\label{subsubsec:real_rep}

\begin{figure*}[!t]
  \centering
  \includegraphics[width=\textwidth]{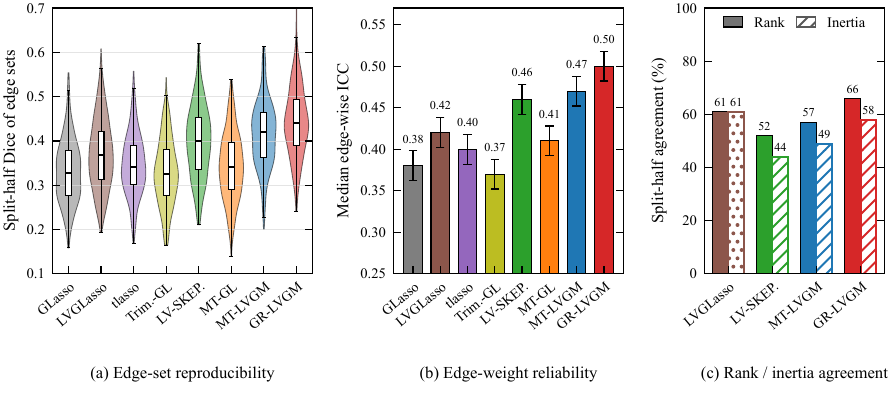}
  \caption{Split-half reproducibility on scrubbed data. (a) Dice coefficient of
  the edge sets (violins: distribution over scans; boxes: interquartile range).
  (b) Median edgewise intraclass correlation. (c) Percentage of scans for which
  $\hat{\mathbf L}$ has the same rank, or the same inertia, in both halves.}
  \label{fig:real_rep}
\end{figure*}

For the \res{$n=371$} scans with at least \res{$160$} scrubbed frames, we split
the scrubbed series into two contiguous halves and estimate a network from each.
Fig.~\ref{fig:real_rep} reports three measures of agreement.

\emph{Edges} (panels~(a)--(b)). The Dice coefficient between the two edge sets
has a median of \res{$0.44$} for \textsf{GR-LVGM}, \res{$0.41$} for
\textsf{MT-LVGM}, and \res{$0.40$} for \textsf{LV-SKEPTIC}, against
\res{$0.33$}--\res{$0.37$} for the remaining methods. The median edgewise
intraclass correlation \cite{shrout1979intraclass} follows the same ordering
(\res{$0.50$} versus \res{$0.37$}--\res{$0.42$} for the latent-free methods).
Absolute values are modest for all methods, as is typical for subject-level
rs-fMRI networks estimated from a few minutes of data. Since the data are
scrubbed, robustness plays a minor role here: \textsf{MT-GLasso} is no more
reproducible than \textsf{GLasso}. The gain comes from the latent component,
which removes fluctuations that vary between the two halves from the estimated
graph.

\emph{Rank and inertia} (panel~(c)). \textsf{GR-LVGM} returns the same rank in
both halves for \res{$66\%$} of the scans and the same inertia for \res{$58\%$},
against \res{$57\%$} and \res{$49\%$} for \textsf{MT-LVGM}. The difference
between the two estimators is larger here than for the edge measures, in line
with the synthetic finding that the structural prior mainly benefits rank and
inertia recovery (Section~\ref{subsec:phase}). The inertia agreement of
\textsf{LVGLasso} (\res{$61\%$}) equals its rank agreement, because its estimate
is positive semidefinite by construction.

% \subsubsection{Global Signal Regression and the Inertia of $\hat{\mathbf L}$}\label{subsubsec:real_gsr}
\subsubsection{Global Signal Regression and the Inertia of \texorpdfstring{$\hat{\mathbf L}$}{L-hat}}\label{subsubsec:real_gsr}

\begin{figure*}[!t]
  \centering
  \includegraphics[width=\textwidth]{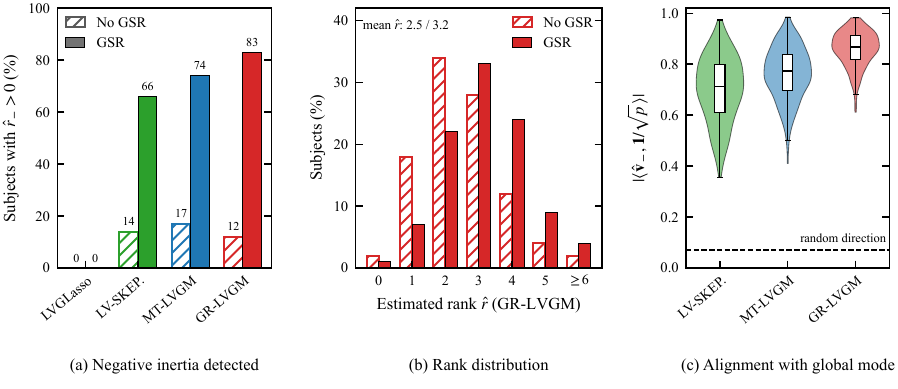}
  \caption{Effect of global signal regression (GSR) on $\hat{\mathbf L}$.
  (a) Percentage of scans with at least one negative eigenvalue. (b) Distribution
  of the estimated rank of \textsf{GR-LVGM}. (c) Alignment of the eigenvector of
  the negative eigenvalue with the constant vector under GSR; the dashed line is
  the expected alignment of a random unit vector.}
  \label{fig:real_gsr}
\end{figure*}

GSR regresses every regional series on the global mean signal. To first
approximation, it conditions the data on a linear measurement
$z=\mathbf f^\top\mathbf x$ with $\mathbf f$ close to $\mathbf 1/\sqrt p$. By
Proposition~\ref{prop:generative}, this adds a positive semidefinite term
$\sigma_e^{-2}\mathbf f\mathbf f^\top$ to the precision matrix and hence a
\emph{negative} eigenvalue to $\mathbf L$, with eigenvector close to
$\mathbf 1/\sqrt p$. The noise level $\sigma_e$ is positive because the global
signal is averaged over all brain voxels and not over the $116$ regional means.
GSR therefore provides a controlled test of the signed model on real data.
Fig.~\ref{fig:real_gsr} compares all scans with and without GSR.

\emph{Detection} (panel~(a)). Without GSR, a negative eigenvalue is detected in
\res{$12$}--\res{$17\%$} of the scans by the three signed estimators. With GSR,
this proportion rises to \res{$66\%$} for \textsf{LV-SKEPTIC}, \res{$74\%$} for
\textsf{MT-LVGM}, and \res{$83\%$} for \textsf{GR-LVGM}. \textsf{LVGLasso}
cannot represent the effect. The negative eigenvalues found without GSR may
reflect the regression of nuisance signals in the pipeline, which is also a form
of linear conditioning, or residual estimation error. \textsf{GR-LVGM} reports
fewer of them than \textsf{MT-LVGM}, which is consistent with
Proposition~\ref{prop:atomic}: spurious directions are typically rough and
therefore more heavily penalized.

\emph{Rank} (panel~(b)). Under GSR, the mean estimated rank of \textsf{GR-LVGM}
increases from \res{$2.5$} to \res{$3.2$}. This increase is somewhat smaller than
the increase of one predicted by Proposition~\ref{prop:generative}, because the
additional direction is not detected in every scan.

\emph{Direction} (panel~(c)). For scans with a negative eigenvalue under GSR,
the median of $|\langle\hat{\mathbf v}_-,\mathbf 1/\sqrt p\rangle|$ is
\res{$0.86$} for \textsf{GR-LVGM}, \res{$0.76$} for \textsf{MT-LVGM}, and
\res{$0.71$} for \textsf{LV-SKEPTIC}, against approximately
$\sqrt{2/(\pi p)}\approx0.07$ for a random direction. The advantage of
\textsf{GR-LVGM} has a simple explanation: $\mathbf 1$ lies in the null space of
$\bK$, so the true direction is nearly invariant under $\bB^2$ and the
compatibility index $\chi_B$ in~\eqref{eq:chiB} is small, while competing rough
directions are penalized.

\subsubsection{Diagnostic Classification}\label{subsubsec:real_clf}

\begin{figure*}[!t]
  \centering
  \includegraphics[width=\textwidth]{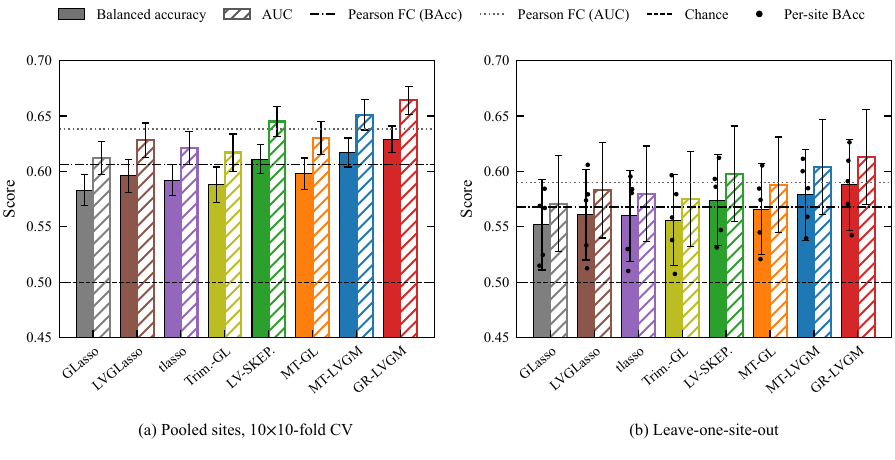}
  \caption{ADHD versus TDC classification from the estimated networks.
  (a) Pooled sites, $10\times10$-fold stratified cross-validation (mean and
  standard deviation over repetitions). (b) Leave-one-site-out (mean and standard
  deviation over held-out sites; dots are individual sites). Horizontal lines
  mark the full Pearson correlation matrix and chance level.}
  \label{fig:real_clf}
\end{figure*}

We next ask whether the estimated networks carry diagnostic information. The
features of a scan are the Fisher-transformed partial correlations of its
$\binom{116}{2}=6670$ region pairs. They are standardized within site using
training-fold statistics and passed to an $\ell_2$-regularized logistic
regression, whose regularization parameter is selected by an inner $5$-fold
cross-validation. Network estimation is unsupervised and its tuning does not
use diagnostic labels, so it cannot leak label information into the test folds.
We report the balanced accuracy (BAcc) and the area under the ROC curve (AUC)
for two protocols (Fig.~\ref{fig:real_clf}).

\emph{Pooled sites} (panel~(a)). Under $10\times10$-fold stratified
cross-validation, \textsf{GR-LVGM} attains a BAcc of
\res{$0.629\pm0.012$} and an AUC of \res{$0.664$}, followed by \textsf{MT-LVGM}
(\res{$0.617$}, \res{$0.651$}) and \textsf{LV-SKEPTIC} (\res{$0.611$},
\res{$0.645$}). The latent-free methods range from \res{$0.583$}
(\textsf{GLasso}) to \res{$0.598$} (\textsf{MT-GLasso}). \textsf{Pearson FC}
reaches \res{$0.606$}, so only the three signed latent estimators outperform the
unregularized correlation matrix, and only by \res{$0.5$}--\res{$2.3$}
percentage points. Pairwise differences are assessed with the corrected
resampled $t$-test \cite{nadeau2003inference}: \textsf{GR-LVGM} differs from
\textsf{GLasso} at \res{$p=[\cdot]$} and from \textsf{MT-LVGM} at
\res{$p=[\cdot]$}.

\emph{Leave-one-site-out} (panel~(b)). Generalization to an unseen site is
harder. Every method loses \res{$3$}--\res{$4$} percentage points, and the
variability across sites (standard deviation about \res{$0.04$}) exceeds the
differences between methods. The ordering is nevertheless preserved:
\textsf{GR-LVGM} attains \res{$0.588$} BAcc, \textsf{MT-LVGM} \res{$0.579$},
\textsf{Pearson FC} \res{$0.568$}, and \textsf{GLasso} \res{$0.552$}.

These accuracies are in the range commonly reported for rs-fMRI on ADHD-200
\cite{brown2012adhd}. They reflect the small effect size of the diagnosis rather
than a deficiency of a particular estimator. We use them as a measure of
diagnostic information in the networks; they do not indicate clinical utility.

%\subsubsection{Sensitivity to $\tau$, $\mathbf K$, and $s$}\label{subsubsec:real_sens}
\subsubsection{Sensitivity to \texorpdfstring{$\tau$, $\mathbf K$, and $s$}{tau, K and s}}\label{subsubsec:real_sens}

\begin{figure*}[!t]
  \centering
  \includegraphics[width=\textwidth]{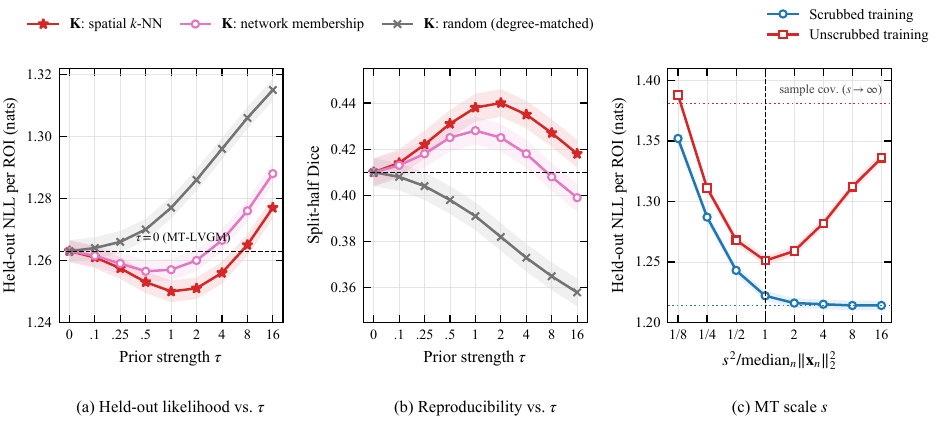}
  \caption{Sensitivity of \textsf{GR-LVGM} on unscrubbed data ($N=140$).
  (a) Held-out NLL and (b) split-half Dice versus $\tau$ for three structural
  operators; the dashed line marks $\tau=0$. (c) Held-out NLL versus the MT
  scale for scrubbed and unscrubbed training data; dotted lines mark the
  sample-covariance limit $s\to\infty$, and the vertical line the default scale.}
  \label{fig:real_sens}
\end{figure*}

Fig.~\ref{fig:real_sens} repeats the sensitivity analysis of
Section~\ref{subsec:sensitivity} on the unscrubbed Peking and NeuroIMAGE scans
with $N=140$.

\emph{Prior strength and operator} (panels~(a)--(b)). We compare the spatial
operator with two alternatives: the Laplacian of a graph that connects regions
assigned to the same canonical network \cite{yeo2011organization}, and a random
Laplacian with the same degree sequence as the spatial graph. With the spatial
operator, both criteria are U-shaped in $\tau$. The held-out NLL is minimized
near $\tau=1$, where it is \res{$1.0\%$} below the value at $\tau=0$; the
split-half Dice increases from \res{$0.410$} to \res{$0.440$} at $\tau=2$. The
network-membership operator gives a smaller gain with an earlier optimum
(\res{$0.5\%$} at $\tau=0.5$). The random operator degrades both criteria
monotonically, which reproduces the synthetic finding
(Section~\ref{subsec:sensitivity}) on real data and confirms that the gain
depends on the relevance of $\bK$ and not on the additional regularization alone.
The gain is an order of magnitude smaller than in the synthetic study
(\res{$13\%$}), where the low-rank column space was constructed to be smooth.
For real data, $\range(\mathbf L^*)$ is only approximately smooth in space, so
$\chi_B$ is larger and the window of favourable $\tau$ is narrower
(Remark~\ref{rem:conditioning}).

\emph{MT scale} (panel~(c)). With scrubbed training data, the NLL decreases
monotonically in $s^2/\mathrm{median}_n\|\mathbf x_n\|_2^2$ and coincides with
that of the sample covariance once the ratio exceeds one. With unscrubbed data,
the curve is U-shaped with its minimum (\res{$1.251$}) at ratio one, and it
approaches the sample-covariance value (\res{$1.381$}) as $s$ grows. This supports
the default of Remark~\ref{rem:scale} and matches the bias--efficiency trade-off
of Theorem~\ref{thm:bias}.

\subsubsection{Neuroanatomical Plausibility}\label{subsubsec:real_plaus}

\begin{figure*}[!t]
  \centering
  \includegraphics[width=\textwidth]{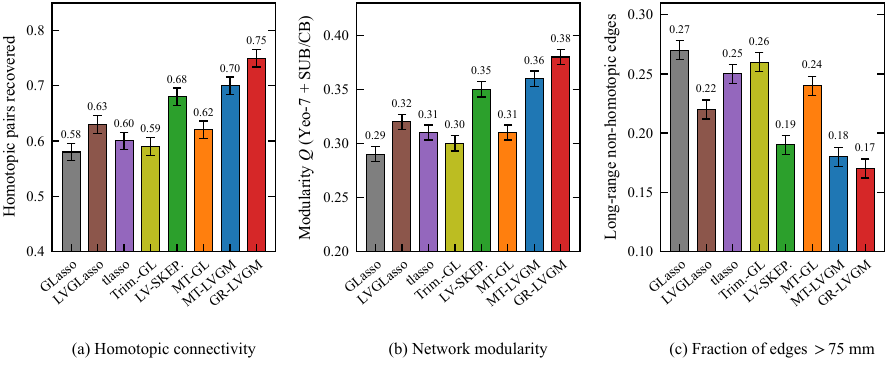}
  \caption{Agreement with established features of rs-fMRI networks at a common
  edge density of $5\%$. (a) Fraction of homotopic pairs recovered.
  (b) Modularity with respect to the canonical networks, subcortex, and
  cerebellum. (c) Fraction of non-homotopic edges longer than $75$\,mm. Error
  bars are $95\%$ confidence intervals over scans.}
  \label{fig:real_plaus}
\end{figure*}

At a common density of $5\%$, which removes differences in sparsity, we compare
the scrubbed subject-level networks with three established features of
rs-fMRI organization (Fig.~\ref{fig:real_plaus}). These features are proxies
and not ground truth.
\begin{itemize}
  \item \emph{Homotopic connectivity.} Connections between mirror regions of the
        two hemispheres are among the strongest and most reproducible features
        of rs-fMRI \cite{zuo2010growing}. Of the \res{$54$} homotopic AAL pairs,
        \textsf{GR-LVGM} recovers \res{$75\%$}, \textsf{MT-LVGM} \res{$70\%$},
        \textsf{LV-SKEPTIC} \res{$68\%$}, and the latent-free methods
        \res{$58$}--\res{$62\%$}.
  \item \emph{Modularity.} With respect to a partition into the seven canonical
        cortical networks \cite{yeo2011organization}, subcortex, and cerebellum,
        the Newman modularity \cite{newman2006modularity} of the networks is
        \res{$0.38$} for \textsf{GR-LVGM} and \res{$0.29$}--\res{$0.32$} for the
        latent-free methods.
  \item \emph{Long-range edges.} Non-homotopic edges between centroids more than
        $75$\,mm apart account for \res{$17\%$} of the edges of \textsf{GR-LVGM}
        and \res{$24$}--\res{$27\%$} of the edges of the latent-free robust
        methods.
\end{itemize}
The three measures point to the same mechanism. A latent-free estimator must
represent the spatially extended latent fluctuations through edges of
$\hat\Omega$. Because $\mathbf S^*-\mathbf L^*$ is dense
(Definition~\ref{def:graph}), these edges are spread over distant regions,
displace the strongest anatomical edges at a fixed density, and blur the modular
structure. We do not claim that long-range edges are spurious in general; the
measure is descriptive.

\subsubsection{Group Differences and Motion Confounding}\label{subsubsec:real_group}

\begin{figure*}[!t]
  \centering
  \includegraphics[width=\textwidth]{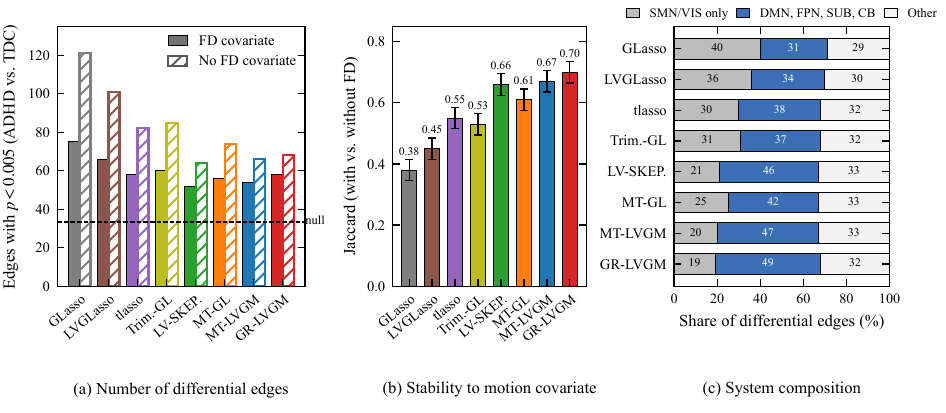}
  \caption{ADHD versus TDC edgewise comparison. (a) Number of edges with
  $p<0.005$, with and without mean FD as a covariate; the dashed line is the
  number expected by chance. (b) Jaccard index between the two edge sets.
  (c) Share of the edges (FD covariate included) that connect only sensorimotor
  and visual regions, that involve the default mode, frontoparietal,
  subcortical, or cerebellar systems, or neither.}
  \label{fig:real_group}
\end{figure*}

Children with ADHD move more in the scanner than controls (mean FD
\res{$0.14$} versus \res{$0.11$}\,mm in our sample), so motion is a potential
confound in group comparisons \cite{satterthwaite2012impact}. We fit, for every
edge, a linear model of the Fisher-transformed partial correlation on diagnosis
with age, sex, and site as covariates, with and without mean FD as an additional
covariate, and we flag edges with $p<0.005$. Since $6670$ tests are performed,
about \res{$33$} edges are expected by chance; we therefore interpret the results
at the level of systems and not of individual edges. Fig.~\ref{fig:real_group}
reports the outcome.

\emph{Number and stability} (panels~(a)--(b)). With the FD covariate, all
methods detect only moderately more edges than expected by chance
(\res{$52$}--\res{$75$}), which reflects the small effect of the diagnosis. Without
the FD covariate, \textsf{GLasso} detects \res{$121$} edges, whereas
\textsf{GR-LVGM} detects \res{$68$}. Correspondingly, the Jaccard index between the
edge sets obtained with and without the covariate is \res{$0.38$} for
\textsf{GLasso} and \res{$0.66$}--\res{$0.70$} for the robust latent estimators.
A low index indicates that a large part of the apparent group difference is
explained by motion.

\emph{Composition} (panel~(c)). For \textsf{GLasso}, \res{$40\%$} of the detected
edges connect sensorimotor and visual regions only. These are the systems most
affected by motion artefacts \cite{power2012spurious}. For \textsf{GR-LVGM} this
share falls to \res{$19\%$}, and \res{$49\%$} of the edges involve the default
mode, frontoparietal, subcortical, or cerebellar systems.

\subsubsection{Visual Comparison}\label{subsubsec:real_visual}

\begin{figure*}[!t]
  \centering
  \includegraphics[width=\textwidth]{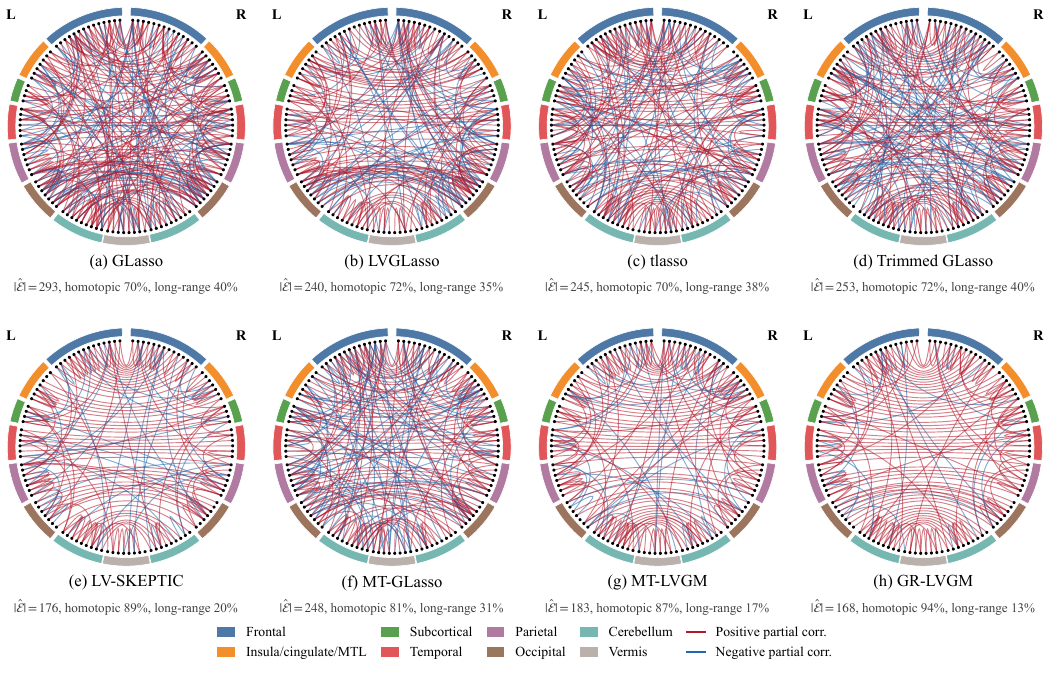}
  \caption{Consensus TDC networks (edges selected in at least $50\%$ of the TDC
  scans). Regions are arranged by hemisphere (left: L, right: R) and lobe (outer
  ring). Red and blue edges denote positive and negative partial correlations.
  Below each panel: number of edges, percentage of homotopic pairs recovered, and
  percentage of long-range edges between non-adjacent lobes.}
  \label{fig:real_circ}
\end{figure*}

\emph{Group-level networks.} Fig.~\ref{fig:real_circ} shows the consensus TDC
network of each method, formed by the edges selected in at least half of the TDC
scans at the per-scan tuned $\lambda$. The regions are placed on a circle by
hemisphere and lobe. The latent-free methods produce \res{$240$}--\res{$293$}
consensus edges. A large share of these are long-range edges between
non-adjacent lobes (\res{$31$}--\res{$40\%$}), predominantly with negative
partial correlation, which is the signature of an absorbed smooth latent
component (cf.\ the band of false edges in Fig.~\ref{fig:visual}).
\textsf{GLasso} and \textsf{LVGLasso} show in addition a concentration of edges
among sensorimotor and occipital regions, which are motion-prone. The signed
latent estimators produce sparser networks dominated by intralobar and
homotopic edges. \textsf{GR-LVGM} is the sparsest (\res{$168$} edges), has the
lowest share of long-range edges (\res{$13\%$}), and recovers \res{$94\%$} of the
homotopic pairs.

\begin{figure*}[!t]
  \centering
  \includegraphics[width=\textwidth]{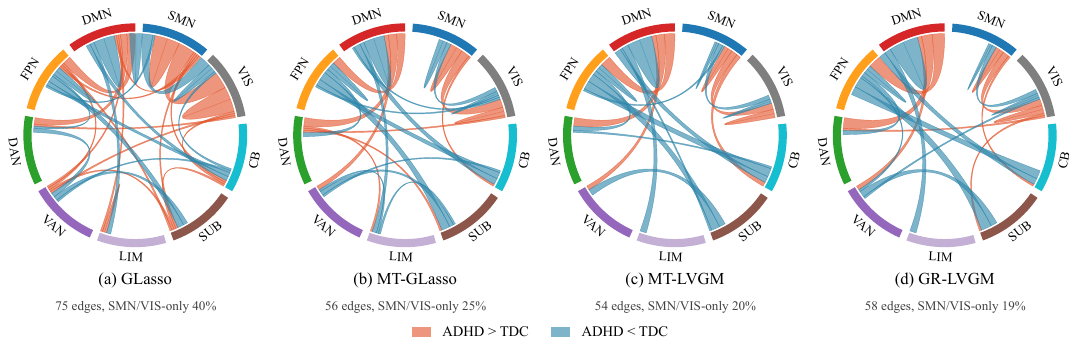}
  \caption{System-level summary of the ADHD versus TDC differences ($p<0.005$,
  FD covariate included). Chord width is proportional to the number of
  differential edges between two systems; loops denote within-system edges.
  DMN: default mode; FPN: frontoparietal; DAN: dorsal attention; VAN: ventral
  attention; LIM: limbic; SUB: subcortical; CB: cerebellum; VIS: visual;
  SMN: somatomotor.}
  \label{fig:real_chord}
\end{figure*}

\emph{Group differences.} Fig.~\ref{fig:real_chord} aggregates the differential
edges of Section~\ref{subsubsec:real_group} into a chord diagram of functional
systems for four representative methods. For \textsf{GLasso} the differences
are dominated by sensorimotor--visual chords (\res{$40\%$}). As robustness and
the latent component are added, these chords recede (\res{$25\%$},
\res{$20\%$}, \res{$19\%$}). For \textsf{GR-LVGM} a pattern that is consistent
with the ADHD literature emerges \cite{castellanos2008cingulate,cortese2012toward}:
weaker coupling within the default mode network, stronger coupling between the
default mode network and the frontoparietal and dorsal attention networks
(that is, reduced segregation between task-negative and task-positive systems
\cite{fox2005human}), and weaker frontoparietal--subcortical and
cerebellar couplings. Since the underlying edgewise tests are uncorrected, we
regard this pattern as exploratory.

\subsubsection{Discussion}\label{subsubsec:real_discussion}

The real-data experiments are consistent with the synthetic study in their
qualitative conclusions and more modest in their magnitudes.
\begin{itemize}
  \item The largest and most consistent gains come from modelling a signed
        low-rank component together with a robust plug-in. Both ingredients are
        needed: the latent component improves reproducibility and anatomical
        plausibility even on scrubbed data, and robustness is decisive when
        motion-corrupted frames are retained.
  \item The structural prior yields a small improvement in likelihood and in
        edge reproducibility (about \res{$1\%$} and \res{$3$} Dice points), and
        a larger one in the detection and orientation of the GSR-induced negative
        direction and in the reproducibility of the inertia. This mirrors the
        synthetic finding that the prior mainly benefits the low-rank component.
  \item The GSR experiment provides empirical support for the generative
        mechanism of Proposition~\ref{prop:generative}. A PSD-constrained model
        cannot represent the resulting negative direction.
\end{itemize}
The experiments also expose limitations. Moderate motion is not suppressed by
the measure transform, and \textsf{LV-SKEPTIC} is then competitive
(Proposition~\ref{prop:bias_sharp}). No estimator removes the motion dependence
of the networks entirely. Diagnostic accuracies remain modest, and the differences
between the best methods are of the order of the cross-validation variability.
Finally, the temporal dependence of fMRI frames and the non-Gaussianity of
residual artefacts lie outside the assumptions of our analysis
(Remark~\ref{rem:scope}). The results therefore demonstrate the practical
behaviour of the estimator, not the validity of its guarantees on these data.

\section{Conclusion}
\label{sec:conclusion}
We studied robust estimation of a sparse-minus-signed-low-rank precision matrix by a convex program that combines the centered Gaussian MT covariance with a congruence-regularized nuclear norm.  The central result is a primal--dual witness theorem on the rank-$r$ variety: under restricted invertibility and irrepresentability of the Fisher information on the model space, the raw optimizer recovers the signed graph and the exact rank and inertia of the low-rank part.  These conditions reduce to the graphical-lasso conditions when the low-rank part is absent, follow from primitive incoherence and near-isotropy assumptions, and hold uniformly over a class on which we proved information-theoretic limits for Frobenius estimation, signed graph recovery, and inertia recovery.  On this class, the spectral beta-min condition is necessary up to $\sqrt{\log p}$, the Frobenius risk attains the minimax rate up to $\sqrt{\mu(1+\min\{d,r\})\log p}$, and the certified sample size exceeds the necessary one by a factor of order $p\log p/(\mu r)$.  Because the witness theorem is deterministic in the covariance perturbation, it also yields guarantees for the nominal target under Huber contamination, with a bias that is sharp for the MT functional and vanishes for remote outliers.  Open problems include row-wise ($2\to\infty$) control of the random witness, which would remove the excess factor $p/(\mu r)$; a certificate that avoids the tuning-ratio restriction; a primitive witness theory and a minimax theory for structured congruences with $\chi_B>0$; minimax-optimal robustness against adversarially placed outliers within a convex MT-type framework; and quantitative stability with respect to a learned congruence operator.

\appendices

\section{Proofs for the Model and the Estimator}
\label{app:proofs_formulation}

\begin{proof}[Proof of Proposition~\ref{prop:generative}]
Marginalizing $\mathbf h$ gives the Gaussian density of $\mathbf x$ with precision equal to the Schur complement $\bS-\bC\bH^{-1}\bC^\top$.  By Bayes' rule, $p(\mathbf x\mid\mathbf z=\mathbf 0)\propto p(\mathbf x)\exp(-\|\bF^\top\mathbf x\|_2^2/(2\sigma_e^2))$, which adds $\sigma_e^{-2}\bF\bF^\top$ to the precision and leaves the mean at zero.  Write $\bP_1\triangleq\bC\bH^{-1}\bC^\top\succeq0$ and $\bP_2\triangleq\sigma_e^{-2}\bF\bF^\top\succeq0$, so $\ker\bP_1=\range(\bP_1)^\perp$ and $\range(\bP_2)=\range(\bF)$.  If $\range(\bP_1)\not\subseteq\range(\bP_2)$, then $\range(\bP_2)^\perp\not\subseteq\ker\bP_1$, so some $\mathbf v\in\range(\bP_2)^\perp$ has $\mathbf v^\top\bL\mathbf v=\mathbf v^\top\bP_1\mathbf v>0$.  If $\range(\bP_2)\not\subseteq\range(\bP_1)$, then $\ker\bP_1\not\subseteq\ker\bP_2$, so some $\mathbf v\in\ker\bP_1$ has $\mathbf v^\top\bL\mathbf v=-\mathbf v^\top\bP_2\mathbf v<0$.
\end{proof}

\begin{proof}[Proof of Proposition~\ref{prop:preservation}]
The product of the $\mathcal N(\mathbf 0,\bSig^*)$ density and $u(\mathbf x;s)$ is proportional to $\exp(-\frac12\mathbf x^\top(\bOm^*+s^{-2}\bI)\mathbf x)$.  Its integral equals $\det(\bOm^*+s^{-2}\bI)^{-1/2}\det(\bOm^*)^{1/2}=\det(\bI+\bSig^*/s^2)^{-1/2}$.  The shift is diagonal.
\end{proof}

\begin{proof}[Proof of Proposition~\ref{prop:atomic}]
For any decomposition, the triangle inequality and $\|\bB\mathbf v\mathbf v^\top\bB\|_*=\|\bB\mathbf v\|_2^2$ give $\|\cD(\bL)\|_*\le\sum_k|c_k|\|\bB\mathbf v_k\|_2^2$.  Conversely, let $\cD(\bL)=\sum_{k=1}^p\mu_k\mathbf w_k\mathbf w_k^\top$ be an eigendecomposition and put $\mathbf v_k\triangleq\bB^{-1}\mathbf w_k/\|\bB^{-1}\mathbf w_k\|_2$ and $c_k\triangleq\mu_k\|\bB^{-1}\mathbf w_k\|_2^2$.  Then $\sum_kc_k\mathbf v_k\mathbf v_k^\top=\bB^{-1}\cD(\bL)\bB^{-1}=\bL$ and $\sum_k|c_k|\|\bB\mathbf v_k\|_2^2=\sum_k|\mu_k|=\|\cD(\bL)\|_*$.  The last identity follows from $\bB^2=\bI+\tau\bK^\top\bK$.
\end{proof}
\section{Deterministic Tools}
\label{app:tools}

This appendix collects the deterministic lemmas used in the proof of Theorem~\ref{thm:pdw}: the geometry of the congruence, an optimality-and-uniqueness certificate, and an explicit chart of the rank-$r$ variety.

\begin{lemma}[Congruence geometry]
\label{lem:congruence}
(i) For every $\bM\in\Sym^p$, $\bmin^2\|\bM\|_\#\le\|\cD(\bM)\|_\#\le\bmax^2\|\bM\|_\#$ for $\#\in\{F,2\}$.
(ii) $\cD(\mathbf T)=\Tt$ and $\cD(\Tt^\perp)=\mathbf T^\perp$.
(iii) For every $\bX\in\Sym^p$, $\|\PTt\cD(\bX)\|_F\le\varpi_B\|\bX\|_F$, where
\begin{align}
\varpi_B\triangleq\min\big\{\bmax^2,\ \sqrt2\,\bmax\|\bB\tilde\bU\|_2\big\}.
\label{eq:varpi}
\end{align}
\end{lemma}
\begin{proof}
(i) follows from submultiplicativity applied to $\cD$ and $\cD^{-1}$.  (ii) $\cD(\bU\bY^\top+\bY\bU^\top)=(\bB\bU)(\bB\bY)^\top+(\bB\bY)(\bB\bU)^\top\in\Tt$, and both spaces have dimension $pr-r(r-1)/2$, so $\cD(\mathbf T)=\Tt$.  For $\bY\in\Tt^\perp$ and $\bH\in\mathbf T$, $\langle\cD(\bY),\bH\rangle=\langle\bY,\cD(\bH)\rangle=0$, so $\cD(\Tt^\perp)\subseteq\mathbf T^\perp$, with equality by dimension.  (iii) The bound by $\bmax^2$ is (i).  Write $\PTt\bY=\bP_{\tilde U}\bY+(\bI-\bP_{\tilde U})\bY\bP_{\tilde U}$; the two terms are Frobenius-orthogonal.  With $\bY=\bB\bX\bB$, $\|\bP_{\tilde U}\bB\bX\bB\|_F\le\|\bP_{\tilde U}\bB\|_2\|\bX\|_F\bmax$ and $\|(\bI-\bP_{\tilde U})\bB\bX\bB\bP_{\tilde U}\|_F\le\bmax\|\bX\|_F\|\bB\bP_{\tilde U}\|_2$, and $\|\bP_{\tilde U}\bB\|_2=\|\bB\bP_{\tilde U}\|_2=\|\bB\tilde\bU\|_2$.
\end{proof}

The factor $\|\bB\tilde\bU\|_2$ measures the structural energy of the low-rank column space: if $\bK\bU=\mathbf 0$, then $\bB\bU=\bU$, $\tilde\bU=\bU$, and $\|\bB\tilde\bU\|_2=1$, so the cost of the prior drops from $\bmax^2$ to $\sqrt2\bmax$.

For a rank-$r$ matrix $\bN$, let $\bP_{\bN}$ be the projector onto $\range(\bN)$ and $\Tt_{\bN}$ its tangent space.

\begin{lemma}[Optimality and uniqueness certificate]
\label{lem:kkt}
Let $\bS^\natural\in\Sym^p$ and $\bN^\natural\in\Sym^p$ with $\rank(\bN^\natural)=r$, $\bOm^\natural\triangleq\bS^\natural-\cD^{-1}(\bN^\natural)\succ0$, $\bG^\natural\triangleq\bG(\bOm^\natural)$, $\bP^\natural\triangleq\bP_{\bN^\natural}$, and $\Tt^\natural\triangleq\Tt_{\bN^\natural}$.  Suppose:
\begin{enumerate}
\item[(K1)] $(\bS^\natural)_{ij}=0$ for $(i,j)\notin\mathcal R$; $(\bG^\natural)_{ii}=0$ for all $i$; and for $(i,j)\in\mathcal E$, $(\bS^\natural)_{ij}\ne0$ and $(\bG^\natural)_{ij}=-\lambda\,\sign((\bS^\natural)_{ij})$;
\item[(K2)] $|(\bG^\natural)_{ij}|<\lambda$ for all $i\ne j$ with $(i,j)\notin\mathcal R$;
\item[(K3)] $P_{\Tt^\natural}\big(\cD^{-1}(\bG^\natural)\big)=\gamma\sgn(\bN^\natural)$;
\item[(K4)] $\|P_{(\Tt^\natural)^\perp}(\cD^{-1}(\bG^\natural))\|_2<\gamma$;
\item[(K5)] $\mathcal R\cap\cD^{-1}(\Tt^\natural)=\{0\}$.
\end{enumerate}
Then $(\bS^\natural,\cD^{-1}(\bN^\natural))$ is the unique minimizer of \eqref{eq:estimator}.
\end{lemma}
\begin{proof}
\emph{Optimality.}  Let $\bZ\triangleq-\bG^\natural/\lambda$ off the diagonal and $\bZ_{ii}=0$.  By (K1)--(K2), $\bZ\in\partial\|\bS^\natural\|_{1,\mathrm{off}}$, the subdifferential with respect to $\langle\cdot,\cdot\rangle$ on $\Sym^p$, which consists of the symmetric $\bZ$ with zero diagonal, $Z_{ij}=\sign(S_{ij})$ if $S_{ij}\ne0$, and $|Z_{ij}|\le1$ otherwise.  Hence $\bG^\natural+\lambda\bZ=0$, where we used $(\bG^\natural)_{ii}=0$.  Next, $\bY\triangleq\cD^{-1}(\bG^\natural)/\gamma=\sgn(\bN^\natural)+\bY_\perp$ with $\bY_\perp\triangleq P_{(\Tt^\natural)^\perp}(\bY)$ by (K3), and $\bP^\natural\bY_\perp=0$, $\|\bY_\perp\|_2<1$ by (K4).  By the characterization of the subdifferential of spectral functions \cite{lewis1996convex}, $\partial\|\bN^\natural\|_*=\{\sgn(\bN^\natural)+\bY':\bP^\natural\bY'=0,\|\bY'\|_2\le1\}$, so $\bY\in\partial\|\bN^\natural\|_*$.  Since $\nabla_{\bS}f(\bS-\cD^{-1}\bN)=\bG$ and $\nabla_{\bN}f(\bS-\cD^{-1}\bN)=-\cD^{-1}(\bG)$, we have $0\in\partial F(\bS^\natural,\bN^\natural)$; $F$ is convex, so $(\bS^\natural,\bN^\natural)$ minimizes \eqref{eq:N_program}.

\emph{Uniqueness.}  Since $-\log\det$ is strictly convex on the positive definite cone, all minimizers share $\bOm^\natural$, hence $\bG^\natural$.  Let $(\bS',\bN')$ be a minimizer.  Fermat's rule gives $-\bG^\natural/\lambda\in\partial\|\bS'\|_{1,\mathrm{off}}$ off the diagonal and $\bY\in\partial\|\bN'\|_*$.  By (K2), $S'_{ij}=0$ for $(i,j)\notin\mathcal R$.  Since $\|\bY\|_2\le1$ and $\langle\bY,\bN'\rangle=\|\bN'\|_*$, writing $\bN'=\sum_k\mu_k\mathbf v_k\mathbf v_k^\top$ gives $\sum_k|\mu_k|=\sum_k\mu_k\mathbf v_k^\top\bY\mathbf v_k\le\sum_k|\mu_k||\mathbf v_k^\top\bY\mathbf v_k|\le\sum_k|\mu_k|$, so $|\mathbf v_k^\top\bY\mathbf v_k|=1$ whenever $\mu_k\ne0$.  The eigenvalues of $\bY$ are $\pm1$ on $\range(\bP^\natural)$ and lie in $(-1,1)$ on its complement; $\mathbf v^\top\bY\mathbf v$ is a convex combination of eigenvalues, and $|\mathbf v^\top\bY\mathbf v|=1$ forces $\mathbf v\in\range(\bP^\natural)$.  Hence $\range(\bN')\subseteq\range(\bN^\natural)$ and $\bN'-\bN^\natural\in\Tt^\natural$.  Also $\bS'-\bS^\natural\in\mathcal R$ and $\bS'-\bS^\natural=\cD^{-1}(\bN'-\bN^\natural)$, so (K5) gives $\bS'=\bS^\natural$ and $\bN'=\bN^\natural$.
\end{proof}

\begin{lemma}[Chart of the rank-$r$ variety]
\label{lem:chart}
For $\bY\in\Tt$ with $\|\bY\|_2\le\nu/4$, let $\bA_Y\triangleq\tilde\bU^\top\bY\tilde\bU$, $\bF_Y\triangleq\tilde\bU_\perp^\top\bY\tilde\bU$, $\bGamma_Y\triangleq\tilde\bGamma+\bA_Y$,
\begin{align}
\varphi(\bY)&\triangleq\tilde\bU_\perp\bF_Y\bGamma_Y^{-1}\bF_Y^\top\tilde\bU_\perp^\top\in\Tt^\perp,\notag\\
\bN(\bY)&\triangleq\bN_0+\bY+\varphi(\bY),
\label{eq:chart}
\end{align}
$\bP_Y\triangleq\bP_{\bN(\bY)}$, $\Tt_Y\triangleq\Tt_{\bN(\bY)}$, and $\theta_Y\triangleq\|\bP_Y-\bP_0\|_2$.  Then:
\begin{enumerate}
\item[(i)] $\bN(\bY)=\bV_Y\bGamma_Y\bV_Y^\top$ with $\bV_Y\triangleq\tilde\bU+\tilde\bU_\perp\bF_Y\bGamma_Y^{-1}$ of full column rank; $\rank\bN(\bY)=r$ and $\bN(\bY)$ has the inertia of $\bN_0$;
\item[(ii)] $\|\varphi(\bY)\|_2\le\frac{4}{3\nu}\|\bY\|_2^2$ and $\|\bN(\bY)-\bN_0\|_2\le\frac43\|\bY\|_2$;
\item[(iii)] $\theta_Y\le4\|\bY\|_2/\nu$;
\item[(iv)] for $\bX\in\Sym^p$ and $\#\in\{2,F\}$, $\|(P_{\Tt_Y}-\PTt)\bX\|_\#\le4\theta_Y\|\bX\|_\#$; for $\bX\in\Tt_Y$, $\|\PTtp\bX\|_2\le2\theta_Y\|\bX\|_2$; and $\|\PTt\sgn(\bN(\bY))\|_2\le1+\theta_Y^2$;
\item[(v)] $\bY\mapsto\varphi(\bY)$, $\bP_Y$, $P_{\Tt_Y}$, $\sgn(\bN(\bY))$ are continuous.
\end{enumerate}
\end{lemma}
\begin{proof}
(i) Since $\bY\in\Tt$, $\tilde\bU_\perp^\top\bY\tilde\bU_\perp=0$, so $\bY=\tilde\bU\bA_Y\tilde\bU^\top+\tilde\bU\bF_Y^\top\tilde\bU_\perp^\top+\tilde\bU_\perp\bF_Y\tilde\bU^\top$.  Weyl's inequality and $\|\bA_Y\|_2\le\|\bY\|_2\le\nu/4$ show that $\bGamma_Y$ is nonsingular with $\|\bGamma_Y^{-1}\|_2\le4/(3\nu)$ and has the inertia of $\tilde\bGamma$.  Expanding $\bV_Y\bGamma_Y\bV_Y^\top$ gives $\tilde\bU\bGamma_Y\tilde\bU^\top+\tilde\bU\bF_Y^\top\tilde\bU_\perp^\top+\tilde\bU_\perp\bF_Y\tilde\bU^\top+\tilde\bU_\perp\bF_Y\bGamma_Y^{-1}\bF_Y^\top\tilde\bU_\perp^\top=\bN(\bY)$.  $\bV_Y$ has full column rank since $\tilde\bU^\top\bV_Y=\bI_r$; Sylvester's law gives the inertia.
(ii) $\|\bF_Y\|_2\le\|\bY\|_2$, so $\|\varphi(\bY)\|_2\le\|\bY\|_2^2\|\bGamma_Y^{-1}\|_2$, and $\|\bN(\bY)-\bN_0\|_2\le\|\bY\|_2(1+\frac{4}{3\nu}\cdot\frac\nu4)$.
(iii) Let $\bE\triangleq\bN(\bY)-\bN_0$, so $\|\bE\|_2\le\nu/3$.  From $(\bI-\bP_Y)\bN(\bY)=0$ we get $(\bI-\bP_Y)\bN_0=-(\bI-\bP_Y)\bE$; multiplying by $\bN_0^\dagger$ and using $\bN_0\bN_0^\dagger=\bP_0$, $\|\bN_0^\dagger\|_2=1/\nu$ gives $\|(\bI-\bP_Y)\bP_0\|_2\le\|\bE\|_2/\nu$.  Similarly $(\bI-\bP_0)\bN(\bY)=(\bI-\bP_0)\bE$ gives $\|(\bI-\bP_0)\bP_Y\|_2\le\|\bE\|_2\|\bN(\bY)^\dagger\|_2$.  By (i) and Weyl's inequality applied to the sorted eigenvalues, the $r$ nonzero eigenvalues of $\bN(\bY)$ have modulus at least $\nu-\|\bE\|_2\ge2\nu/3$, so $\|\bN(\bY)^\dagger\|_2\le3/(2\nu)$.  Since $\bP_Y-\bP_0=\bP_Y(\bI-\bP_0)-(\bI-\bP_Y)\bP_0$ and $\|\bP_Y(\bI-\bP_0)\|_2=\|(\bI-\bP_0)\bP_Y\|_2$, $\theta_Y\le\frac52\|\bE\|_2/\nu\le\frac{10}{3}\|\bY\|_2/\nu$.
(iv) With $\bP=\bP_Y$, $\bP'=\bP_0$: $P_{\Tt_Y}\bX-\PTt\bX=(\bP-\bP')\bX+\bX(\bP-\bP')-(\bP-\bP')\bX\bP-\bP'\bX(\bP-\bP')$, each term bounded by $\theta_Y\|\bX\|_\#$.  For $\bX\in\Tt_Y$, $\bX=\bP\bX+(\bI-\bP)\bX\bP$, so $(\bI-\bP')\bX(\bI-\bP')=(\bI-\bP')\bP\bX(\bI-\bP')+(\bI-\bP')(\bI-\bP)\bX\bP(\bI-\bP')$, and both terms contain a factor of norm at most $\theta_Y$.  For $\bE_Y\triangleq\sgn(\bN(\bY))=\bP\bE_Y\bP$, $\|\bE_Y\|_2=1$ and $\PTt\bE_Y=\bE_Y-(\bI-\bP')\bP\bE_Y\bP(\bI-\bP')$, whose second term has norm at most $\theta_Y^2$.
(v) $\varphi$ is continuous because $\bGamma_Y^{-1}$ is.  With $\bC_Y\triangleq\bV_Y^\top\bV_Y\succ0$ and $\bM_Y\triangleq\bC_Y^{1/2}\bGamma_Y\bC_Y^{1/2}$ (nonsingular), $\bP_Y=\bV_Y\bC_Y^{-1}\bV_Y^\top$ and $\sgn(\bN(\bY))=\bV_Y\bC_Y^{-1/2}\bM_Y(\bM_Y^2)^{-1/2}\bC_Y^{-1/2}\bV_Y^\top$, all continuous; $P_{\Tt_Y}$ is continuous in $\bP_Y$.
\end{proof}

\section{Proof of the Witness Theorem and Its Consequences}
\label{app:pdw}

\begin{proof}[Proof of Theorem~\ref{thm:pdw}]
Let $\bZ_{\mathcal E}\in\mathcal R$ have entries $\sign((\bS_0)_{ij})$ on $\mathcal E$ and zeros elsewhere (in particular on the diagonal), and let $\mathbb B\triangleq\{\mathbf u=(\bDelta_S,\bY)\in\cX:\|\mathbf u\|_g\le\varrho\}$, a convex compact set.

\emph{Step 0: a priori bounds on $\mathbb B$.}  Let $\mathbf u=(\bDelta_S,\bY)\in\mathbb B$.  Then $\|\bY\|_2\le\varrho\gamma=\xi\nu\le\nu/32$, so Lemma~\ref{lem:chart} applies and $\theta\triangleq\theta_Y\le4\xi\le\frac18$.  Define
\[
\bDelta(\mathbf u)\triangleq\bDelta_S-\cD^{-1}(\bY+\varphi(\bY))=\cA\mathbf u-\cD^{-1}(\varphi(\bY)),
\]
the deviation of $\bOm(\mathbf u)\triangleq(\bS_0+\bDelta_S)-\cD^{-1}(\bN(\bY))$ from $\bOm_0$.  Since $\bDelta_S\in\mathcal R$, $\|\bDelta_S\|_2\le\mu_{\mathcal R}\varrho\lambda$, so $\|\cA\mathbf u\|_2\le\varrho h=\zeta\omega_-$.  By Lemma~\ref{lem:chart}(ii) and $\gamma/\bmin^2\le h$,
\begin{align}
\|\cD^{-1}(\varphi(\bY))\|_2\le\frac{4\varrho^2\gamma^2}{3\nu\bmin^2}=\frac43\,\xi\,\frac{\varrho\gamma}{\bmin^2}\le\frac43\xi\zeta\omega_-,
\label{eq:phi_small}
\end{align}
hence $\|\bDelta(\mathbf u)\|_2\le(1+\frac43\xi)\zeta\omega_-\le\frac{25}{24}\zeta\omega_-\le\frac{25}{96}\omega_-$.  Thus $\bOm(\mathbf u)\succeq\frac12\omega_-\bI$.  With $\cR(\bDelta)\triangleq(\bOm_0+\bDelta)^{-1}-\bSig_0+\bSig_0\bDelta\bSig_0=\bSig_0\bDelta\bSig_0\bDelta(\bOm_0+\bDelta)^{-1}$,
\begin{align}
\bG(\mathbf u)\triangleq\bG(\bOm(\mathbf u))=\bW+\cI(\bDelta(\mathbf u))-\cR(\bDelta(\mathbf u)).
\label{eq:G_expansion}
\end{align}
Using $\|\bSig_0\|_2=1/\omega_-$ and $\|\cI\bX\|_\infty\le\|\cI\bX\|_2\le\|\bX\|_2/\omega_-^2$, define and bound
\begin{align*}
n_1&\triangleq\|\cR(\bDelta(\mathbf u))\|_2\le\frac{2\|\bDelta(\mathbf u)\|_2^2}{\omega_-^3}\le\frac{2(25/24)^2\zeta^2}{\omega_-},\\
n_2&\triangleq\|\cI\cD^{-1}(\varphi(\bY))\|_2\le\frac43\frac{\xi\zeta}{\omega_-},\\
n_3&\triangleq\|\PTt(P_{\Tt_Y}-\PTt)\cD^{-1}\cI\cA\mathbf u\|_2\le2\cdot4\theta\cdot\frac{\varrho h}{\bmin^2\omega_-^2}\le\frac{32\xi\zeta}{\bmin^2\omega_-},
\end{align*}
so $n_1+n_2\le3\zeta(\zeta+\xi)/\omega_-$.  With $\iota\triangleq\max\{\lambda^{-1},4\bmin^{-2}\gamma^{-1}\}=\omega_-\Lambda$, set
\begin{align*}
T_B&\triangleq\iota(n_1+n_2)\le3\zeta(\zeta+\xi)\Lambda,\qquad T_C\triangleq n_3/\gamma\le8\xi\zeta\Lambda,\\
T_D&\triangleq\frac{4\theta}{\bmin^2\gamma}\Big(\frac{\varrho h}{\omega_-^2}+n_1+n_2\Big)\le4\xi\zeta\Lambda\big(1+3(\zeta+\xi)\big)\le8\xi\zeta\Lambda.
\end{align*}
Hence $\theta^2+T_B+T_C+T_D\le16\xi^2+\zeta(3\zeta+19\xi)\Lambda=\mathfrak n$.

\emph{Step 1: fixed-point equation.}  Define $\mathcal F:\mathbb B\to\cX$ by
\begin{align*}
\mathcal F(\mathbf u)\triangleq\Big(&\PR\bG(\mathbf u)+\lambda\bZ_{\mathcal E},\\
&\PTt P_{\Tt_Y}\big[\gamma\sgn(\bN(\bY))-\cD^{-1}(\bG(\mathbf u))\big]\Big).
\end{align*}
Since $\theta<1/4$, Lemma~\ref{lem:chart}(iv) gives $\|\PTt\bX\|_F\ge(1-4\theta)\|\bX\|_F>0$ for nonzero $\bX\in\Tt_Y$, so $\PTt$ is injective on $\Tt_Y$; as $\sgn(\bN(\bY))\in\Tt_Y$, the second component vanishes iff $P_{\Tt_Y}\cD^{-1}(\bG(\mathbf u))=\gamma\sgn(\bN(\bY))$.  Substituting \eqref{eq:G_expansion}, using $\cH\mathbf u=(\PR\cI\cA\mathbf u,-\PTt\cD^{-1}\cI\cA\mathbf u)$ and $\PTt P_{\Tt_Y}=\PTt+\PTt(P_{\Tt_Y}-\PTt)$, we obtain $\mathcal F(\mathbf u)=\cH\mathbf u+\mathbf q(\mathbf u)$ with $\mathbf q=(\mathbf q_1,\mathbf q_2)\in\cX$,
\begin{align*}
\mathbf q_1={}&\lambda\bZ_{\mathcal E}+\PR\bW-\PR\big[\cI\cD^{-1}\varphi(\bY)+\cR(\bDelta)\big],\\
\mathbf q_2={}&\gamma\PTt\sgn(\bN(\bY))-\PTt P_{\Tt_Y}\cD^{-1}\bW\\
&+\PTt P_{\Tt_Y}\cD^{-1}\big[\cI\cD^{-1}\varphi(\bY)+\cR(\bDelta)\big]\\
&-\PTt(P_{\Tt_Y}-\PTt)\cD^{-1}\cI\cA\mathbf u.
\end{align*}
By \eqref{eq:proj_facts}, $\|\PTt P_{\Tt_Y}\bX\|_2\le4\|\bX\|_2$, and $\|\cD^{-1}\bX\|_2\le\bmin^{-2}\|\bX\|_2$.  Hence, with Lemma~\ref{lem:chart}(iv),
\[
\frac{\|\mathbf q_1\|_\infty}{\lambda}\le1+\varepsilon_W+T_B,\
\frac{\|\mathbf q_2\|_2}{\gamma}\le1+\theta^2+4\varepsilon_W+T_B+T_C,
\]
so $\|\mathbf q(\mathbf u)\|_g\le1+4\varepsilon_W+\mathfrak n\le1+\frac\alpha3+\frac\alpha8<2$ by (C2)--(C3).  The map $\Psi(\mathbf u)\triangleq-\cH^{-1}\mathbf q(\mathbf u)$ is continuous on $\mathbb B$ by Lemma~\ref{lem:chart}(v) and continuity of $\bDelta\mapsto(\bOm_0+\bDelta)^{-1}$, and $\|\Psi(\mathbf u)\|_g\le\kappa_H\|\mathbf q(\mathbf u)\|_g\le2\kappa_H=\varrho$.  By Brouwer's fixed-point theorem there is $\mathbf u^\natural=(\bDelta_S^\natural,\bY^\natural)\in\mathbb B$ with $\Psi(\mathbf u^\natural)=\mathbf u^\natural$, i.e.\ $\mathcal F(\mathbf u^\natural)=0$.  Set $\bS^\natural\triangleq\bS_0+\bDelta_S^\natural$, $\bN^\natural\triangleq\bN(\bY^\natural)$, $\bG^\natural\triangleq\bG(\mathbf u^\natural)$, $\bP^\natural\triangleq\bP_{Y^\natural}$, $\Tt^\natural\triangleq\Tt_{Y^\natural}$, and $\theta\triangleq\theta_{Y^\natural}$.

\emph{Step 2: strict dual feasibility.}  Write $\mathbf e\triangleq\cA^*(\bW-\cI\cD^{-1}\varphi(\bY^\natural)-\cR(\bDelta(\mathbf u^\natural)))$, so that $\cA^*\bG^\natural=\cM\mathbf u^\natural+\mathbf e$ by \eqref{eq:G_expansion}.  Since $\mathbf u^\natural=-\cH^{-1}\mathbf q(\mathbf u^\natural)$, Assumption~\ref{ass:irrep} gives $\|P_{\cX^\perp}\cM\mathbf u^\natural\|_g\le(1-\alpha)\|\mathbf q(\mathbf u^\natural)\|_g\le(1-\alpha)(1+4\varepsilon_W+\mathfrak n)$.  The first component of $P_{\cX^\perp}\cA^*\bG^\natural$ is $\PRp\bG^\natural$, so
\begin{align*}
\frac{\|\PRp\bG^\natural\|_\infty}{\lambda}&\le\varepsilon_W+T_B+(1-\alpha)(1+4\varepsilon_W+\mathfrak n)\\
&\le1-\alpha+5\varepsilon_W+2\mathfrak n\le1-\frac\alpha3.
\end{align*}
The second component is $-\PTtp\cD^{-1}\bG^\natural$, and $\PTtp$ is a spectral contraction, so
\[
\|\PTtp\cD^{-1}\bG^\natural\|_2\le\gamma\big[\varepsilon_W+T_B+(1-\alpha)(1+4\varepsilon_W+\mathfrak n)\big].
\]
By Lemma~\ref{lem:chart}(iv), $\|(P_{(\Tt^\natural)^\perp}-\PTtp)\cD^{-1}\bG^\natural\|_2\le4\theta\|\cD^{-1}\bG^\natural\|_2$ and $\|\cD^{-1}\bG^\natural\|_2\le\|\cD^{-1}\bW\|_2+\bmin^{-2}(\varrho h/\omega_-^2+n_1+n_2)$, so this difference is at most $\gamma(4\theta\varepsilon_W+T_D)\le\gamma(\frac12\varepsilon_W+T_D)$.  Therefore
\[
\frac{\|P_{(\Tt^\natural)^\perp}\cD^{-1}\bG^\natural\|_2}{\gamma}\le1-\alpha+\tfrac{11}{2}\varepsilon_W+2\mathfrak n\le1-\frac{7\alpha}{24}.
\]
This proves (d) once we identify the minimizer below.

\emph{Step 3: certificate and uniqueness.}  We verify (K1)--(K5) of Lemma~\ref{lem:kkt} for $(\bS^\natural,\bN^\natural)$.  First, $\bOm^\natural=\bOm(\mathbf u^\natural)\succ0$ by Step~0, and $\rank\bN^\natural=r$ by Lemma~\ref{lem:chart}(i).  (K1): $\bS^\natural$ is supported on $\mathcal R$ because $\bS_0$ and $\bDelta_S^\natural$ are; $\|\bDelta_S^\natural\|_\infty\le\varrho\lambda=2\kappa_H\lambda$ and (C4) give $\sign((\bS^\natural)_{ij})=\sign((\bS_0)_{ij})\ne0$ on $\mathcal E$; and the first component of $\mathcal F(\mathbf u^\natural)=0$ is $\PR\bG^\natural=-\lambda\bZ_{\mathcal E}$, which states $(\bG^\natural)_{ii}=0$ and $(\bG^\natural)_{ij}=-\lambda\sign((\bS^\natural)_{ij})$ on $\mathcal E$.  (K3) is the second component of $\mathcal F(\mathbf u^\natural)=0$, as shown in Step~1.  (K2) and (K4) follow from Step~2.  (K5): let $(\bA,\bX)\in\mathcal R\times\Tt^\natural$ with $\bA=\cD^{-1}(\bX)$, and put $\mathbf u'=(\bA,\PTt\bX)\in\cX$.  Then $\cA\mathbf u'=\cD^{-1}(\PTtp\bX)$, and $\|\PTtp\bX\|_2\le2\theta\|\bX\|_2$ by Lemma~\ref{lem:chart}(iv).  Hence
\begin{align*}
\frac{(1-2\theta)\|\bX\|_2}{\gamma}&\le\frac{\|\PTt\bX\|_2}{\gamma}\le\|\mathbf u'\|_g\le\kappa_H\|\cH\mathbf u'\|_g\\
&\le\kappa_H\,\iota\,\frac{2\theta\|\bX\|_2}{\bmin^2\omega_-^2},
\end{align*}
because the components of $\cH\mathbf u'$ have $\|\PR\cI\cA\mathbf u'\|_\infty$ and $\frac12\|\PTt\cD^{-1}\cI\cA\mathbf u'\|_2\bmin^2$ both at most $\|\cA\mathbf u'\|_2/\omega_-^2$.  Since $\theta\le4\xi$ and $\zeta\Lambda=\varrho h\iota/\omega_-^2\ge2\kappa_H\gamma\iota/(\bmin^2\omega_-^2)$, the right side is at most $4\xi\zeta\Lambda\|\bX\|_2/\gamma\le\frac{4}{19}\mathfrak n\|\bX\|_2/\gamma<\frac34\|\bX\|_2/\gamma\le(1-2\theta)\|\bX\|_2/\gamma$, which forces $\bX=0$ and then $\bA=0$.  Lemma~\ref{lem:kkt} shows that $(\hat\bS,\hat\bL)=(\bS^\natural,\cD^{-1}(\bN^\natural))$ is the unique minimizer.

\emph{Step 4: conclusions.}  (a) follows from (K1).  (b): $\rank\hat\bL=\rank\bN^\natural=r$, and $\hat\bL=\cD^{-1}(\bN^\natural)$ has the inertia of $\bN^\natural$, which equals that of $\bN_0$ and hence of $\bL_0$.  (c): $\|\hat\bS-\bS_0\|_\infty=\|\bDelta_S^\natural\|_\infty\le2\kappa_H\lambda$, and $\|\bN^\natural-\bN_0\|_2\le\frac43\|\bY^\natural\|_2\le\frac43\varrho\gamma$ by Lemma~\ref{lem:chart}(ii).  The Frobenius bounds follow from $|\mathcal R|=m$, $\rank(\bN^\natural-\bN_0)\le2r$, and Lemma~\ref{lem:congruence}(i).
\end{proof}

\begin{proof}[Proof of Proposition~\ref{prop:identifiability}]
Apply Theorem~\ref{thm:pdw} (and Remark~\ref{rem:ravikumar} if a rank is zero)
deterministically with $\hat{\bm\Sigma}_u\triangleq\bm\Sigma_0$, so that $\bW=0$ for both
targets, and with $\gamma=\rho\lambda$. The constants $\kappa_H$ and $\alpha$
depend on $(\lambda,\gamma)$ only through $\rho$. As $\lambda\downarrow0$,
the quantities $\xi$ and $\zeta$ in~\eqref{eq:pdw_scalars} are $O(\lambda)$ and
$\Lambda=O(\lambda^{-1})$, so $n=O(\lambda)$; hence (C1)--(C4) hold for both
decompositions for all sufficiently small $\lambda$, (C2) trivially. The unique
minimizer $\hat \bS_\lambda$ then satisfies
$\|\hat \bS_\lambda-\bS_k\|_\infty\le2\kappa_H^{(k)}\lambda$ for $k=0,1$, so
$\|\bS_0-\bS_1\|_\infty\le2(\kappa_H^{(0)}+\kappa_H^{(1)})\lambda\to0$. Thus
$\bS_0=\bS_1$ and $\bL_0=\bS_0-\bm\Sigma_0^{-1}=\bL_1$.
\end{proof}

\begin{proof}[Proof of Corollary~\ref{cor:pdw_nominal}]
On the intersection of the events of Proposition~\ref{prop:noise}, each at level $\eta/2$, $\varepsilon_W\le\alpha/12$, which is (C2).  Apply Theorem~\ref{thm:pdw}.
\end{proof}

\section{Verification and Stability of the Witness Conditions}
\label{app:primitive}

\begin{proof}[Proof of Theorem~\ref{thm:primitive_2}]
\emph{Invariance.}  If $(\bI-\bP_U)\bB^2\bP_U=0$, then $\bB^2\bP_U=\bP_U\bB^2\bP_U$, and transposing gives $\bP_U\bB^2=\bP_U\bB^2\bP_U$; hence $\bB^2$, and therefore $\bB=(\bB^2)^{1/2}$, commutes with $\bP_U$.  The converse is immediate.  Consequently $\range(\bB\bU)=\range(\bU)$, so we may take $\tilde\bU=\bU$ and $\Tt=\mathbf T$.  By Lemma~\ref{lem:congruence}(ii), $\cD$ maps $\mathbf T$ and $\mathbf T^\perp$ onto themselves; being self-adjoint, $\cD$ and $\cD^{-1}$ commute with $\PT$.  Moreover $\|\cD\bX\|_2\le\bmax^2\|\bX\|_2$ and $\|\cD^{-1}\bX\|_2\le\bmin^{-2}\|\bX\|_2$.

\emph{Reduction.}  Write $\bm\Sigma_0=\sigma_0(\bI+\Delta)$ with $\|\Delta\|_{\infty\to\infty}=\delta_\infty$.  Then $\mathcal I=\sigma_0^2(\mathrm{Id}+\mathcal E)$ with $\mathcal E(\bX)\triangleq\Delta\bX+\bX\Delta+\Delta\bX\Delta$.  Since $\Delta$ is symmetric, $\|\Delta\|_2\le\|\Delta\|_{\infty\to\infty}$ and $\|\bX\Delta\|_\infty\le\|\bX\|_\infty\|\Delta\|_{1\to1}=\|\bX\|_\infty\delta_\infty$; hence, with $\varepsilon\triangleq2\delta_\infty+\delta_\infty^2\le1/(40\bmax^4)$,
\begin{equation}\label{eq:E-bounds}
\|\mathcal E \bX\|_\infty\le\varepsilon\|\bX\|_\infty,\qquad
\|\mathcal E \bX\|_2\le\varepsilon\|\bX\|_2\qquad (\bX\in\mathbb S^p).
\end{equation}
Accordingly $\mathcal M=\sigma_0^2(\mathcal M_0+\mathcal M_{\mathcal E})$ with $\mathcal M_0=\mathcal A^*\mathcal A$ and $\mathcal M_{\mathcal E}=\mathcal A^*\mathcal E\mathcal A$, and $\mathcal H=\sigma_0^2(\mathcal H_0+\mathcal H_{\mathcal E})$.  The operator $P_{\mathcal X^\perp}\mathcal M\mathcal H^{-1}$ is invariant under this scaling and $\|\mathcal H^{-1}\|_g$ scales as $\sigma_0^{-2}$, so we may take $\sigma_0=1$.

\emph{Isotropic operator.}  For $\mathbf u=(\bA,\bY)\in\mathcal X$ put $\bZ\triangleq\cD^{-1}\bY\in\mathbf T$.  Then $\mathcal A\mathbf u=\bA-\bZ$ and, since $\cD^{-1}\bZ\in\mathbf T$,
\begin{align*}
\mathcal H_0\mathbf u&=\big(\bA-P_{\mathcal R}\bZ,\ \cD^{-1}\bZ-\PT\cD^{-1}\bA\big),\\
P_{\mathcal X^\perp}\mathcal M_0\mathbf u&=\big(-P_{\mathcal R^\perp}\bZ,\ -\PTp\cD^{-1}\bA\big).
\end{align*}
Put $s_1\triangleq\mu_{\mathcal R}\lambda/\gamma$ and $s_2\triangleq\xi(\mathbf T)\gamma/\lambda$.  Then (P2) reads $s_1\le\bmin^2/4$ and $s_2\le1/(4\bmax^2)$, and $\beta\triangleq2\mu_{\mathcal R}\xi(\mathbf T)=2s_1s_2\le1/8$ because $\bmin\le\bmax$.  We use $\|\bA\|_2\le\mu_{\mathcal R}\|\bA\|_\infty$ for $\bA\in\mathcal R$, $\|\bX\|_\infty\le\xi(\mathbf T)\|\bX\|_2$ for $\bX\in\mathbf T$, and~\eqref{eq:proj_facts}.  Let $\mathbf q=(\mathbf q_1,\mathbf q_2)\in\mathcal X$ with $\|\mathbf q\|_g\le1$.  The equation $\mathcal H_0\mathbf u=\mathbf q$ is equivalent to $\bA=\mathbf q_1+P_{\mathcal R}\bZ$ and, after applying $\cD$, which commutes with $\PT$, to
\[
(\mathrm{Id}-\mathcal K)\bZ=\cD\mathbf q_2+\PT\mathbf q_1\quad\text{on }\mathbf T,\qquad \mathcal K\triangleq\PT P_{\mathcal R}.
\]
Since $\|\mathcal K\bZ\|_2\le2\mu_{\mathcal R}\|\bZ\|_\infty\le\beta\|\bZ\|_2$, a Neumann series gives a unique solution with $\|\bZ\|_2\le(\bmax^2\gamma+2\mu_{\mathcal R}\lambda)/(1-\beta)$.  Hence, with $\bY=\cD\bZ$,
\begin{equation}\label{eq:H0inv}
\begin{aligned}
\frac{\|\bY\|_2}{\gamma}&\le\frac{\bmax^2(\bmax^2+2s_1)}{1-\beta}\le\frac{12}{7}\bmax^4,\\
\frac{\|\bA\|_\infty}{\lambda}&\le 1+\frac{\bmax^2s_2+\beta}{1-\beta}\le\frac{10}{7},
\end{aligned}
\end{equation}
so $\mathcal H_0$ is invertible with $\|\mathcal H_0^{-1}\|_g\le\frac{12}{7}\bmax^4$.  Moreover $\|P_{\mathcal R^\perp}\bZ\|_\infty/\lambda\le s_2\|\bZ\|_2/\gamma\le(\bmax^2s_2+\beta)/(1-\beta)\le3/7$ and $\|\PTp\cD^{-1}\bA\|_2/\gamma\le\bmin^{-2}s_1\|\bA\|_\infty/\lambda\le5/14$, so $\mathcal J_0\triangleq P_{\mathcal X^\perp}\mathcal M_0\mathcal H_0^{-1}$ satisfies $\|\mathcal J_0\|_g\le3/7$.

\emph{Perturbation.}  We have $\mathcal H_{\mathcal E}\mathbf u=(P_{\mathcal R}\mathcal E(\bA-\bZ),\,-\PT\cD^{-1}\mathcal E(\bA-\bZ))$ and $P_{\mathcal X^\perp}\mathcal M_{\mathcal E}\mathbf u=(P_{\mathcal R^\perp}\mathcal E(\bA-\bZ),\,-\PTp\cD^{-1}\mathcal E(\bA-\bZ))$.  For $\|\mathbf u\|_g\le1$, $\|\bZ\|_2\le\gamma/\bmin^2$ and $\|\bZ\|_\infty\le\xi(\mathbf T)\gamma/\bmin^2$, so $\|\bA-\bZ\|_\infty\le\lambda(1+s_2/\bmin^2)\le\frac54\lambda$ and $\|\bA-\bZ\|_2\le\gamma(s_1+\bmin^{-2})\le\gamma(\bmin^2/4+\bmin^{-2})$.  By~\eqref{eq:E-bounds} and~\eqref{eq:proj_facts}, and since $\bmin\ge1$,
\begin{align*}
\|\mathcal H_{\mathcal E}\|_g&\le\max\Big\{\tfrac54\varepsilon,\ \big(\tfrac12+2\bmin^{-4}\big)\varepsilon\Big\}\le\tfrac52\varepsilon,\\
\|P_{\mathcal X^\perp}\mathcal M_{\mathcal E}\|_g&\le\tfrac54\varepsilon.
\end{align*}
Thus $\|\mathcal H_0^{-1}\mathcal H_{\mathcal E}\|_g\le\frac{30}{7}\bmax^4\varepsilon\le\frac{3}{28}$, so $\mathcal H=\mathcal H_0(\mathrm{Id}+\mathcal H_0^{-1}\mathcal H_{\mathcal E})$ is invertible and $\|\mathcal H^{-1}\|_g\le\frac{12}{7}\bmax^4\big/\big(1-\frac{3}{28}\big)=\frac{48}{25}\bmax^4<2\bmax^4$.  Finally, $\mathcal H^{-1}=\mathcal H_0^{-1}-\mathcal H_0^{-1}\mathcal H_{\mathcal E}\mathcal H^{-1}$ gives $P_{\mathcal X^\perp}\mathcal M\mathcal H^{-1}=\mathcal J_0-\mathcal J_0\mathcal H_{\mathcal E}\mathcal H^{-1}+P_{\mathcal X^\perp}\mathcal M_{\mathcal E}\mathcal H^{-1}$, whence
\begin{align*}
\|P_{\mathcal X^\perp}\mathcal M\mathcal H^{-1}\|_g
&\le\tfrac37+\tfrac37\cdot\tfrac52\varepsilon\cdot2\bmax^4+\tfrac54\varepsilon\cdot2\bmax^4\\
&=\tfrac37+\tfrac{65}{14}\bmax^4\varepsilon\le\tfrac{61}{112}<\tfrac35 .
\end{align*}
The window is nonempty iff $4\mu_{\mathcal R}/\bmin^2\le1/(4\bmax^2\xi(\mathbf T))$, and the last claim follows from $\xi(\mathbf T)\le2\max_i\|\bU^\top e_i\|_2$.
\end{proof}

\begin{proof}[Proof of Proposition~\ref{prop:window}]
Take $q=(0,q_2)$ with $q_2\in\mathcal T$, $\|q_2\|_2=\gamma$, and
$\|P_{\mathcal R^\perp}q_2\|_\infty=\xi_{\rm off}(\mathcal T)\gamma$. As in the
proof of Theorem~\ref{thm:primitive_2} (with $\bB=\bI$, so that $\bZ=\bY$),
$\mathcal H^{-1}q=\sigma_0^{-2}(P_{\mathcal R}\bY,\bY)$ with $\bY=q_2+\mathcal K\bY$
and $\|\bY\|_2\le\gamma/(1-\beta)\le2\gamma$. The first component of
$P_{\mathcal X^\perp}\mathcal M\mathcal H^{-1}q$ is $-P_{\mathcal R^\perp}\bY$,
and
$\|P_{\mathcal R^\perp}\bY\|_\infty\ge\xi_{\rm off}(\mathcal T)\gamma-\xi(\mathcal T)\|\mathcal K\bY\|_2
\ge(\xi_{\rm off}(\mathcal T)-2\beta\xi(\mathcal T))\gamma$.
Since $\|q\|_g=1$, Assumption~\ref{ass:irrep} gives~\eqref{eq:window-nec}.
For the lower bound on $\xi_{\rm off}(\mathcal T)$, fix $i$, let
$w\triangleq P_Ue_i/\|P_Ue_i\|_2\in\mathrm{range}(\bU)$, choose $j\neq i$ with
$(i,j)\notin\mathcal R$, and put $X\triangleq we_j^\top+e_jw^\top\in\mathcal T$.
Then $\|\bX\|_2\le2$ and $X_{ij}=w_i=\|\bU^\top e_i\|_2$. Finally,
$\sum_i\|\bU^\top e_i\|_2^2=r$.
\end{proof}

The constants of Theorem~\ref{thm:primitive_2} are attached to the isotropic reference $\sigma_0\bI$.  The next result shows that the witness conditions are stable, with an explicit radius, around \emph{any} reference covariance at which they are known to hold; its proof contains the perturbation step of Theorem~\ref{thm:primitive_2} as the special case $\bSig_{\rm ref}=\sigma_0\bI$.

\begin{theorem}[Quantitative transfer of the witness conditions]
\label{thm:transfer}
Fix $\bB$, the target $(\bS_0,\bL_0)$ with model space $\cX=\mathcal R\times\Tt$, and $(\lambda,\gamma)$.  Let $\bSig_{\rm ref}\succ0$, $\cI_{\rm ref}(\bX)\triangleq\bSig_{\rm ref}\bX\bSig_{\rm ref}$, $\cM_{\rm ref}\triangleq\cA^*\cI_{\rm ref}\cA$, and $\cH_{\rm ref}\triangleq P_{\cX}\cM_{\rm ref}|_{\cX}$.  Suppose that $\cH_{\rm ref}$ is invertible, and put $\kappa_{\rm ref}\triangleq\|\cH_{\rm ref}^{-1}\|_{g\to g}$ and $1-\alpha_{\rm ref}\triangleq\|P_{\cX^\perp}\cM_{\rm ref}\cH_{\rm ref}^{-1}\|_{g\to g}$, with $\alpha_{\rm ref}\in(0,1]$.  Let
\begin{align*}
\epsilon_H&\triangleq\|P_{\cX}(\cM-\cM_{\rm ref})|_{\cX}\|_{g\to g},\\
\epsilon_\perp&\triangleq\|P_{\cX^\perp}(\cM-\cM_{\rm ref})|_{\cX}\|_{g\to g}.
\end{align*}
\begin{enumerate}
\item[(i)] If $\kappa_{\rm ref}\epsilon_H<1$, then Assumption~\ref{ass:invert} holds at $\bSig_0$ with $\kappa_H\le\kappa_{\rm ref}/(1-\kappa_{\rm ref}\epsilon_H)$, and
\[
\|P_{\cX^\perp}\cM\cH^{-1}\|_{g\to g}\le(1-\alpha_{\rm ref})(1+\kappa_H\epsilon_H)+\kappa_H\epsilon_\perp .
\]
In particular, if $\kappa_{\rm ref}\epsilon_H\le\frac12$ and $\epsilon_H+\epsilon_\perp\le\alpha_{\rm ref}/(4\kappa_{\rm ref})$, then Assumptions~\ref{ass:invert}--\ref{ass:irrep} hold with $\kappa_H\le2\kappa_{\rm ref}$ and $\alpha\ge\alpha_{\rm ref}/2$.
\item[(ii)] With $\vartheta_\infty\triangleq\|\bSig_0-\bSig_{\rm ref}\|_{\infty\to\infty}(\|\bSig_0\|_{\infty\to\infty}+\|\bSig_{\rm ref}\|_{\infty\to\infty})$ and $\vartheta_2\triangleq\|\bSig_0-\bSig_{\rm ref}\|_2(\|\bSig_0\|_2+\|\bSig_{\rm ref}\|_2)$,
\begin{align}
\epsilon_H&\le\max\Big\{\Big(1+\frac{\xi(\mathbf T)\gamma}{\bmin^2\lambda}\Big)\vartheta_\infty,\ \frac{2}{\bmin^2}\Big(\frac{\mu_{\mathcal R}\lambda}{\gamma}+\frac1{\bmin^2}\Big)\vartheta_2\Big\},\notag\\
\epsilon_\perp&\le\max\Big\{\Big(1+\frac{\xi(\mathbf T)\gamma}{\bmin^2\lambda}\Big)\vartheta_\infty,\ \frac{1}{\bmin^2}\Big(\frac{\mu_{\mathcal R}\lambda}{\gamma}+\frac1{\bmin^2}\Big)\vartheta_2\Big\}.
\label{eq:transfer_eps}
\end{align}
\end{enumerate}
\end{theorem}
\begin{proof}
(i) Write $\cH=\cH_{\rm ref}+\bE_H$ with $\bE_H\triangleq P_{\cX}(\cM-\cM_{\rm ref})|_{\cX}$.  Since $\|\cH_{\rm ref}^{-1}\bE_H\|_{g\to g}\le\kappa_{\rm ref}\epsilon_H<1$, the operator $\cH=\cH_{\rm ref}(\mathrm{Id}+\cH_{\rm ref}^{-1}\bE_H)$ is invertible and $\|\cH^{-1}\|_{g\to g}\le\kappa_{\rm ref}/(1-\kappa_{\rm ref}\epsilon_H)$.  Next, $\cH_{\rm ref}\cH^{-1}=(\cH-\bE_H)\cH^{-1}=\mathrm{Id}-\bE_H\cH^{-1}$, so
\begin{align*}
P_{\cX^\perp}\cM\cH^{-1}={}&\big(P_{\cX^\perp}\cM_{\rm ref}\cH_{\rm ref}^{-1}\big)\big(\mathrm{Id}-\bE_H\cH^{-1}\big)\\
&+P_{\cX^\perp}(\cM-\cM_{\rm ref})\cH^{-1},
\end{align*}
and the bound follows by submultiplicativity.  Under the last two conditions, $\kappa_H\le2\kappa_{\rm ref}$ and $\|P_{\cX^\perp}\cM\cH^{-1}\|_{g\to g}\le1-\alpha_{\rm ref}+\kappa_H(\epsilon_H+\epsilon_\perp)\le1-\alpha_{\rm ref}/2$.

(ii) Let $\Delta\cI\triangleq\cI-\cI_{\rm ref}$ and $\bE\triangleq\bSig_0-\bSig_{\rm ref}$, so that $\Delta\cI(\bX)=\bE\bX\bSig_0+\bSig_{\rm ref}\bX\bE$.  For matrices $\bF,\bX,\bF'$, $|(\bF\bX\bF')_{ij}|\le\|\bX\|_\infty\sum_k|F_{ik}|\sum_l|F'_{lj}|$, and $\|\bF'\|_{1\to1}=\|\bF'\|_{\infty\to\infty}$ for symmetric $\bF'$; hence $\|\Delta\cI(\bX)\|_\infty\le\vartheta_\infty\|\bX\|_\infty$, and similarly $\|\Delta\cI(\bX)\|_2\le\vartheta_2\|\bX\|_2$.  Let $\mathbf u=(\bA,\bY)\in\cX$ with $\|\mathbf u\|_g\le1$.  By Lemma~\ref{lem:congruence}(ii), $\cD^{-1}\bY\in\mathbf T$, with $\|\cD^{-1}\bY\|_2\le\gamma/\bmin^2$ and $\|\cD^{-1}\bY\|_\infty\le\xi(\mathbf T)\gamma/\bmin^2$.  Therefore $\|\cA\mathbf u\|_\infty\le\lambda+\xi(\mathbf T)\gamma/\bmin^2$ and $\|\cA\mathbf u\|_2\le\mu_{\mathcal R}\lambda+\gamma/\bmin^2$.  Since $(\cM-\cM_{\rm ref})\mathbf u=(\Delta\cI\cA\mathbf u,\,-\cD^{-1}\Delta\cI\cA\mathbf u)$, the bounds \eqref{eq:transfer_eps} follow from $\|\PR\bX\|_\infty,\|\PRp\bX\|_\infty\le\|\bX\|_\infty$, $\|\PTt\bX\|_2\le2\|\bX\|_2$, $\|\PTtp\bX\|_2\le\|\bX\|_2$, and $\|\cD^{-1}\bX\|_2\le\bmin^{-2}\|\bX\|_2$.
\end{proof}

\begin{remark}[Local classes around a reference model]
\label{rem:local}
For $\bSig_{\rm ref}=\sigma_0\bI$ and $\bB$ as in Theorem~\ref{thm:primitive_2}, the isotropic step of its proof gives $\kappa_{\rm ref}\le\frac{12}{7}\bmax^4/\sigma_0^2$ and $\alpha_{\rm ref}\ge\frac47$, and $\vartheta_\infty,\vartheta_2\le\sigma_0^2\varepsilon$, so Theorem~\ref{thm:transfer} reproduces Theorem~\ref{thm:primitive_2} up to constants.  More generally, under a window of the form (P2) the prefactors in \eqref{eq:transfer_eps} are absolute constants, and Theorem~\ref{thm:transfer}(i) shows that the witness conditions hold on the set of $\bSig_0$ with $\max\{\vartheta_\infty,\vartheta_2\}\le c\,\alpha_{\rm ref}/\kappa_{\rm ref}$ for an absolute constant $c$.  Any reference at which $(\kappa_{\rm ref},\alpha_{\rm ref})$ are known therefore generates a common class on which Theorem~\ref{thm:pdw} applies uniformly.  The isotropic reference is used in Section~\ref{sec:lower} because it is the only one for which both the witness constants and the least-favorable packings are explicit.  Theorem~\ref{thm:transfer} also quantifies Proposition~\ref{prop:stability} with respect to $\bSig_0$; stability with respect to $\bB$ remains qualitative.
\end{remark}

\begin{remark}[Source of the excess sample size]
\label{rem:gap_source}
The source of the quadratic dependence on $p$ in \eqref{eq:C3_sample} can be located precisely.  In the proof of Theorem~\ref{thm:pdw}, the terms $T_C$ and $T_D$ are bounded through $\Lambda$ only for convenience: they are at most of order $\kappa_H^2\gamma/(\nu\omega_-^2)$ and impose a condition of the same type as (C1).  The excess comes from $T_B$, i.e., from the second-order remainder $\cR(\bDelta)$ and the curvature term $\cI\cD^{-1}\varphi(\bY)$, which enter the sparse block of the fixed-point map, are measured there in $\|\cdot\|_\infty$ relative to $\lambda$, and are bounded through their spectral norms ($n_1$ and $n_2$).  Both are quadratic in the off-diagonal block $\tilde\bU_\perp\bF_Y=(\bI-\bP_0)\bY\tilde\bU$ of the witness: $\varphi(\bY)=\tilde\bU_\perp\bF_Y\bGamma_Y^{-1}\bF_Y^\top\tilde\bU_\perp^\top$, and $\cR(\bDelta)$ contains the product of two such blocks.  If the rows of $(\bI-\bP_0)\bY\tilde\bU$ were bounded at the incoherence scale $\sqrt{\mu r/p}\,\|\bY\|_2$, the entrywise size of these products would drop by the factor $\mu r/p$; a heuristic count (which we do not claim as a theorem) then indicates that (C3) would be implied by (C1) up to constants and logarithmic factors.  Establishing such row-wise ($2\to\infty$) control for the random witness, for instance by leave-one-out arguments \cite{abbe2020entrywise}, is the open step.  By Theorem~\ref{thm:inertia_converse}, the resulting excess factor $p/(\mu r)$ in the sample size is not information-theoretic.
\end{remark}

\section{Concentration of the Centered Empirical MT Covariance}
\label{sec:conc}

For a generic law $P$, let $\mu_u\triangleq\E u(\mathbf x;s)$, $\mathbf a_u\triangleq\E[\mathbf xu]$, $\bV_u\triangleq\E[\mathbf x\mathbf x^\top u]$, $\mathbf m_u=\mathbf a_u/\mu_u$, and $\bM_u\triangleq\bV_u/\mu_u=\bSig_u+\mathbf m_u\mathbf m_u^\top$.  With $\hat{\mathbf a}_u\triangleq N^{-1}\sum_n\mathbf x_nu_n$ and $\hat\bV_u\triangleq N^{-1}\sum_n\mathbf x_n\mathbf x_n^\top u_n$, we have $\hat\bSig_u=\hat\bV_u/\hat\mu_u-\hat{\mathbf m}_u\hat{\mathbf m}_u^\top$.  We use Bernstein's inequality in the form: if $Y_1,\dots,Y_N$ are i.i.d.\ with $|Y_n-\E Y_n|\le b$ and $\mathrm{Var}(Y_n)\le v$, then for $t>0$,
\begin{align}
\Pr\Big(\big|\bar Y-\E Y_1\big|>\sqrt{2vt/N}+2bt/(3N)\Big)\le2e^{-t},
\label{eq:bernstein}
\end{align}
which follows from the classical tail bound $2\exp\{-s^2/(2(Nv+bs/3))\}$ for the sum, since $s=\sqrt{2Nvt}+2bt/3$ satisfies $s^2\ge2t(Nv+bs/3)$.  The matrix version \cite{tropp2012user} holds with $2e^{-t}$ replaced by $2d\,e^{-t}$ for $d\times d$ symmetric summands, $b$ an almost-sure bound on $\|Y_n-\E Y_n\|_2$, and $v\ge\|\E(Y_n-\E Y_n)^2\|_2$.

\begin{lemma}[Entrywise concentration]
\label{lem:conc_inf}
Fix $\eta\in(0,1)$, let $\ell_\mu\triangleq\log(6/\eta)$, $\ell_a\triangleq\log(6p/\eta)$, $\ell_V\triangleq\log(6p^2/\eta)$, and let $v_\mu\ge\E[u^2]$, $v_a\ge\max_i\E[x_i^2u^2]$, $v_V\ge\max_{i,j}\E[x_i^2x_j^2u^2]$.  Define
\begin{align}
t_\mu&\triangleq\sqrt{2v_\mu\ell_\mu/N}+\tfrac{2\ell_\mu}{3N},\notag\\
t_a&\triangleq\sqrt{2v_a\ell_a/N}+\tfrac{4R_1\ell_a}{3N},\notag\\
t_V&\triangleq\sqrt{2v_V\ell_V/N}+\tfrac{2R_2\ell_V}{3N},\notag\\
\delta_{m}&\triangleq\tfrac{2}{\mu_u}\big(t_a+\|\mathbf m_u\|_\infty t_\mu\big),\notag\\
\delta_{M}&\triangleq\tfrac{2}{\mu_u}\big(t_V+\|\bM_u\|_\infty t_\mu\big),\notag\\
\varepsilon_\infty(\eta)&\triangleq\delta_M+\big(2\|\mathbf m_u\|_\infty+\delta_m\big)\delta_m.
\label{eq:eps_inf}
\end{align}
If $t_\mu\le\mu_u/2$, then $\Pr(\|\hat\bSig_u-\bSig_u\|_\infty>\varepsilon_\infty(\eta))\le\eta$.  Distribution-free admissible choices are $v_\mu=\mu_u$, $v_a=R_2\mu_u$, and $v_V=R_2\mu_u\tr(\bM_u)/4$.
\end{lemma}
\begin{proof}
Since $0\le u\le1$, $|x_iu|\le R_1$, and $|x_ix_ju|\le\frac12\|\mathbf x\|_2^2u\le R_2/2$ by \eqref{eq:R1R2}, the centered summands are bounded by $1$, $2R_1$, and $R_2$, respectively, and their variances are bounded by the second moments.  Applying \eqref{eq:bernstein} with $t=\ell_\mu$ to $u$, with $t=\ell_a$ to each of the $p$ coordinates of $\mathbf xu$, and with $t=\ell_V$ to each of the at most $p^2$ entries of $\mathbf x\mathbf x^\top u$, a union bound gives, with probability at least $1-\eta$, $|\hat\mu_u-\mu_u|\le t_\mu$, $\|\hat{\mathbf a}_u-\mathbf a_u\|_\infty\le t_a$, and $\|\hat\bV_u-\bV_u\|_\infty\le t_V$.  On this event $\hat\mu_u\ge\mu_u/2$ and
\[
\frac{\hat{\mathbf a}_u}{\hat\mu_u}-\frac{\mathbf a_u}{\mu_u}=\frac{\hat{\mathbf a}_u-\mathbf a_u}{\hat\mu_u}+\frac{\mathbf a_u}{\mu_u}\cdot\frac{\mu_u-\hat\mu_u}{\hat\mu_u},
\]
so $\|\hat{\mathbf m}_u-\mathbf m_u\|_\infty\le\delta_m$; likewise $\|\hat\bV_u/\hat\mu_u-\bM_u\|_\infty\le\delta_M$.  Finally $\hat{\mathbf m}_u\hat{\mathbf m}_u^\top-\mathbf m_u\mathbf m_u^\top=(\hat{\mathbf m}_u-\mathbf m_u)\hat{\mathbf m}_u^\top+\mathbf m_u(\hat{\mathbf m}_u-\mathbf m_u)^\top$ has $\|\cdot\|_\infty\le(2\|\mathbf m_u\|_\infty+\delta_m)\delta_m$.  For the distribution-free choices: $\E u^2\le\E u$; $x_i^2u^2\le\|\mathbf x\|_2^2u\cdot u\le R_2u$; and $(x_ix_ju)^2\le\frac{R_2}{2}|x_ix_j|u\le\frac{R_2}4\|\mathbf x\|_2^2u$ with $\E[\|\mathbf x\|_2^2u]=\mu_u\tr\bM_u$.
\end{proof}

\begin{lemma}[Operator-norm concentration in a fixed geometry]
\label{lem:conc_op}
Let $\bJ\in\Sym^p$ be nonsingular and fixed.  Put $R_1^J\triangleq\|\bJ\|_2R_1$, $R_2^J\triangleq\|\bJ\|_2^2R_2$, $K_M^J\triangleq\|\bJ\bM_u\bJ\|_2$, $K_m^J\triangleq\|\bJ\mathbf m_u\|_2$, and let $v_V^J\ge\|\E[\|\bJ\mathbf x\|_2^2u^2\,\bJ\mathbf x\mathbf x^\top\bJ]\|_2$ and $v_a^J\ge\E[\|\bJ\mathbf x\|_2^2u^2]$; admissible distribution-free choices are $v_V^J=R_2^J\mu_uK_M^J$ and $v_a^J=R_2^J\mu_u$.  With $t_\mu$ as in Lemma~\ref{lem:conc_inf}, $\ell_2\triangleq\log(6p/\eta)$, $\ell_2'\triangleq\log(6(p+1)/\eta)$, define
\begin{align}
t_{V}^J&\triangleq\sqrt{2v_V^J\ell_2/N}+\tfrac{2R_2^J\ell_2}{3N},\notag\\
t_{a}^J&\triangleq\sqrt{2v_a^J\ell_2'/N}+\tfrac{4R_1^J\ell_2'}{3N},\notag\\
\delta_m^J&\triangleq\tfrac{2}{\mu_u}(t_a^J+K_m^Jt_\mu),\notag\\
\delta_M^J&\triangleq\tfrac{2}{\mu_u}(t_V^J+K_M^Jt_\mu),\notag\\
\varepsilon_2(\eta;\bJ)&\triangleq\delta_M^J+(2K_m^J+\delta_m^J)\delta_m^J.
\label{eq:eps2}
\end{align}
If $t_\mu\le\mu_u/2$, then $\Pr(\|\bJ(\hat\bSig_u-\bSig_u)\bJ\|_2>\varepsilon_2(\eta;\bJ))\le\eta$.
\end{lemma}
\begin{proof}
Let $\tilde{\mathbf x}=\bJ\mathbf x$, so $\bJ\hat\bSig_u\bJ=\bJ\hat\bV_u\bJ/\hat\mu_u-(\bJ\hat{\mathbf m}_u)(\bJ\hat{\mathbf m}_u)^\top$, where $\bJ\hat\bV_u\bJ$ averages $\bZ_n\triangleq\tilde{\mathbf x}_n\tilde{\mathbf x}_n^\top u_n$ and $\bJ\hat{\mathbf a}_u$ averages $\tilde{\mathbf x}_nu_n$, with weights depending on $\mathbf x_n$.  Since $\bZ_n\succeq0$, $\E\bZ_n\succeq0$, and $\|\bZ_n\|_2=\|\tilde{\mathbf x}_n\|_2^2u_n\le R_2^J$, we have $\|\bZ_n-\E\bZ_n\|_2\le R_2^J$, and $\E(\bZ_n-\E\bZ_n)^2\preceq\E\bZ_n^2=\E[\|\tilde{\mathbf x}\|_2^2u^2\tilde{\mathbf x}\tilde{\mathbf x}^\top]$.  Moreover $\bZ_n^2=(\|\tilde{\mathbf x}_n\|_2^2u_n)\bZ_n\preceq R_2^J\bZ_n$, so $\|\E\bZ_n^2\|_2\le R_2^J\mu_uK_M^J$.  The matrix Bernstein inequality with $t=\ell_2$ gives $\|\bJ(\hat\bV_u-\bV_u)\bJ\|_2\le t_V^J$ with probability at least $1-\eta/3$.  For $\mathbf z_n=\tilde{\mathbf x}_nu_n$, apply the matrix Bernstein inequality to the $(p+1)\times(p+1)$ self-adjoint dilations $\begin{psmallmatrix}0&(\mathbf z_n-\E\mathbf z_n)^\top\\ \mathbf z_n-\E\mathbf z_n&0\end{psmallmatrix}$, whose norm is $\|\mathbf z_n-\E\mathbf z_n\|_2\le2R_1^J$ and whose squared expectation has norm at most $\E\|\mathbf z_n\|_2^2\le v_a^J$; with $t=\ell_2'$ this gives $\|\bJ(\hat{\mathbf a}_u-\mathbf a_u)\|_2\le t_a^J$ with probability at least $1-\eta/3$.  The bound $|\hat\mu_u-\mu_u|\le t_\mu$ holds with probability at least $1-\eta/3$.  On the intersection, the ratio and outer-product perturbation steps of Lemma~\ref{lem:conc_inf}, carried out in $\|\cdot\|_2$, give the claim.  The distribution-free bound $v_a^J=R_2^J\mu_u$ follows from $\|\tilde{\mathbf x}\|_2^2u^2\le R_2^Ju$.
\end{proof}

Taking $\bJ=\bI$ controls $\|\bW\|_2$; taking $\bJ=\bB^{-1}$ controls the dual noise $\|\cD^{-1}(\bW)\|_2$ in \eqref{eq:dual_noise} directly, rather than through the cruder $\bmin^{-2}\|\bW\|_2$.

\begin{proposition}[Nominal Gaussian moment calculus]
\label{prop:moments}
Let $\mathbf x\sim\mathcal N(\mathbf 0,\bSig^*)$, $\bSig'\triangleq(\bOm^*+2s^{-2}\bI)^{-1}$, $\mu_2\triangleq\E u^2=\det(\bI+2\bSig^*/s^2)^{-1/2}$, $\bar\sigma\triangleq\max_i(\bSig_u^*)_{ii}$, and $\bG_J\triangleq\bJ\bSig_u^*\bJ$.  Then:
\begin{enumerate}
\item[(a)] $\mu_0\ge\exp(-\tr\bSig^*/(2s^2))$ and $\mu_2/\mu_0^2\le\exp(\|\bSig^*\|_F^2/(2s^4))$;
\item[(b)] $\mathbf m_u=\mathbf 0$, $\bM_u=\bSig_u^*$, and $v_\mu=\mu_2$, $v_a=\mu_2\bar\sigma$, $v_V=3\mu_2\bar\sigma^2$ are admissible in Lemma~\ref{lem:conc_inf};
\item[(c)] $K_m^J=0$, $K_M^J=\|\bG_J\|_2$, and $v_V^J=\mu_2(\tr\bG_J+2\|\bG_J\|_2)\|\bG_J\|_2$, $v_a^J=\mu_2\tr\bG_J$ are admissible in Lemma~\ref{lem:conc_op}.
\end{enumerate}
\end{proposition}
\begin{proof}
Since $u(\mathbf x;s)^2=u(\mathbf x;s/\sqrt2)$, Proposition~\ref{prop:preservation} gives $\E[u^2g(\mathbf x)]=\mu_2\E_{\mathbf y\sim\mathcal N(\mathbf 0,\bSig')}g(\mathbf y)$ for integrable $g$, and $\mu_2=\det(\bI+2\bSig^*/s^2)^{-1/2}$.  Moreover $\bOm^*+2s^{-2}\bI\succeq\bOm^*+s^{-2}\bI$ implies $\bSig'\preceq\bSig_u^*$.  (a) With $a_k=\lambda_k(\bSig^*)/s^2$, $\log\mu_0=-\frac12\sum_k\log(1+a_k)\ge-\frac12\sum_ka_k$, and $\mu_2/\mu_0^2=\prod_k(1+a_k)(1+2a_k)^{-1/2}$, where $(1+a)^2/(1+2a)=1+a^2/(1+2a)\le e^{a^2}$.  (b) Symmetry gives $\mathbf m_u=\mathbf 0$, and Proposition~\ref{prop:preservation} gives $\bM_u=\bSig_u^*$.  By Isserlis' theorem, $\E[y_i^2y_j^2]=\Sigma'_{ii}\Sigma'_{jj}+2(\Sigma'_{ij})^2\le3\max_k(\Sigma'_{kk})^2\le3\bar\sigma^2$, $\E[y_i^2]=\Sigma'_{ii}\le\bar\sigma$, and $\E[u^2]=\mu_2$.  (c) For $\mathbf w=\bJ\mathbf y\sim\mathcal N(\mathbf 0,\bG')$ with $\bG'=\bJ\bSig'\bJ\preceq\bG_J$, Isserlis' theorem gives $\E[\|\mathbf w\|_2^2\mathbf w\mathbf w^\top]=\tr(\bG')\bG'+2\bG'^2$, whose norm is at most $(\tr\bG_J+2\|\bG_J\|_2)\|\bG_J\|_2$; also $\E\|\mathbf w\|_2^2=\tr\bG'\le\tr\bG_J$.
\end{proof}

\begin{corollary}[Explicit nominal noise levels]
\label{cor:explicit}
Under the nominal model, let $\ell_\mu,\ell_a,\ell_V,\ell_2,\ell_2'$ be as above, $\bG_B\triangleq\bB^{-1}\bSig_u^*\bB^{-1}$, and $\mathsf r_B\triangleq\tr\bG_B/\|\bG_B\|_2$.  If $t_\mu\le\mu_0/2$, then, with probability at least $1-\eta$ each,
\begin{align}
\|\bW\|_\infty&\le\varepsilon_\infty\triangleq\tfrac{2}{\mu_0}\big(t_V+\bar\sigma t_\mu\big)+\tfrac{4}{\mu_0^2}t_a^2,\label{eq:eps_inf_nom}\\
\|\cD^{-1}(\bW)\|_2&\le\varepsilon_{2,B}\triangleq\tfrac{2}{\mu_0}\big(t_V^B+\|\bG_B\|_2t_\mu\big)+\tfrac{4}{\mu_0^2}(t_a^B)^2,\label{eq:eps2_nom}
\end{align}
where the $t$'s are computed with the admissible proxies of Proposition~\ref{prop:moments} and $\bJ=\bB^{-1}$.  In particular, the leading terms are
\begin{align*}
\tfrac{2}{\mu_0}\sqrt{\tfrac{2v_V\ell_V}{N}}&\le2\sqrt6\,e^{\|\bSig^*\|_F^2/(4s^4)}\bar\sigma\sqrt{\tfrac{\ell_V}{N}},\\
\tfrac{2}{\mu_0}\sqrt{\tfrac{2v_V^B\ell_2}{N}}&\le2\sqrt2\,e^{\|\bSig^*\|_F^2/(4s^4)}\|\bG_B\|_2\sqrt{\tfrac{(\mathsf r_B+2)\ell_2}{N}}.
\end{align*}
\end{corollary}
\begin{proof}
Combine Lemmas~\ref{lem:conc_inf}--\ref{lem:conc_op} with Proposition~\ref{prop:moments}, noting $\mathbf m_u=\mathbf 0$ and $\|\bM_u\|_\infty\le\bar\sigma$ (by $|\Sigma_{ij}|\le(\Sigma_{ii}\Sigma_{jj})^{1/2}$).  The leading-term bounds use $\mu_2/\mu_0^2\le e^{\|\bSig^*\|_F^2/(2s^4)}$.
\end{proof}

\begin{remark}[Explicit dependence on $p$, $s$, and $\bB$]
\label{rem:scale}
If $s^2\ge\tr\bSig^*$, then $\|\bSig^*\|_F\le\tr\bSig^*\le s^2$, so $\mu_0\ge e^{-1/2}$ and $\mu_2/\mu_0^2\le e^{1/2}$.  Since $\mu_2\le1$, $R_1\le s$, and $R_2\le s^2$, \eqref{eq:eps_inf_nom} then yields, for an absolute constant $C$,
\begin{align*}
\varepsilon_\infty&\le C\Big(\bar\sigma\sqrt{\tfrac{\ell_V}{N}}+\tfrac{s^2\ell_V}{N}+\tfrac{\bar\sigma\ell_a}{N}+\tfrac{s^2\ell_a^2}{N^2}\Big),\\
\varepsilon_{2,B}&\le C\Big(\|\bG_B\|_2\sqrt{\tfrac{(\mathsf r_B+2)\ell_2'}{N}}+\tfrac{s^2\ell_2'}{\bmin^2N}\\
&\qquad\quad+\tfrac{\tr\bG_B\,\ell_2'}{N}+\tfrac{s^2\ell_2'^2}{\bmin^2N^2}\Big).
\end{align*}
The first terms are the usual parametric rates; the $s^2/N$ terms are the price of the bounded, redescending weight.  If instead $s$ is fixed while $p\to\infty$, then $\mu_0$ can decay like $e^{-\tr\bSig^*/(2s^2)}$, and the condition $t_\mu\le\mu_0/2$ requires $N$ to grow accordingly.  The $\bB$-whitened effective rank $\mathsf r_B=\tr(\bB^{-1}\bSig_u^*\bB^{-1})/\|\bB^{-1}\bSig_u^*\bB^{-1}\|_2$ can be much smaller than $\tr\bSig_u^*/\|\bSig_u^*\|_2$ when $\bB$ damps many directions; this is the mechanism by which the structural prior lowers the low-rank regularization level $\gamma$.  Proposition~\ref{prop:rB} quantifies this reduction.
\end{remark}

\begin{proof}[Proof of Proposition~\ref{prop:noise}]
The first statement is Corollary~\ref{cor:explicit}, and the second is Remark~\ref{rem:scale}.
\end{proof}

\begin{proof}[Proof of Proposition~\ref{prop:rB}]
Since $\bB^2=\bI+\tau\bK^\top\bK$, the eigenvalues of $\bB^{-2}$ are $(1+\tau\varsigma_k)^{-1}$, and $\|\bB^{-2}\|_2=\bmin^{-2}$.  From $\sigma_-\bB^{-2}\preceq\bG_B\preceq\sigma_+\bB^{-2}$ we obtain $\sigma_-\bmin^{-2}\le\|\bG_B\|_2\le\sigma_+\bmin^{-2}$ and $\tr\bG_B=\tr(\bSig_u^*\bB^{-2})\le\sigma_+\tr(\bB^{-2})$.  (i) $\|\bG_B\|_2^2(\mathsf r_B+2)=\|\bG_B\|_2\tr\bG_B+2\|\bG_B\|_2^2\le\sigma_+^2(\bmin^{-2}\tr(\bB^{-2})+2\bmin^{-4})$; for $\bB=\bI$, $\|\bG_I\|_2\tr\bG_I\ge\sigma_-^2p$.  (ii) follows from the same bounds.  (iii) The matrix $\bK_{\mathrm{diff}}^\top\bK_{\mathrm{diff}}$ is the path-graph Laplacian, whose eigenvalues are $4\sin^2(\pi k/(2p))$, $k=0,\dots,p-1$; hence $\varsigma_1=0$ and $\bmin=1$.  Since $\sin x\ge2x/\pi$ on $[0,\pi/2]$, the $k$th term of $\tr(\bB^{-2})$ is at most $f(k)\triangleq(1+4\tau k^2/p^2)^{-1}$, and since $f$ is decreasing, $\sum_{k=0}^{p-1}f(k)\le1+\int_0^\infty f(x)\,dx=1+\pi p/(4\sqrt\tau)$.  The consequence follows from (i) and the leading terms in Corollary~\ref{cor:explicit}.
\end{proof}

\section{Proofs of the Information-Theoretic Limits}
\label{app:lower}

\begin{remark}[Size and locality of $\mathcal F$]
\label{rem:class_size}
Condition (F3) restricts $\bOm=\bS-\bL$ to a neighborhood of radius $1/200$ of the identity in $\|\cdot\|_{\infty\to\infty}$, which is a strong form of near-isotropy.  Three points delimit its role.  (i) The normalization at $\bI$ is without loss of generality: the Gaussian model is equivariant under $\mathbf x\mapsto c\mathbf x$, which maps $\mathcal F$ onto the class in which (F3) reads $\|\bS-\bL-c^{-2}\bI\|_{\infty\to\infty}\le c^{-2}/200$ and rescales the Frobenius loss by $c^{-2}$; what (F3) imposes is \emph{relative} near-isotropy.  (ii) Lower bounds are local by nature: a lower bound over a subclass is a lower bound over every model containing it, and the least-favorable configurations of Theorems~\ref{thm:lower}--\ref{thm:inertia_converse} lie near the identity.  Restricting the class therefore weakens only the upper-bound side of the comparison.  (iii) The radius is the price of verifying Assumptions~\ref{ass:invert}--\ref{ass:irrep} primitively in the mixed norm \eqref{eq:gnorm}, which requires the Fisher operator to be controlled simultaneously in $\|\cdot\|_\infty$ and $\|\cdot\|_2$; $\|\cdot\|_{\infty\to\infty}$ is the natural norm that does both.  By Theorem~\ref{thm:transfer} and Remark~\ref{rem:local}, a common class can be centered at any reference model with known witness constants.  We do not claim that near-isotropy is necessary for the witness conditions: for $r=0$ they reduce to the irrepresentability condition of \cite{ravikumar2011high}, which holds far beyond near-isotropic models.
\end{remark}

\begin{proof}[Proof of Lemma~\ref{lem:class}]
Since $\bS-\bL-\bI$ is symmetric, $\|\bS-\bL-\bI\|_2\le1/200$, which gives the eigenvalue
bound. The bound on $\xi(\mathcal T)$ follows from (F2) and
$\xi(\mathcal T)\le2\max_i\|\bU^\top e_i\|_2$. Put
$\bG\triangleq(\bS-\bL-\bI)/(1+s^{-2})$, so that $\|\bG\|_{\infty\to\infty}\le1/200$ and
$\bm\Sigma_0/\sigma_0=(\bI+\bG)^{-1}$. Then
$\|\bm\Sigma_0/\sigma_0-\bI\|_{\infty\to\infty}=\|\bG(\bI+\bG)^{-1}\|_{\infty\to\infty}\le\frac{1/200}{1-1/200}<\frac1{100}$.
Finally, $1/(4\bar\xi)\le1/(4\xi(\mathcal T))$, and the window is nonempty
because $16\mu_{\mathcal R}\bar\xi\le1$. Apply Theorem~\ref{thm:primitive_2}.
\end{proof}

\begin{lemma}[Local Kullback--Leibler bound]
\label{lem:KL}
If $\|\bOm_j-\bI\|_2\le1/4$ for $j=1,2$, then $\mathrm{KL}\big(\mathcal N(\mathbf 0,\bOm_1^{-1})^{\otimes N}\,\|\,\mathcal N(\mathbf 0,\bOm_2^{-1})^{\otimes N}\big)\le N\|\bOm_1-\bOm_2\|_F^2$.
\end{lemma}
\begin{proof}
For one observation, $2\,\mathrm{KL}=\tr(\bM)-p-\log\det\bM$ with $\bM=\bOm_1^{-1/2}\bOm_2\bOm_1^{-1/2}$, whose eigenvalues lie in $[3/5,5/3]$.  Since $x-1-\log x=(x-1)^2g(x)$ with $g(x)=\int_0^1t\,(1+t(x-1))^{-1}dt$ decreasing, $x-1-\log x\le g(3/5)(x-1)^2\le0.7(x-1)^2$ on that interval.  Hence $\mathrm{KL}\le0.35\|\bM-\bI\|_F^2\le0.35\|\bOm_1^{-1}\|_2^2\|\bOm_2-\bOm_1\|_F^2\le0.35\cdot\frac{16}{9}\|\bOm_2-\bOm_1\|_F^2$.  Tensorize.
\end{proof}

\begin{proof}[Proof of Theorem~\ref{thm:lower}]
\emph{Assouad in probability.} Let $\{P_\theta:\theta\in\{\pm1\}^M\}$ satisfy
$\mathrm{KL}(P_\theta\|P_{\theta'})\le1/2$ whenever
$\mathrm{Ham}(\theta,\theta')=1$. By~\cite[Thm.~2.12]{tsybakov2009introduction}, every
$\hat\theta$ admits $\theta$ with $\mathbb E_\theta\mathrm{Ham}(\hat\theta,\theta)\ge M/4$.
Since $\mathrm{Ham}\le M$, we have
$\mathbb E\,\mathrm{Ham}\le\frac M8+\frac{7M}8\Pr(\mathrm{Ham}\ge\frac M8)$, so
$\Pr_\theta(\mathrm{Ham}(\hat\theta,\theta)\ge M/8)\ge1/7$. Suppose further that
$\theta\mapsto X_\theta$ satisfies $\|X_\theta-X_{\theta'}\|_F^2\ge c_1\mathrm{Ham}(\theta,\theta')$.
Given $\tilde X$, let $\hat\theta\in\arg\min_\theta\|\tilde X-X_\theta\|_F$; then
$\|\tilde X-X_\theta\|_F\ge\frac12\|X_{\hat\theta}-X_\theta\|_F$, so
\begin{equation}\label{eq:assouad-prob}
\Pr_\theta\big(\|\tilde X-X_\theta\|_F\ge\tfrac12\sqrt{c_1M/8}\big)\ge\tfrac17 .
\end{equation}
All parameters below satisfy (F3), hence $\|\bm\Omega-\bI\|_2\le1/4$, and
Lemma~\ref{lem:KL} bounds the Kullback--Leibler divergence.

\emph{Auxiliary low-rank basis.} Let $p_0\triangleq\lfloor p/2\rfloor\ge p/3$,
$\mathcal I_0\triangleq[p_0]$, and $\mathcal I_1\triangleq[p]\setminus\mathcal I_0$,
so that $|\mathcal I_1|\ge p/2$. Let $\bV_0\in\mathbb R^{p\times r}$ vanish outside
$\mathcal I_0$, with columns given by the DCT-II vectors
$(\bV_0)_{ik}=\sqrt{2/p_0}\cos(\pi(2i-1)k/(2p_0))$, $i\in\mathcal I_0$,
$k\in[r]$. Then $\bV_0^\top \bV_0=\bI_r$ and $\|\bV_0^\top e_i\|_2^2\le2r/p_0\le6r/p$,
so $|(\bV_0\bV_0^\top)_{ij}|\le 2r/p_0$ and $\|\bV_0\bV_0^\top\|_{\infty\to\infty}\le2r$; by Cauchy--Schwarz, the same bounds hold for $\bV_0\bm\Pi_\pi\bV_0^\top$, where $\bm\Pi_\pi\triangleq\diag(\bI_{\pi_+},-\bI_{\pi_-})$.
Write $\nu\triangleq\nu_{\min}$.

\emph{Sparse part.} Let $\mathsf G$ be the disjoint union of
$\lfloor p/(d+1)\rfloor$ cliques on $d+1$ vertices; it has $M\ge pd/4$ edges.
Let $a\triangleq1/(4\sqrt N)$ and $b_0\triangleq\beta_{\min}+a$. For
$\theta\in\{\pm1\}^M$ put $S_\theta\triangleq I+\sum_e(b_0+a\theta_e)F_e$, where $F_e\triangleq e_ie_j^\top+e_je_i^\top$ for $e=\{i,j\}$, and
$\bL_{\rm b}\triangleq\nu \bV_0\bm\Pi_\pi\bV_0^\top$, so that $\In(\bL_{\rm b})=(\pi_+,\pi_-)$. Then (F1) holds, (F2) holds with $\bU=\bV_0$,
and $\|\bS_\theta-\bL_{\rm b}-\bI\|_{\infty\to\infty}\le d(\beta_{\min}+2a)+2\nu r
\le\frac1{400}+\frac1{600}<\frac1{200}$ by~\eqref{eq:lb-cond}; hence
$(\bS_\theta,\bL_{\rm b})\in\mathcal F^{\pi_+,\pi_-}$. Moreover
$\|\bS_\theta-\bS_{\theta'}\|_F^2=8a^2\mathrm{Ham}(\theta,\theta')$, so neighbors have
$\mathrm{KL}\le8Na^2=1/2$. By~\eqref{eq:assouad-prob} with $c_1=8a^2$, there is
$\theta$ with $\|\tilde S-S_\theta\|_F\ge\frac a2\sqrt M\ge\frac1{16}\sqrt{pd/N}$
with probability at least $1/7$.

\emph{Low-rank part.} Let $a'\triangleq p/(32N)$ and, for
$\Theta\in\{\pm1\}^{|\mathcal I_1|\times r}$, let $\bX_\Theta\in\mathbb R^{p\times r}$
equal $\sqrt{a'/p}\,\Theta$ on the rows $\mathcal I_1$ and vanish on
$\mathcal I_0$. Define
\[
\bL_\Theta\triangleq\nu\,\bZ_\Theta\bm\Pi_\pi\bZ_\Theta^\top,\qquad \bZ_\Theta\triangleq \bV_0+\bX_\Theta/\nu .
\]
Since $\bV_0^\top \bX_\Theta=0$, $\bZ_\Theta^\top \bZ_\Theta=\bI+\bX_\Theta^\top \bX_\Theta/\nu^2\succeq \bI$,
and, with $\mathbf C_\Theta\triangleq(\bZ_\Theta^\top\bZ_\Theta)^{1/2}$, $\bL_\Theta=\nu\,\bU_\Theta\mathbf C_\Theta\bm\Pi_\pi\mathbf C_\Theta\bU_\Theta^\top$ for the orthonormal $\bU_\Theta$ defined below.  By Ostrowski's theorem \cite[Thm.~4.5.9]{horn2012matrix}, the eigenvalues of $\mathbf C_\Theta\bm\Pi_\pi\mathbf C_\Theta$ are $\theta_k\lambda_k(\bm\Pi_\pi)$ with $\theta_k\in[\lambda_{\min}(\mathbf C_\Theta^2),\lambda_{\max}(\mathbf C_\Theta^2)]\subset[1,\infty)$.  Hence $\mathrm{rank}\,\bL_\Theta=r$, $\In(\bL_\Theta)=(\pi_+,\pi_-)$, and its nonzero eigenvalues have modulus at least $\nu$.
With $\bU_\Theta=\bZ_\Theta(\bZ_\Theta^\top \bZ_\Theta)^{-1/2}$ we have
$\|\bU_\Theta^\top e_i\|_2\le\|\bZ_\Theta^\top e_i\|_2$, which is at most
$\sqrt{6r/p}$ for $i\in\mathcal I_0$ and equals $\sqrt{a'r/p}/\nu\le\sqrt{r/p}$
for $i\in\mathcal I_1$, because $\|\bX_\Theta\|_2\le\|\bX_\Theta\|_F\le\sqrt{a'r}\le\nu/8$
by~\eqref{eq:lb-cond}. Hence (F2) holds. Bounding entries by Cauchy--Schwarz
and summing rows,
$\|\bL_\Theta\|_{\infty\to\infty}\le2\nu r+r\sqrt{6a'}+a'r/\nu\le3\nu r\le\frac1{400}$,
so $(\bI,\bL_\Theta)\in\mathcal F^{\pi_+,\pi_-}$. For $\mathbf D\triangleq \bX_\Theta-\bX_{\Theta'}$,
\begin{align*}
\bL_\Theta-\bL_{\Theta'}={}&\bV_0\bm\Pi_\pi\mathbf D^\top+\mathbf D\bm\Pi_\pi\bV_0^\top\\
&+\nu^{-1}\big(\bX_\Theta\bm\Pi_\pi\bX_\Theta^\top-\bX_{\Theta'}\bm\Pi_\pi\bX_{\Theta'}^\top\big).
\end{align*}
The first two terms have Frobenius norm $\sqrt2\|\mathbf D\|_F$ because $\bV_0^\top \mathbf D=0$ and $\bm\Pi_\pi$ is orthogonal,
and the last has norm at most $2\nu^{-1}\max\|\bX\|_2\|\mathbf D\|_F\le\frac14\|\mathbf D\|_F$.
Since $\|\mathbf D\|_F^2=\frac{4a'}{p}\mathrm{Ham}(\Theta,\Theta')$,
\[
\tfrac{4a'}{p}\mathrm{Ham}\le\|\bL_\Theta-\bL_{\Theta'}\|_F^2\le\tfrac{16a'}{p}\mathrm{Ham},
\]
so neighbors have $\mathrm{KL}\le16Na'/p=1/2$. With $M'=|\mathcal I_1|r\ge pr/2$,
\eqref{eq:assouad-prob} gives $\|\tilde \bL-\bL_\Theta\|_F\ge\frac12\sqrt{a'M'/(2p)}
\ge\frac1{23}\sqrt{pr/N}$ with probability at least $1/7$.

\emph{Combination.} Let $t\triangleq\frac1{48}(\sqrt{pd/N}+\sqrt{pr/N})$. Then
$t\le\frac1{24}\max\{\sqrt{pd/N},\sqrt{pr/N}\}$, so the sparse subfamily (if
$d\ge r$) or the low-rank subfamily (otherwise) yields~\eqref{eq:lb-prob}. The
moment bound follows from $\mathbb E Z^2\ge t^2\Pr(Z\ge t)$.
\end{proof}

\begin{proof}[Proof of Theorem~\ref{thm:support_converse}]
Let $\bV_0$ and $\bm\Pi_\pi$ be as in the proof of Theorem~\ref{thm:lower}, $\nu\triangleq\nu_{\min}$, $\beta\triangleq\beta_{\min}$, and $\bL_{\rm b}\triangleq\nu\bV_0\bm\Pi_\pi\bV_0^\top$.  For each of the $M\triangleq p(p-1)/2$ unordered pairs $e=\{i,j\}$ put $\bS_e\triangleq\bI+\beta F_e$.  Each $(\bS_e,\bL_{\rm b})$ lies in $\mathcal F^{\pi_+,\pi_-}$: $\bS_e$ has a single edge of magnitude $\beta$, (F2) holds as in Theorem~\ref{thm:lower}, and $\|\bS_e-\bL_{\rm b}-\bI\|_{\infty\to\infty}\le\beta+2\nu r\le\frac1{800}+\frac1{600}<\frac1{200}$.  In particular $\|\bOm_e-\bI\|_2\le1/4$, and Lemma~\ref{lem:KL} gives, for $e\ne e'$,
\[
\mathrm{KL}\big(P_e^{\otimes N}\|P_{e'}^{\otimes N}\big)\le N\beta^2\|F_e-F_{e'}\|_F^2=4N\beta^2\le\tfrac14\log p
\]
by \eqref{eq:support_conv}.  Let $\hat e$ be the unique edge of $\hat{\mathsf G}$ if it has exactly one edge, and a fixed pair otherwise, so that $\{\hat{\mathsf G}=\mathsf G(\bS_e)\}\subseteq\{\hat e=e\}$.  If $V$ is uniform on the $M$ pairs, the mutual information between $V$ and the sample is at most $\max_{e\ne e'}\mathrm{KL}(P_e^{\otimes N}\|P_{e'}^{\otimes N})$, and Fano's inequality \cite{yu1997assouad} yields
\[
\max_e\Pr_e(\hat e\ne e)\ge1-\frac{\frac14\log p+\log2}{\log M}\ge\frac34-\frac{\log 2}{\log p}\ge\frac12,
\]
because $M\ge p^2/4$ gives $\log M\ge\log p$ for $p\ge4$, and $\log2/\log p\le\frac14$ for $p\ge16$.
\end{proof}

\begin{proof}[Proof of Theorem~\ref{thm:inertia_converse}]
\emph{Construction.}  Let $p_0\triangleq\lfloor p/2\rfloor$, $\mathcal I_0\triangleq[p_0]$, $\mathcal I_1\triangleq[p]\setminus\mathcal I_0$, and $p_1\triangleq|\mathcal I_1|\ge p/2$.  Let $\bV_0\in\R^{p\times(r-1)}$ consist of the first $r-1$ columns of the matrix $\bV_0$ of Theorem~\ref{thm:lower} (empty if $r=1$), and $\bm\Pi_0\triangleq\diag(\bI_{\pi_+-1},-\bI_{\pi_-})$.  For $\varepsilon\in\{\pm1\}^{p_1}$ let $\mathbf w_\varepsilon\triangleq\varepsilon/\sqrt{p_1}\in\R^{p_1}$, let $\mathbf v_\varepsilon\in\R^p$ equal $\mathbf w_\varepsilon$ on $\mathcal I_1$ and vanish on $\mathcal I_0$, and for $\sigma\in\{+1,-1\}$ put, with $\nu\triangleq\nu_{\min}$,
\[
\bL_{\varepsilon,\sigma}\triangleq\nu\big(\bV_0\bm\Pi_0\bV_0^\top+\sigma\,\mathbf v_\varepsilon\mathbf v_\varepsilon^\top\big).
\]
Since $\bV_0$ and $\mathbf v_\varepsilon$ have disjoint supports, $\bU\triangleq[\bV_0,\mathbf v_\varepsilon]$ is orthonormal, $\bL_{\varepsilon,\sigma}=\bU\,\nu\diag(\bm\Pi_0,\sigma)\,\bU^\top$, and $\In(\bL_{\varepsilon,+})=(\pi_+,\pi_-)$, $\In(\bL_{\varepsilon,-})=(\pi_+-1,\pi_-+1)$.  Moreover $(\bI,\bL_{\varepsilon,\sigma})\in\mathcal F$: (F1) is vacuous; (F2) holds since $\|\bU^\top e_i\|_2^2\le2(r-1)/p_0\le6r/p$ for $i\in\mathcal I_0$ and $\|\bU^\top e_i\|_2^2=1/p_1\le2/p$ for $i\in\mathcal I_1$; and, since $\bL_{\varepsilon,\sigma}$ is block diagonal with respect to $\mathcal I_0\cup\mathcal I_1$, $\|\bL_{\varepsilon,\sigma}\|_{\infty\to\infty}\le\nu\max\{2(r-1),\,\|\mathbf w_\varepsilon\mathbf w_\varepsilon^\top\|_{\infty\to\infty}\}\le2\nu r\le\frac1{200}$, which is (F3).

\emph{Reduction to one block.}  The precision $\bI-\bL_{\varepsilon,\sigma}$ is block diagonal, and its $\mathcal I_0$-block does not depend on $(\varepsilon,\sigma)$.  Hence, under $P_{\varepsilon,\sigma}$, the subvectors $\mathbf x_{\mathcal I_0}$ and $\mathbf x_{\mathcal I_1}$ are independent, the law of $\mathbf x_{\mathcal I_0}$ is fixed, and $\mathbf x_{\mathcal I_1}\sim\mathcal N(\mathbf 0,\bSig_\sigma(\mathbf w_\varepsilon))$ with, by the Sherman--Morrison formula,
\begin{align*}
\bSig_\sigma(\mathbf w)&\triangleq(\bI_{p_1}-\sigma\nu\mathbf w\mathbf w^\top)^{-1}=\bI_{p_1}+\beta_\sigma\mathbf w\mathbf w^\top,\\
\beta_+&\triangleq\frac{\nu}{1-\nu},\qquad \beta_-\triangleq-\frac{\nu}{1+\nu}.
\end{align*}
Let $\bar P_\sigma\triangleq2^{-p_1}\sum_\varepsilon P_{\varepsilon,\sigma}^{\otimes N}$ and $\bar Q_\sigma\triangleq2^{-p_1}\sum_\varepsilon\mathcal N(\mathbf 0,\bSig_\sigma(\mathbf w_\varepsilon))^{\otimes N}$.  Since $\bar P_\sigma$ is the product of a fixed law with $\bar Q_\sigma$, $\mathrm{TV}(\bar P_+,\bar P_-)=\mathrm{TV}(\bar Q_+,\bar Q_-)$.

\emph{A $\chi^2$ bound.}  Let $Q_0\triangleq\mathcal N(\mathbf 0,\bI_{p_1})$ and let $\phi_{\bSig}$ denote the $\mathcal N(\mathbf 0,\bSig)$ density.  For unit $\mathbf w,\mathbf w'$, the matrix $\bSig_\sigma(\mathbf w)^{-1}+\bSig_\sigma(\mathbf w')^{-1}-\bI=\bI-\sigma\nu(\mathbf w\mathbf w^\top+\mathbf w'\mathbf w'^\top)$ is positive definite because $2\nu<1$, and Gaussian integration gives, with $\bSig\triangleq\bSig_\sigma(\mathbf w)$ and $\bSig'\triangleq\bSig_\sigma(\mathbf w')$,
\begin{align*}
\int\frac{\phi_{\bSig}\,\phi_{\bSig'}}{\phi_{\bI}}
&=\det\big(\bSig+\bSig'-\bSig\bSig'\big)^{-1/2}\\
&=\big(1-\beta_\sigma^2\langle\mathbf w,\mathbf w'\rangle^2\big)^{-1/2},
\end{align*}
where the first step uses $\det\bSig\det\bSig'\det(\bSig^{-1}+\bSig'^{-1}-\bI)=\det(\bSig+\bSig'-\bSig\bSig')$ and the second uses $\bSig+\bSig'-\bSig\bSig'=\bI-\beta_\sigma^2\langle\mathbf w,\mathbf w'\rangle\mathbf w\mathbf w'^\top$.  Therefore, with $\varepsilon,\varepsilon'$ independent and uniform and $t\triangleq\langle\mathbf w_\varepsilon,\mathbf w_{\varepsilon'}\rangle$,
\[
1+\chi^2\big(\bar Q_\sigma\|Q_0^{\otimes N}\big)=\E_{\varepsilon,\varepsilon'}\big(1-\beta_\sigma^2t^2\big)^{-N/2}\le\E\exp\big(N\beta_\sigma^2t^2\big),
\]
where we used $\beta_\sigma^2t^2\le\beta_+^2\le(4\nu/3)^2\le\frac12$ and $-\log(1-x)\le2x$ on $[0,\frac12]$.  The variable $t$ has the law of $S/\sqrt{p_1}$ with $S\triangleq p_1^{-1/2}\sum_{i=1}^{p_1}\varepsilon_i$, which satisfies $\E e^{\theta S}=\cosh(\theta/\sqrt{p_1})^{p_1}\le e^{\theta^2/2}$.  For $g\sim\mathcal N(0,1)$ independent of $S$ and $0\le a<\frac12$, $\E e^{aS^2}=\E e^{\sqrt{2a}\,gS}\le\E e^{ag^2}=(1-2a)^{-1/2}$.  With $a=N\beta_\sigma^2/p_1\le\frac{16}{9}\nu^2\cdot\frac{2N}{p}\le\frac29$ by \eqref{eq:inertia_conv}, we obtain $\chi^2(\bar Q_\sigma\|Q_0^{\otimes N})\le(5/9)^{-1/2}-1<0.35$.

\emph{Testing.}  Since $\mathrm{TV}\le\frac12\sqrt{\chi^2}$, $\mathrm{TV}(\bar Q_+,\bar Q_-)\le\mathrm{TV}(\bar Q_+,Q_0^{\otimes N})+\mathrm{TV}(Q_0^{\otimes N},\bar Q_-)<0.6$.  For any $\hat\jmath$, let $\psi\triangleq\mathbf 1\{\hat\jmath=(\pi_+,\pi_-)\}$.  Then
\begin{align*}
&\bar P_+\big(\hat\jmath\ne(\pi_+,\pi_-)\big)+\bar P_-\big(\hat\jmath\ne(\pi_+-1,\pi_-+1)\big)\\
&\quad\ge\bar P_+(\psi=0)+\bar P_-(\psi=1)\ge1-\mathrm{TV}(\bar P_+,\bar P_-)>\frac25,
\end{align*}
so one of the two mixtures errs with probability greater than $1/5$, and since each $\bar P_\sigma$ is an average of $P_{\varepsilon,\sigma}^{\otimes N}$, so does some $(\bI,\bL_{\varepsilon,\sigma})\in\mathcal F$.  The final statement follows by contraposition from \eqref{eq:inertia_conv}.
\end{proof}

\begin{proof}[Proof of Corollary~\ref{cor:sandwich}]
By construction, $\gamma\ge4\mu_{\mathcal R}\lambda$ in both cases of
\eqref{eq:tuning-class}: if $\lambda=30\bar\varepsilon_\infty$, this holds by the
definition of $\gamma$; if $\lambda=4\bar\xi\gamma$, it follows from
$16\mu_{\mathcal R}\bar\xi\le1$. Also $\gamma/\lambda\le1/(4\bar\xi)$, so
Lemma~\ref{lem:class} yields Assumptions~\ref{ass:invert}--\ref{ass:irrep} with
$(\bar\kappa,\frac25)$. On the event of Corollary~\ref{cor:explicit},
$\varepsilon_W\le\frac1{30}=\frac{\alpha}{12}$, which is (C2). An inspection of
the proof of Theorem~\ref{thm:pdw} shows that it remains valid when $\kappa_H$,
$\alpha^{-1}$, $\nu^{-1}$, and $\omega_-^{-1}$ are replaced by upper bounds,
since these enter only through upper estimates; and (C4) holds by
$\beta_{\min}>2\bar\kappa\lambda$. Theorem~\ref{thm:pdw}(c) gives
\eqref{eq:unif-upper}. For~\eqref{eq:unif-rate}, Remark~\ref{rem:scale} with $s^2=2p$ and
$N\ge p^2\ell$ gives $\bar\varepsilon_\infty\lesssim\sqrt{\ell/N}$ and
$\bar\varepsilon_2\lesssim\sqrt{p\ell/N}$. Since $d+1\le\sqrt p$,
\[
\lambda\lesssim\sqrt{\ell/N}\,(1+\sqrt{\mu r}),\qquad
\sqrt r\,\gamma\lesssim\sqrt{pr\ell/N},
\]
and $\sqrt{pd\,\mu r}\le\sqrt{\mu\min\{d,r\}}\,(\sqrt{pd}+\sqrt{pr})$.
\end{proof}

\section{Proofs of the Robustness Results}
\label{app:robust}

\begin{proof}[Proof of Theorem~\ref{thm:bias}]
If $\mu_\Psi=0$ the claim is trivial.  Otherwise let $\mathbf m_\Psi,\bSig_\Psi$ be the mean and covariance under $Q_\Psi$.  The nominal transformed mean is zero, so the mixture covariance identity gives
\begin{align}
\bSig_u(P)=(1-\epsilon_u)\bSig_u^*+\epsilon_u\bSig_\Psi+\epsilon_u(1-\epsilon_u)\mathbf m_\Psi\mathbf m_\Psi^\top.
\label{eq:mixture}
\end{align}
With $\mathbf a_\Psi=\E_\Psi[\mathbf xu]$ and $\bV_\Psi=\E_\Psi[\mathbf x\mathbf x^\top u]$, \eqref{eq:R1R2} gives $\|\bV_\Psi\|_2\le R_2$, and Cauchy--Schwarz gives $\|\mathbf a_\Psi\|_2^2\le\mu_\Psi R_2$.  Since $\bSig_\Psi\preceq\bV_\Psi/\mu_\Psi$, $\epsilon_u\|\bSig_\Psi\|_2\le\epsilon R_2/Z_\epsilon$ and $\epsilon_u(1-\epsilon_u)\|\mathbf m_\Psi\|_2^2\le\epsilon_u\|\mathbf a_\Psi\|_2^2/\mu_\Psi^2\le\epsilon R_2/Z_\epsilon$; also $\epsilon_u\le\epsilon/Z_\epsilon$ since $\mu_\Psi\le1$.  Thus $\|\bSig_u(P)-\bSig_u^*\|_2\le\epsilon(\|\bSig_u^*\|_2+2R_2)/Z_\epsilon$, and $Z_\epsilon\ge(1-\epsilon)\mu_0$.  The inverse bound follows from $(\bSig+\bE)^{-1}-\bSig^{-1}=-\bSig^{-1}\bE(\bI+\bSig^{-1}\bE)^{-1}\bSig^{-1}$ and a Neumann series.
\end{proof}

\begin{proof}[Proof of Proposition~\ref{prop:bias_sharp}]
By Theorem~\ref{thm:point_mass}, $\bSig_u(P)-\bSig_u^*=-\epsilon_u\bSig_u^*+\epsilon_u(1-\epsilon_u)\mathbf r\mathbf r^\top$ with $\epsilon_u=\epsilon e^{-1}/((1-\epsilon)\mu_0+\epsilon e^{-1})$.  The assumption gives $\epsilon_u\le\frac12$ and $\epsilon_u\ge\epsilon e^{-1}/(2(1-\epsilon)\mu_0)$.  Since $\|\bSig_u^*\|_2=s^2\lambda_{\max}(\bSig^*)/(s^2+\lambda_{\max}(\bSig^*))\le s^2/2$, testing with $\hat{\mathbf r}=\mathbf r/\|\mathbf r\|_2$ yields $\hat{\mathbf r}^\top(\bSig_u(P)-\bSig_u^*)\hat{\mathbf r}\ge\epsilon_u[(1-\epsilon_u)2s^2-s^2/2]\ge\epsilon_us^2/2$.
\end{proof}

\begin{proof}[Proof of Theorem~\ref{thm:point_mass}]
The point mass has mean $\mathbf r$ and zero covariance under $Q_u$, so \eqref{eq:mixture} gives $(1-\epsilon_u)\bSig_u^*+\epsilon_u(1-\epsilon_u)\mathbf r\mathbf r^\top$, which is \eqref{eq:Sigma_uc}.  The Sherman--Morrison formula gives $\bOm_{u,c}=(1-\epsilon_u)^{-1}[\bOm_u^*-\epsilon_u\bOm_u^*\mathbf r\mathbf r^\top\bOm_u^*/(1+\epsilon_u\mathbf r^\top\bOm_u^*\mathbf r)]$; substituting $\bOm_u^*=\bS_u^*-\bL^*$ gives \eqref{eq:point_mass_decomp}.  As $\|\mathbf r\|_2\to\infty$, $u(\mathbf r;s)=e^{-\|\mathbf r\|_2^2/(2s^2)}$ gives $\epsilon_u\to0$ and $\epsilon_u\|\mathbf r\|_2^2\to0$; the rank-one term has norm at most $\epsilon_u\|\mathbf r\|_2^2\|\bOm_u^*\|_2^2/(1-\epsilon_u)$, and the rescalings contribute $O(\epsilon_u)$.
\end{proof}

\begin{proof}[Proof of Corollary~\ref{cor:robust_nominal}]
$\bW=(\hat\bSig_u-\bSig_u(P))+(\bSig_u(P)-\bSig_u^*)$; the first term is controlled by Lemmas~\ref{lem:conc_inf} and~\ref{lem:conc_op} under $P$, the second by Theorem~\ref{thm:bias} together with $\|\cdot\|_\infty\le\|\cdot\|_2$ and $\|\cD^{-1}(\cdot)\|_2\le\bmin^{-2}\|\cdot\|_2$.  This gives \eqref{eq:robust_noise}, and hence (C2).  Apply Theorem~\ref{thm:pdw} with $\bSig_0=\bSig_u^*$.
\end{proof}

\section{Joint-Cone Theory and Thresholded Recovery}
\label{sec:theory}

This appendix develops a complementary deterministic theory, based on the single joint cone generated by the two regularizers, that yields Frobenius bounds without irrepresentability, at the price of dimension-dependent constants (Remark~\ref{rem:cone_dimension}).  We keep the notation of Section~\ref{sec:setup}, including $\xi(\mathbf T)$ and $\chi_B$.

\subsection{Deterministic Ingredients}

\begin{lemma}[Decomposability]
\label{lem:decomp}
For all $\bDelta_S,\bDelta_L\in\Sym^p$,
\begin{align}
&\|\bS_0+\bDelta_S\|_{1,\mathrm{off}}-\|\bS_0\|_{1,\mathrm{off}}\ge\|\PRp\bDelta_S\|_1-\|\PR\bDelta_S\|_1,\label{eq:decomp_S}\\
&\|\cD(\bL_0+\bDelta_L)\|_*-\|\cD(\bL_0)\|_*\notag\\
&\qquad\ge\|\PTtp\cD(\bDelta_L)\|_*-\|\PTt\cD(\bDelta_L)\|_*.\label{eq:decomp_L}
\end{align}
\end{lemma}
\begin{proof}
For \eqref{eq:decomp_S}, the off-diagonal entries outside $\mathcal E$ contribute $\|\PRp\bDelta_S\|_1$ and those in $\mathcal E$ change by at least $-\|P_{\mathcal E}\bDelta_S\|_1\ge-\|\PR\bDelta_S\|_1$.  For \eqref{eq:decomp_L}, with $\cD(\bL_0)=\tilde\bU\tilde\bGamma\tilde\bU^\top$, the matrix $\sgn(\cD(\bL_0))+\bY$ is a subgradient of $\|\cdot\|_*$ at $\cD(\bL_0)$ for every $\bY\in\Tt^\perp$ with $\|\bY\|_2\le1$ \cite{lewis1996convex}.  Choosing $\bY$ dual to $\PTtp\cD(\bDelta_L)$ and using $|\langle\sgn(\cD(\bL_0)),\PTt\cD(\bDelta_L)\rangle|\le\|\PTt\cD(\bDelta_L)\|_*$ gives the claim.
\end{proof}

\begin{lemma}[Local curvature]
\label{lem:curv}
Let $h_0(\bOm)\triangleq\tr(\bOm\hat\bSig_u)-\log\det\bOm$.  For every $\bDelta\in\Sym^p$ with $\|\bDelta\|_2\le\omega_-/2$,
\begin{align}
h_0(\bOm_0+\bDelta)-h_0(\bOm_0)-\langle\nabla h_0(\bOm_0),\bDelta\rangle\ge\frac{2}{9\omega_+^2}\|\bDelta\|_F^2.
\label{eq:curv}
\end{align}
\end{lemma}
\begin{proof}
$\nabla^2h_0(\bOm)[\bH,\bH]=\tr(\bH\bOm^{-1}\bH\bOm^{-1})\ge\|\bH\|_F^2/\lambda_{\max}(\bOm)^2$.  Along $\bOm_0+t\bDelta$, $t\in[0,1]$, Weyl's inequality gives $\lambda_{\max}\le\omega_++\omega_-/2\le3\omega_+/2$ and $\lambda_{\min}\ge\omega_-/2>0$.  The integral form of the Taylor remainder with $\int_0^1(1-t)\,dt=1/2$ yields \eqref{eq:curv}.
\end{proof}

\subsection{The Weighted Joint Cone}

For $c>0$ define the closed cone
\begin{align}
\cC(c)\triangleq\Big\{&(\bA,\bH)\in\Sym^p\times\Sym^p:\notag\\
&\lambda\|\PRp\bA\|_1+\gamma\|\PTtp\cD(\bH)\|_*\notag\\
&\le c\big[\lambda\|\PR\bA\|_1+\gamma\|\PTt\cD(\bH)\|_*\big]\Big\}.
\label{eq:cone}
\end{align}

\begin{lemma}[Joint cone generated by the estimator]
\label{lem:cone}
Suppose
\begin{align}
\lambda\ge4\|\bW\|_\infty,\qquad \gamma\ge4\|\cD^{-1}(\bW)\|_2.
\label{eq:dual_noise}
\end{align}
If $t\in[0,1]$ and $(\bS_0+t\bDelta_S,\bL_0+t\bDelta_L)$ is feasible with objective value in \eqref{eq:estimator} not exceeding that of $(\bS_0,\bL_0)$, then $(t\bDelta_S,t\bDelta_L)\in\cC(5/3)$ and, with $\bar\bDelta\triangleq t\bDelta$,
\begin{align}
&\mathcal B(\bar\bDelta)+\tfrac34\lambda\|\PRp t\bDelta_S\|_1+\tfrac34\gamma\|\PTtp\cD(t\bDelta_L)\|_*\notag\\
&\quad\le\tfrac54\lambda\|\PR t\bDelta_S\|_1+\tfrac54\gamma\|\PTt\cD(t\bDelta_L)\|_*,
\label{eq:basic}
\end{align}
where $\mathcal B(\bar\bDelta)\triangleq h_0(\bOm_0+\bar\bDelta)-h_0(\bOm_0)-\langle\bW,\bar\bDelta\rangle\ge0$.  In particular, this holds for every minimizer with $t=1$.
\end{lemma}
\begin{proof}
$\nabla h_0(\bOm_0)=\hat\bSig_u-\bSig_0=\bW$.  The objective comparison and Lemma~\ref{lem:decomp} give
$\mathcal B(\bar\bDelta)+\lambda\|\PRp t\bDelta_S\|_1+\gamma\|\PTtp\cD(t\bDelta_L)\|_*\le\lambda\|\PR t\bDelta_S\|_1+\gamma\|\PTt\cD(t\bDelta_L)\|_*-\langle\bW,t\bDelta_S\rangle+\langle\bW,t\bDelta_L\rangle$.
By H\"older's inequality, $|\langle\bW,\bA\rangle|\le\|\bW\|_\infty(\|\PR\bA\|_1+\|\PRp\bA\|_1)$, and since $\langle\bW,\bH\rangle=\langle\cD^{-1}(\bW),\cD(\bH)\rangle$, $|\langle\bW,\bH\rangle|\le\|\cD^{-1}(\bW)\|_2(\|\PTt\cD(\bH)\|_*+\|\PTtp\cD(\bH)\|_*)$.  Inserting \eqref{eq:dual_noise} gives \eqref{eq:basic}; dropping $\mathcal B\ge0$ and multiplying by $4/3$ gives membership in $\cC(5/3)$.
\end{proof}

\subsection{Cone--Kernel Separation}

In this appendix only, let $\cA(\bA,\bH)\triangleq\bA-\bH$ on $\Sym^p\times\Sym^p$ with norm $\|(\bA,\bH)\|_\oplus\triangleq(\|\bA\|_F^2+\|\bH\|_F^2)^{1/2}$, so $\ker\cA=\{(\bM,\bM)\}$.  Fix $c_0\triangleq5/3$ and set
\begin{align}
\alpha_{\cC}&\triangleq\min\big\{\|\bA-\bH\|_F:(\bA,\bH)\in\cC(c_0),\ \|(\bA,\bH)\|_\oplus=1\big\},\notag\\
\kappa_{\mathrm{id}}&\triangleq\alpha_{\cC}/\sqrt2.
\label{eq:alphaC}
\end{align}

\begin{proposition}[Reformulation of restricted injectivity]
\label{prop:transversality}
The following are equivalent: (i) $\alpha_{\cC}>0$; (ii) $\cC(c_0)\cap\ker\cA=\{0\}$; (iii) $\Phi(\bM)>0$ for all $\bM\ne0$, where
\begin{align}
\Phi(\bM)\triangleq{}&\lambda\big(\|\PRp\bM\|_1-c_0\|\PR\bM\|_1\big)\notag\\
&+\gamma\big(\|\PTtp\cD(\bM)\|_*-c_0\|\PTt\cD(\bM)\|_*\big).
\label{eq:Phi}
\end{align}
Under these conditions, $\|\bA-\bH\|_F\ge\kappa_{\mathrm{id}}(\|\bA\|_F+\|\bH\|_F)$ on $\cC(c_0)$, and $\kappa_{\mathrm{id}}=\min\{\mathrm{dist}(z,\ker\cA):z\in\cC(c_0),\|z\|_\oplus=1\}$.
\end{proposition}
\begin{proof}
(ii)$\Leftrightarrow$(iii) is the definition of $\cC(c_0)$ evaluated at $(\bM,\bM)$.  The unit section of the closed cone $\cC(c_0)$ is compact and $z\mapsto\|\cA z\|_F$ is continuous, so the minimum in \eqref{eq:alphaC} is attained; it is positive iff (ii) holds.  Homogeneity gives $\|\bA-\bH\|_F\ge\alpha_{\cC}\|(\bA,\bH)\|_\oplus\ge\kappa_{\mathrm{id}}(\|\bA\|_F+\|\bH\|_F)$.  The orthogonal projection of $(\bA,\bH)$ onto $(\ker\cA)^\perp$ is $\frac12(\bA-\bH,\bH-\bA)$, whose norm is $\|\bA-\bH\|_F/\sqrt2$.
\end{proof}

Proposition~\ref{prop:transversality} identifies the relevant geometry, but it is a reformulation.  We now characterize the margin $\delta_{\mathrm{sep}}\triangleq\min_{\|\bM\|_F=1}\Phi(\bM)$ by duality and give primitive sufficient conditions.  Let $\mathbb B_\infty(\mathcal V)$ denote the $\|\cdot\|_\infty$ unit ball of a coordinate subspace $\mathcal V$, $\mathbb B_2(\mathcal V)$ the spectral unit ball of $\mathcal V$, and $\mathbb B_F$ the Frobenius unit ball of $\Sym^p$.  Define the compact convex sets
\begin{align}
\mathsf G&\triangleq\lambda\,\mathbb B_\infty(\mathcal R^\perp)+\gamma\,\cD\big(\mathbb B_2(\Tt^\perp)\big),\notag\\
\mathsf H&\triangleq c_0\big[\lambda\,\mathbb B_\infty(\mathcal R)+\gamma\,\cD\big(\PTt\mathbb B_2(\Sym^p)\big)\big].
\label{eq:GH}
\end{align}

\begin{proposition}[Dual characterization of the separation margin]
\label{prop:dual_margin}
$\Phi=\sigma_{\mathsf G}-\sigma_{\mathsf H}$, where $\sigma$ denotes the support function, and for every $\varepsilon\ge0$,
\begin{align}
\delta_{\mathrm{sep}}\ge\varepsilon\iff\mathsf H+\varepsilon\,\mathbb B_F\subseteq\mathsf G.
\label{eq:dual_margin}
\end{align}
In particular, if $\delta_{\mathrm{sep}}\ge0$, then $\delta_{\mathrm{sep}}=\max\{\varepsilon\ge0:\mathsf H+\varepsilon\mathbb B_F\subseteq\mathsf G\}$.
\end{proposition}
\begin{proof}
By duality of $\ell_1/\ell_\infty$, $\sigma_{\lambda\mathbb B_\infty(\mathcal R^\perp)}(\bM)=\lambda\|\PRp\bM\|_1$.  Since $\PTtp$ is a spectral-norm contraction that fixes $\Tt^\perp$, $\{\PTtp\bY:\|\bY\|_2\le1\}=\mathbb B_2(\Tt^\perp)$, and by self-adjointness of $\cD$, $\sigma_{\cD(\mathbb B_2(\Tt^\perp))}(\bM)=\sup_{\|\bY\|_2\le1}\langle\bY,\PTtp\cD(\bM)\rangle=\|\PTtp\cD(\bM)\|_*$.  Similarly $\sigma_{\mathsf H}(\bM)=c_0[\lambda\|\PR\bM\|_1+\gamma\|\PTt\cD(\bM)\|_*]$.  Support functions are additive over Minkowski sums, so $\Phi=\sigma_{\mathsf G}-\sigma_{\mathsf H}$.  Since $\sigma_{\mathsf H+\varepsilon\mathbb B_F}=\sigma_{\mathsf H}+\varepsilon\|\cdot\|_F$ and closed convex sets are ordered by inclusion iff their support functions are ordered pointwise, $\mathsf H+\varepsilon\mathbb B_F\subseteq\mathsf G$ iff $\Phi(\bM)\ge\varepsilon\|\bM\|_F$ for all $\bM$.  By positive homogeneity of $\Phi$, the latter holds iff $\delta_{\mathrm{sep}}=\min_{\|\bM\|_F=1}\Phi(\bM)\ge\varepsilon$; the minimum exists by continuity and compactness.
\end{proof}

\begin{theorem}[Primitive sufficient conditions for cone--kernel separation]
\label{thm:primitive}
Assume $2\mu_{\mathcal R}\xi(\mathbf T)\le1/2$ and let $\varepsilon\ge0$.  Put
\begin{align*}
q_Q&\triangleq c_0\mu_{\mathcal R}\lambda+2c_0\bmax^2\gamma+\varepsilon,\\
q_\perp&\triangleq c_0\mu_{\mathcal R}\lambda+4c_0\bmax^2\bmin^{-2}\chi_B\gamma+\varepsilon.
\end{align*}
If
\begin{align}
4\xi(\mathbf T)\,q_Q\le\lambda
\quad\text{and}\quad
q_\perp+4\mu_{\mathcal R}\xi(\mathbf T)\,q_Q\le\bmin^2\gamma,
\label{eq:primitive}
\end{align}
then $\delta_{\mathrm{sep}}\ge\varepsilon$.  In particular, if \eqref{eq:primitive} holds with strict inequalities for $\varepsilon=0$, then $\alpha_{\cC}>0$.
\end{theorem}
\begin{proof}
By Proposition~\ref{prop:dual_margin} it suffices to write every $\bQ\in\mathsf H+\varepsilon\mathbb B_F$ as $\bX+\bY$ with $\bX\in\mathcal R^\perp$, $\|\bX\|_\infty\le\lambda$, and $\bY\in\cD(\Tt^\perp)=\mathbf T^\perp$ (Lemma~\ref{lem:congruence}(ii)), $\|\cD^{-1}(\bY)\|_2\le\gamma$.

\emph{Size of $\bQ$.}  Write $\bQ=c_0\lambda\bZ_{\mathcal R}+c_0\gamma\cD(\bY_T)+\varepsilon\bE$ with $\bZ_{\mathcal R}\in\mathbb B_\infty(\mathcal R)$, $\bY_T=\PTt\bY$, $\|\bY\|_2\le1$, $\|\bE\|_F\le1$.  Then $\|c_0\lambda\bZ_{\mathcal R}\|_2\le c_0\mu_{\mathcal R}\lambda$ and $\|\bY_T\|_2\le2$ by \eqref{eq:proj_facts}, so $\|\bQ\|_2\le q_Q$.  Next we bound $\|\PTp\cD(\bY_T)\|_2$.  Since $\bY_T=\bP_{\tilde U}\bY_T+(\bI-\bP_{\tilde U})\bY_T\bP_{\tilde U}$,
\begin{align*}
\PTp\cD(\bY_T)={}&(\bI-\bP_U)\bB\bP_{\tilde U}\bY_T\bB(\bI-\bP_U)\\
&+(\bI-\bP_U)\bB(\bI-\bP_{\tilde U})\bY_T\bP_{\tilde U}\bB(\bI-\bP_U).
\end{align*}
Now $\bP_{\tilde U}=\bB\bU(\bU^\top\bB^2\bU)^{-1}\bU^\top\bB$, so $(\bI-\bP_U)\bB\bP_{\tilde U}=(\bI-\bP_U)\bB^2\bU(\bU^\top\bB^2\bU)^{-1}\bU^\top\bB$ has norm at most $\chi_B\bmin^{-2}\bmax$, and $\|\bP_{\tilde U}\bB(\bI-\bP_U)\|_2$ is the same.  Hence $\|\PTp\cD(\bY_T)\|_2\le2\chi_B\bmax^2\bmin^{-2}\|\bY_T\|_2\le4\chi_B\bmax^2\bmin^{-2}$, and $\|\PTp\bQ\|_2\le q_\perp$ by \eqref{eq:proj_facts}.

\emph{Construction.}  Let $\mathcal K\bUps\triangleq\PT\PR\bUps$ on $(\mathbf T,\|\cdot\|_2)$.  For $\bUps\in\mathbf T$, $\|\mathcal K\bUps\|_2\le2\|\PR\bUps\|_2\le2\mu_{\mathcal R}\|\bUps\|_\infty\le2\mu_{\mathcal R}\xi(\mathbf T)\|\bUps\|_2\le\frac12\|\bUps\|_2$.  Hence $\bUps\triangleq(\mathrm{Id}-\mathcal K)^{-1}\PT\bQ\in\mathbf T$ exists with $\|\bUps\|_2\le2\|\PT\bQ\|_2\le4q_Q$.  Set $\bX\triangleq\PRp\bUps$ and $\bY\triangleq\bQ-\bX$.  Then $\PT\bX=\bUps-\PT\PR\bUps=\PT\bQ$, so $\PT\bY=0$ and $\bY\in\mathbf T^\perp$.  Moreover $\|\bX\|_\infty\le\|\bUps\|_\infty\le\xi(\mathbf T)\|\bUps\|_2\le4\xi(\mathbf T)q_Q\le\lambda$.  Finally $\bY=\PTp\bY=\PTp\bQ+\PTp\PR\bUps$, so $\|\bY\|_2\le q_\perp+\mu_{\mathcal R}\xi(\mathbf T)\|\bUps\|_2\le q_\perp+4\mu_{\mathcal R}\xi(\mathbf T)q_Q\le\bmin^2\gamma$ and $\|\cD^{-1}(\bY)\|_2\le\bmin^{-2}\|\bY\|_2\le\gamma$.  If the inequalities are strict for $\varepsilon=0$, they remain valid for some $\varepsilon>0$ by continuity, and $\delta_{\mathrm{sep}}>0$ is (iii) of Proposition~\ref{prop:transversality}.
\end{proof}

\begin{remark}[Feasible window]
\label{rem:window}
For $\tau=0$ ($\bmin=\bmax=1$, $\chi_B=0$) and $\varepsilon=0$, \eqref{eq:primitive} reads $4\xi(c_0\mu_{\mathcal R}\lambda+2c_0\gamma)\le\lambda$ and $c_0\mu_{\mathcal R}(1+4\mu_{\mathcal R}\xi)\lambda+8c_0\mu_{\mathcal R}\xi\gamma\le\gamma$, i.e.
\[
\frac{c_0\mu_{\mathcal R}(1+4\mu_{\mathcal R}\xi)}{1-8c_0\mu_{\mathcal R}\xi}\le\frac\gamma\lambda\le\frac{1-4c_0\mu_{\mathcal R}\xi}{8c_0\xi}.
\]
If $\mu_{\mathcal R}\xi(\mathbf T)\le1/(32c_0^2)$, the left side is at most $2c_0\mu_{\mathcal R}$ and the right side is at least $(32c_0^2\mu_{\mathcal R})(1-1/(8c_0))/(8c_0)>2c_0\mu_{\mathcal R}$, so the window has nonempty interior.  This is a rank--sparsity incoherence condition of the type in \cite{chandrasekaran2011rank}.  For $\tau>0$, \eqref{eq:primitive} additionally requires $\chi_B$ to be small: the prior helps identifiability when the low-rank column space is nearly invariant under the structural energy.
\end{remark}

\begin{proposition}[From the margin to the restricted gain]
\label{prop:certificate}
Let $L_S\triangleq\lambda(\sqrt{p^2-m}+c_0\sqrt m)$ and $L_L\triangleq\gamma\bmax^2(\sqrt p+c_0\sqrt{2r})$.  If $\delta_{\mathrm{sep}}>0$, then $\kappa_{\mathrm{id}}\ge\delta_{\mathrm{sep}}/(\delta_{\mathrm{sep}}+L_S+L_L)$.
\end{proposition}
\begin{proof}
Let $f_S(\bA)\triangleq\lambda(\|\PRp\bA\|_1-c_0\|\PR\bA\|_1)$ and $f_L(\bH)\triangleq\gamma(\|\PTtp\cD(\bH)\|_*-c_0\|\PTt\cD(\bH)\|_*)$.  Since $\|\PRp\bE\|_1\le\sqrt{p^2-m}\|\bE\|_F$, $\|\PR\bE\|_1\le\sqrt m\|\bE\|_F$, $\|\PTtp\cD(\bE)\|_*\le\sqrt p\bmax^2\|\bE\|_F$, and $\|\PTt\cD(\bE)\|_*\le\sqrt{2r}\bmax^2\|\bE\|_F$, the reverse triangle inequality shows that $f_S$ and $f_L$ are $L_S$- and $L_L$-Lipschitz in $\|\cdot\|_F$.  For $(\bA,\bH)\in\cC(c_0)$, put $\bM=(\bA+\bH)/2$ and $\bE=\bA-\bH$.  Then $f_S(\bA)+f_L(\bH)\le0$, so $\delta_{\mathrm{sep}}\|\bM\|_F\le f_S(\bM)+f_L(\bM)\le\frac{L_S+L_L}2\|\bE\|_F$, and $\|\bA\|_F+\|\bH\|_F\le2\|\bM\|_F+\|\bE\|_F\le(1+(L_S+L_L)/\delta_{\mathrm{sep}})\|\bE\|_F$.
\end{proof}

\begin{remark}[Dimension dependence of the cone route]
\label{rem:cone_dimension}
Since $\delta_{\mathrm{sep}}$ scales like $\lambda$ and $\gamma$, the lower bound of Proposition~\ref{prop:certificate} can be as small as order $1/p$, and $R_N$ below scales with $\kappa_{\mathrm{id}}^{-2}$.  This is inherent to controlling the full joint cone by Frobenius geometry: cone members may carry off-model mass that is large in Frobenius norm.  Dimension-adequate constants are obtained in Section~\ref{sec:pdw} by the primal--dual witness route, where the error is confined to the model spaces and only the restricted Fisher information matters.
\end{remark}

\subsection{A Noncircular Deterministic Frobenius Bound}

\begin{theorem}[Joint estimation error]
\label{thm:main}
Assume \eqref{eq:dual_noise} and $\alpha_{\cC}>0$.  Let
\begin{align}
\Psi(\lambda,\gamma)\triangleq\lambda\sqrt m+\gamma\sqrt{2r}\,\varpi_B,\
R_N\triangleq\frac{45\,\omega_+^2}{8\,\kappa_{\mathrm{id}}^2}\Psi(\lambda,\gamma).
\label{eq:RN}
\end{align}
If $R_N<\omega_-/2$, then every minimizer of \eqref{eq:estimator} satisfies $\|\hat\bS-\bS_0\|_F+\|\hat\bL-\bL_0\|_F\le R_N$ and $\|\hat\bOm-\bOm_0\|_2<\omega_-/2$.
\end{theorem}
\begin{proof}
Let $t_0\triangleq\min\{1,\omega_-/(2\|\bDelta\|_2)\}$ ($t_0=1$ if $\bDelta=0$) and $(\bar\bDelta_S,\bar\bDelta_L,\bar\bDelta)=t_0(\bDelta_S,\bDelta_L,\bDelta)$.  The scaled point is feasible because $\|\bar\bDelta\|_2\le\omega_-/2$, and if $F$ denotes the objective of \eqref{eq:estimator}, convexity and optimality give $F(\bS_0+\bar\bDelta_S,\bL_0+\bar\bDelta_L)\le(1-t_0)F(\bS_0,\bL_0)+t_0F(\hat\bS,\hat\bL)\le F(\bS_0,\bL_0)$.  Lemma~\ref{lem:cone} gives $(\bar\bDelta_S,\bar\bDelta_L)\in\cC(5/3)$, so $\|\bar\bDelta\|_F\ge\kappa_{\mathrm{id}}\bar q$ with $\bar q\triangleq\|\bar\bDelta_S\|_F+\|\bar\bDelta_L\|_F$.  Lemma~\ref{lem:curv} and \eqref{eq:basic} give
\begin{align*}
\frac{2}{9\omega_+^2}\|\bar\bDelta\|_F^2&\le\frac54\big(\lambda\|\PR\bar\bDelta_S\|_1+\gamma\|\PTt\cD(\bar\bDelta_L)\|_*\big)\\
&\le\frac54\Psi(\lambda,\gamma)\,\bar q,
\end{align*}
using $\|\PR\bar\bDelta_S\|_1\le\sqrt m\|\bar\bDelta_S\|_F$, $\rank(\PTt\cD(\bar\bDelta_L))\le2r$, and Lemma~\ref{lem:congruence}(iii).  Hence $\bar q\le R_N$.  If $t_0<1$, then $\omega_-/2=\|\bar\bDelta\|_2\le\|\bar\bDelta\|_F\le\bar q\le R_N<\omega_-/2$, a contradiction; so $t_0=1$.
\end{proof}

\begin{corollary}[Nominal Frobenius rate via the joint cone]
\label{cor:nominal}
Under the nominal model, with $(\bS_0,\bL_0,\bOm_0)=(\bS_u^*,\bL^*,\bOm_u^*)$, let $\varepsilon_\infty$ and $\varepsilon_{2,B}$ be as in Corollary~\ref{cor:explicit} with $\eta$ replaced by $\eta/2$.  Choose $\lambda\ge4\varepsilon_\infty$ and $\gamma\ge4\varepsilon_{2,B}$, assume $\alpha_{\cC}>0$, and assume $R_N<\lambda_{\min}(\bOm_u^*)/2$.  Then, with probability at least $1-\eta$, $\|\hat\bS-\bS_u^*\|_F+\|\hat\bL-\bL^*\|_F\le R_N$.
\end{corollary}
\begin{proof}
On the intersection of the two events, \eqref{eq:dual_noise} holds; apply Theorem~\ref{thm:main}.
\end{proof}

\subsection{Thresholded Recovery without Irrepresentability}
\label{sec:threshold}

When irrepresentability is doubtful, structure can still be recovered from the Frobenius bound of Theorem~\ref{thm:main} by post-processing.  For $\vartheta>0$ let $[\mathcal T_\vartheta(\bS)]_{ij}\triangleq S_{ij}\mathbf 1\{|S_{ij}|>\vartheta\}$ for $i\ne j$, with the diagonal unchanged, and for $\bM=\mathbf Q\diag(\xi_j)\mathbf Q^\top$ let $\mathcal H_\vartheta(\bM)\triangleq\mathbf Q\diag(\xi_j\mathbf 1\{|\xi_j|>\vartheta\})\mathbf Q^\top$.  Set $\bar\bS\triangleq\mathcal T_{R_N}(\hat\bS)$ and $\bar\bL\triangleq\cD^{-1}[\mathcal H_{\bmax^2R_N}(\cD(\hat\bL))]$.

\begin{theorem}[Thresholded recovery]
\label{thm:threshold}
On the event of Theorem~\ref{thm:main}, if $\min_{(i,j)\in\mathcal E}|(\bS_0)_{ij}|>2R_N$ and $\sigma_r(\cD(\bL_0))>2\bmax^2R_N$, then $\sign(\bar S_{ij})=\sign((\bS_0)_{ij})$ for all $i\ne j$ and $\rank(\bar\bL)=r$.
\end{theorem}
\begin{proof}
$\|\hat\bS-\bS_0\|_\infty\le\|\hat\bS-\bS_0\|_F\le R_N$, so null entries are removed and entries of $\mathcal E$ survive with their signs.  Also $\|\cD(\hat\bL)-\cD(\bL_0)\|_2\le\bmax^2R_N$, and Weyl's inequality shows that exactly $r$ eigenvalues of $\cD(\hat\bL)$ exceed $\bmax^2R_N$ in modulus; $\cD^{-1}$ preserves rank.
\end{proof}

The beta-min condition here is of Frobenius scale, $R_N\gtrsim\sqrt{m}\lambda$, whereas Theorem~\ref{thm:pdw} requires only order $\kappa_H\lambda$ and certifies the raw optimizer.

\begin{remark}[Comparison with the joint-cone route]\label{rem:compare}
Theorem~\ref{thm:pdw} needs irrepresentability, which Theorem~\ref{thm:main} does not, but in exchange it certifies the raw optimizer, requires only an order-$\kappa_H\lambda$ beta-min condition rather than the Frobenius-scale condition of the thresholded estimator in Section~\ref{sec:threshold}, and its constants involve the restricted Fisher information rather than $\kappa_{\mathrm{id}}$.
\end{remark}

\subsection{Robustness via the Joint Cone}

\begin{corollary}[Frobenius guarantee for the nominal target under contamination]
\label{cor:robust_cone}
On the event of Corollary~\ref{cor:robust_nominal}, if $\lambda\ge4(\varepsilon_\infty(P)+\delta_\Sigma)$, $\gamma\ge4(\varepsilon_2(P;\bB^{-1})+\bmin^{-2}\delta_\Sigma)$, and the nominal joint cone satisfies $\alpha_{\cC}>0$ and $R_N<\lambda_{\min}(\bOm_u^*)/2$, then $\|\hat\bS-\bS_u^*\|_F+\|\hat\bL-\bL^*\|_F\le R_N$.
\end{corollary}
\begin{proof}
By \eqref{eq:robust_noise}, \eqref{eq:dual_noise} holds; apply Theorem~\ref{thm:main} with $\bSig_0=\bSig_u^*$.
\end{proof}

\begin{corollary}[Point-mass contamination with the contaminated target]
\label{cor:point_mass}
Let $r_c\triangleq\rank(\bL_{u,c})\le r+1$, construct the joint cone at $(\bS_{u,c},\bL_{u,c})$, assume its restricted gain is positive with constant $\kappa_c$, and let $R_{N,c}$ be the radius \eqref{eq:RN} with this target, rank $r_c$, and concentration bounds under $P$.  If $R_{N,c}<\lambda_{\min}(\bOm_{u,c})/2$, then on the corresponding concentration event
\[
\|\hat\bS-\bS_u^*\|_F+\|\hat\bL-\bL^*\|_F\le R_{N,c}+\|\bS_{u,c}-\bS_u^*\|_F+\|\bL_{u,c}-\bL^*\|_F.
\]
\end{corollary}
\begin{proof}
Apply Theorem~\ref{thm:main} to the target $(\bS_{u,c},\bL_{u,c})$ and use the triangle inequality.
\end{proof}

\section{Simultaneous Guarantees over a Finite Candidate Family}
\label{sec:family}

\begin{theorem}[Simultaneous finite-family guarantee]
\label{thm:family}
Let $\Theta$ be a finite family of tuples $\theta=(s_\theta,\bB_\theta,\lambda_\theta,\gamma_\theta)$ with $M=|\Theta|$, and let $(\hat\bS_\theta,\hat\bL_\theta)$ solve \eqref{eq:estimator} with these parameters.  Let $R_\theta(\eta')$ be the radius of Corollary~\ref{cor:nominal} at failure probability $\eta'$, and assume its deterministic conditions hold for every $\theta$.  Then with probability at least $1-\eta$, simultaneously for all $\theta\in\Theta$,
\begin{align}
\|\hat\bS_\theta-\bS_u^*(s_\theta)\|_F+\|\hat\bL_\theta-\bL^*\|_F\le R_\theta(\eta/M).
\end{align}
Consequently, for every data-dependent $\hat\theta\in\Theta$, $\|P_{\mathrm{off}}(\hat\bS_{\hat\theta}-\bS^*)\|_F+\|\hat\bL_{\hat\theta}-\bL^*\|_F\le R_{\hat\theta}(\eta/M)$ on the same event.  The same union-bound argument applies verbatim to the conclusions of Corollary~\ref{cor:pdw_nominal}.
\end{theorem}
\begin{proof}
Apply Corollary~\ref{cor:nominal} to each $\theta$ at level $\eta/M$ and take a union bound; the event does not depend on how $\hat\theta$ is selected.  Proposition~\ref{prop:preservation} gives $P_{\mathrm{off}}\bS_u^*(s_\theta)=P_{\mathrm{off}}\bS^*$.
\end{proof}

This is a uniform guarantee, not an oracle inequality: it does not compare $R_{\hat\theta}$ with $\min_\theta R_\theta$.

\section{Misspecified Congruence Operator}
\label{sec:misspec}

Let $\hat\bB\in\Sym^p$ be positive definite, e.g.\ $\hat\bB=(\bI+\hat\tau\hat\bK^\top\hat\bK)^{1/2}$ for a misspecified pair $(\hat\bK,\hat\tau)$, and let $\hat\cD(\bL)=\hat\bB\bL\hat\bB$.

\begin{proposition}[Invariance of the estimand]\label{prop:invariance}
Let $\hat \bB\succ0$ be fixed, or independent of the sample used in~\eqref{eq:estimator}. The
target $(\bS_0,\bL_0)$ in~\eqref{eq:target}, and hence the estimand graph $\mathsf G(\bS_0)$ and
the rank and inertia of $\bL_0$, are determined by $\bm\Sigma_0$ alone and do not
depend on the congruence operator. Theorems~\ref{thm:main},
\ref{thm:threshold}, and~\ref{thm:pdw} hold verbatim for the estimator built with
$\hat \bB$, with every $\bB$-dependent quantity ($\widetilde{\mathcal T}$,
$\mathcal D^{-1}$, $\varpi_B$, $\chi_B$, $\nu$, $\kappa_{\rm id}$, $\kappa_H$,
$\alpha$) computed at $\hat \bB$.
\end{proposition}

\begin{proof}
The proofs use only that $\bB$ is symmetric positive definite and
nonrandom. If $\hat \bB$ is computed from an independent sample, apply them
conditionally on $\hat \bB$.
\end{proof}

\begin{remark}[What misspecification does and does not affect]
\label{rem:misspec}
Proposition~\ref{prop:invariance} concerns the estimand, not the estimator. For
fixed $(\lambda,\gamma)$, program~\eqref{eq:estimator} built with $\hat \bB$ is a different
optimization problem, and in general
$(\hat \bS_{\hat \bB},\hat\bL_{\hat \bB})\neq(\hat \bS_\bB,\hat \bL_\bB)$. In particular, the
finite-sample regularization bias, which is present for every $\bB$ because both
penalties shrink, depends on $\hat \bB$. Misspecification of the congruence
operator does not alter the underlying estimand, although it affects the
finite-sample regularization geometry and hence the estimation error through
the tangent space $\widetilde{\mathcal T}$, the dual noise level
$\|\hat \bB^{-1}\bW\hat \bB^{-1}\|_2$, the conditioning constant $\kappa_H$, the
irrepresentability margin $\alpha$, and the spectral gap $\nu$. What is
invariant are the qualitative conclusions: whenever the conditions of
Theorem~\ref{thm:pdw} hold at $\hat \bB$, the estimator built with $\hat \bB$
recovers the same signed graph $\mathsf G(\bS_0)$ and the same rank and inertia
as the one built with $\bB$.
\end{remark}

The next result shows that the PDW conditions are stable. On $\Sym^p\times\Sym^p$ let $P_{\cX_B}$ be the projector onto $\cX_B\triangleq\mathcal R\times\Tt_B$, where $\Tt_B$ is the tangent space at $\bB\bL_0\bB$, let $\cM_B\triangleq\cA_B^*\cI\cA_B$ with $\cA_B(\bA,\bY)=\bA-\bB^{-1}\bY\bB^{-1}$, $\cH_B\triangleq P_{\cX_B}\cM_B|_{\cX_B}$, and define the full-space quantities
\begin{align*}
\kappa_H'(\bB)&\triangleq\|\cH_B^{-1}P_{\cX_B}\|_{g\to g},\\
1-\alpha'(\bB)&\triangleq\|P_{\cX_B^\perp}\cM_B\cH_B^{-1}P_{\cX_B}\|_{g\to g}.
\end{align*}
Since $P_{\cX_B}\mathbf q=\mathbf q$ for $\mathbf q\in\cX_B$, Assumptions~\ref{ass:invert}--\ref{ass:irrep} hold at $\bB$ with $\kappa_H\le\kappa_H'(\bB)$ and $\alpha\ge\alpha'(\bB)$ whenever $\alpha'(\bB)>0$.

\begin{proposition}[Stability of the witness conditions]
\label{prop:stability}
If $\cH_{\bB}$ is invertible, then $\bB'\mapsto\kappa_H'(\bB')$ and $\bB'\mapsto\alpha'(\bB')$ are continuous at $\bB$ on the cone of positive definite matrices.  In particular, if $\alpha'(\bB)>0$, there is a neighborhood of $\bB$ on which $\kappa_H'\le2\kappa_H'(\bB)$ and $\alpha'\ge\alpha'(\bB)/2$, and on which $\nu$, $\bmin$, $\bmax$, and $\|\bB\tilde\bU\|_2$ vary continuously.
\end{proposition}
\begin{proof}
The projector onto $\range(\bB'\bU)$ is $\bB'\bU(\bU^\top\bB'^2\bU)^{-1}\bU^\top\bB'$, continuous in $\bB'$; hence $P_{\Tt_{B'}}$ and $P_{\cX_{B'}}$ are continuous, and so is $\cM_{B'}$.  The operator $\mathcal K_{B'}\triangleq P_{\cX_{B'}}\cM_{B'}P_{\cX_{B'}}+P_{\cX_{B'}^\perp}$ on the fixed space $\Sym^p\times\Sym^p$ is block diagonal with respect to $\cX_{B'}\oplus\cX_{B'}^\perp$, invertible iff $\cH_{B'}$ is, and satisfies $\mathcal K_{B'}^{-1}P_{\cX_{B'}}=\cH_{B'}^{-1}P_{\cX_{B'}}$.  Inversion is continuous on invertible operators and $\mathcal K_{B'}$ is continuous, so $\mathcal K_{B'}$ is invertible near $\bB$ and $\cH_{B'}^{-1}P_{\cX_{B'}}$ is continuous there; all operator norms on a finite-dimensional space are continuous.  The eigenvalues of $\bB'\bL_0\bB'$ and the singular values of $\bB'$ are continuous.
\end{proof}

\begin{proposition}[Perturbation of conditioning and dual noise]
\label{prop:sensitivity}
Let $\hat\bB=\bB+\bE$ be symmetric with $\delta_B\triangleq\|\bE\|_2/\bmin<1$.  Then $\hat\bB\succ0$, $(\kappa_B-\delta_B)_+/(1+\delta_B)\le\kappa_{\hat B}\le(\kappa_B+\delta_B)/(1-\delta_B)$, and $\|\hat\bB^{-1}\bW\hat\bB^{-1}\|_2\le\|\bW\|_2/(\bmin^2(1-\delta_B)^2)$, and
\begin{equation}\label{eq:dualnoise-transfer}
\|\hat \bB^{-1}\bW\hat \bB^{-1}\|_2\le\frac{\|\bB^{-1}\bW\bB^{-1}\|_2}{(1-\delta_B)^2}.
\end{equation}
\end{proposition}
\begin{proof}
Weyl's inequality gives $|\sigma_j(\hat\bB)-\sigma_j(\bB)|\le\|\bE\|_2$ and $\lambda_{\min}(\hat\bB)\ge\bmin(1-\delta_B)>0$, from which the three bounds follow. For~\eqref{eq:dualnoise-transfer}, write
$\hat \bB^{-1}\bW\hat \bB^{-1}=(\hat \bB^{-1}\bB)(\bB^{-1}\bW\bB^{-1})(\bB\hat \bB^{-1})$ and note
that $\hat \bB^{-1}\bB=(\bI+\bB^{-1}\bE)^{-1}$ with $\|\bB^{-1}\bE\|_2\le\delta_B<1$, so
$\|\hat \bB^{-1}\bB\|_2=\|\bB\hat \bB^{-1}\|_2\le(1-\delta_B)^{-1}$.
\end{proof}

\begin{corollary}[Recovery under a nearby misspecified operator]
\label{cor:misspec}
Suppose $\alpha'(\bB)>0$ and let $\mathcal U_B$ be the neighborhood of
Proposition~\ref{prop:stability}, shrunk so that $\nu(\hat \bB)\ge\nu(\bB)/2$ and
$\delta_B\le1/2$ for $\hat \bB\in\mathcal U_B$. Let $\hat \bB\in\mathcal U_B$ be
independent of the sample. On the event of Corollary~\ref{cor:explicit} computed
at $\bB$, choose
\[
\lambda\ge\frac{24\,\varepsilon_\infty}{\alpha'(\bB)},\
\gamma\ge\frac{96\,\varepsilon_{2,B}}{\alpha'(\bB)} .
\]
If (C1), (C3), and (C4) hold with
$(\kappa_H,\alpha,\nu,b_-)=(2\kappa_H'(\bB),\,\alpha'(\bB)/2,\,\nu(\bB)/2,\,b_-(1-\delta_B))$,
then the estimator built with $\hat B$ satisfies conclusions (a)--(d) of
Theorem~\ref{thm:pdw} for the same estimand $(\bS_0,\bL_0)$; in particular,
$\|\hat \bL_{\hat \bB}-\bL_0\|_F\le\frac{16}{3}\sqrt{2r}\,\kappa'_H(\bB)\,\gamma/(b_-^2(1-\delta_B)^2)$.
\end{corollary}

\begin{proof}
By Proposition~\ref{prop:stability}, Assumptions~\ref{ass:invert}--\ref{ass:irrep}
hold at $\hat \bB$ with $\kappa_H\le2\kappa'_H(\bB)$ and $\alpha\ge\alpha'(\bB)/2$.
By~\eqref{eq:dualnoise-transfer} and $\delta_B\le1/2$,
$\|\hat{\mathcal D}^{-1}(\bW)\|_2\le4\varepsilon_{2,B}\le\gamma\,\alpha'(\bB)/24$,
and $\|\bW\|_\infty\le\lambda\,\alpha'(\bB)/24$; hence (C2) holds at $\hat \bB$.
Apply Theorem~\ref{thm:pdw} at $\hat \bB$ together with
Proposition~\ref{prop:invariance}.
\end{proof}
\begin{remark}
Proposition~\ref{prop:stability} is qualitative in $\bB$.  For perturbations of the Fisher information at fixed $\bB$, Theorem~\ref{thm:transfer} gives an explicit radius; a quantitative radius in $\bB$ would require perturbation bounds for $\cH_B^{-1}$ in the mixed norm $\|\cdot\|_g$ across different model spaces $\cX_B$, which we do not pursue.  No estimator-level Lipschitz bound in $\hat\bB$ is claimed.
\end{remark}

\section{Proof of the Convergence of the ADMM}
\label{app:admm}

\begin{proof}[Proof of Theorem~\ref{thm:admm}]
The functions $(\bOm,\bZ)\mapsto f(\bOm)+\gamma\|\bZ\|_*$ and $(\bS,\bL)\mapsto\lambda\|\bS\|_{1,\mathrm{off}}$ are proper, closed, and convex.  The point $(\bOm,\bZ,\bS,\bL)=(\bI,\mathbf 0,\bI,\mathbf 0)$ is feasible and lies in the relative interior of the domain.  Since the constraints are linear, a primal minimizer together with this Slater point yields a Karush--Kuhn--Tucker (KKT) point \cite[Cor.~28.2.2, Thm.~28.3]{rockafellar1970convex}.  The coefficient operator of $\mathbf x$ is the identity, so $\rho\,\mathrm{Id}\succ0$, and $\mathcal T+\rho\cQ^*\cQ=\rho\eta\,\mathrm{Id}\succ0$ while $\mathcal T\succeq0$ by \eqref{eq:Qnorm}.  These are exactly the hypotheses of the convergence theorem for the semi-proximal ADMM with unit dual step \cite[Thm.~B.1]{fazel2013hankel}, which yields convergence of the primal sequence to a solution and of the multiplier sequence to a dual solution.  The equivalence of \eqref{eq:split} and \eqref{eq:estimator} transfers the conclusion.
\end{proof}

\bibliographystyle{unsrt}
\bibliography{references_complete}

\end{document}